%% file: arXiv.tex
\documentclass[journal, 10pt, twocolumn]{IEEEtran}
\pdfoutput=1

\usepackage{cite}
\usepackage[pdftex]{graphicx}
\usepackage[cmex10]{amsmath}
\usepackage{algorithmic}
\usepackage{amssymb}
\usepackage{array}
\usepackage{mdwmath}
\usepackage{mdwtab}
\usepackage{eqparbox}
\usepackage{multirow}
\usepackage{fixltx2e}
\usepackage{enumerate}
\usepackage[hyphens]{url}
\usepackage{hyperref}
\usepackage{breakurl}
\usepackage{tabularx}
\usepackage{color}
\usepackage{dsfont}
\usepackage{bm}
\usepackage{bbm}

\newcommand{\convas}{\overset{\textrm{a.s.}}{\longrightarrow}}
\newcommand{\argmin}{\mathop{\rm arg \, min}}

\usepackage{multibib}
\newcites{Supp}{References}
\bstctlcite[@auxoutSupp]{BSTcontrol2}

\begin{document}

%%%%%%%%%%%%%%%%%%%%%%%%%%%%%%%%%%%%%%%%%%%%%%%%%%
% BEGIN MAIN
%%%%%%%%%%%%%%%%%%%%%%%%%%%%%%%%%%%%%%%%%%%%%%%%%%

\title{Low-rank and Adaptive Sparse Signal (LASSI) Models for Highly Accelerated Dynamic Imaging}

\author{Saiprasad~Ravishankar,~\IEEEmembership{Member,~IEEE,}~Brian~E.~Moore,~\IEEEmembership{Student~Member,~IEEE,}~Raj~Rao~Nadakuditi,~\IEEEmembership{Member,~IEEE,} and~Jeffrey~A.~Fessler,~\IEEEmembership{Fellow,~IEEE}%

\thanks{DOI: 10.1109/TMI.2017.2650960. Copyright (c) 2016 IEEE. Personal use of this material is permitted. However, permission to use this material for any other purposes must be obtained from the IEEE by sending a request to pubs-permissions@ieee.org.}

\thanks{S. Ravishankar and B. E. Moore are equal contributors. This work was supported in part by the following grants: ONR grant N00014-15-1-2141, DARPA Young Faculty Award  D14AP00086, ARO MURI grants  W911NF-11-1-0391 and 2015-05174-05, NIH grants R01 EB023618 and P01 CA 059827, and a UM-SJTU seed grant.}

\thanks{S. Ravishankar, B. E. Moore, R. R. Nadakuditi,  and J. A. Fessler are with the Department of Electrical Engineering and Computer Science, University of Michigan, Ann Arbor, MI, 48109 USA emails: (ravisha, brimoor, rajnrao, fessler)@umich.edu.}}

\maketitle

\begin{abstract}
\input{abstract3}
\end{abstract}


\IEEEpeerreviewmaketitle

\vspace{-0.34in}
\section{Introduction} \label{sec1}
\input{introduction6}

\vspace{-0.1in}
\section{Models and Problem Formulations} \label{sec2}
\input{formulations5}

\vspace{-0.1in}
\section{Algorithms and Properties} \label{sec3}
\input{algorithms6}

\vspace{-0.12in}
\section{Numerical Experiments} \label{sec4}
%\vspace{-0.05in}

\subsection{Framework} \label{sec4a}

\input{framework6}

\vspace{-0.1in}
\input{numericalconvergence5}

\vspace{-0.15in}
\input{resultsandcomparisons7}

\vspace{-0.15in}

\input{parametermodelstudies6}

\vspace{-0.15in}
\section{Conclusions} \label{sec5}  
\input{conclusion5}
\vspace{-0.12in}

\bibliographystyle{./IEEEtran}
\bibliography{./IEEEabrv,./arXiv}

%%%%%%%%%%%%%%%%%%%%%%%%%%%%%%%%%%%%%%%%%%%%%%%%%%
% END MAIN
%%%%%%%%%%%%%%%%%%%%%%%%%%%%%%%%%%%%%%%%%%%%%%%%%%

\newpage
\clearpage

%%%%%%%%%%%%%%%%%%%%%%%%%%%%%%%%%%%%%%%%%%%%%%%%%%
% BEGIN SUPPLEMENT
%%%%%%%%%%%%%%%%%%%%%%%%%%%%%%%%%%%%%%%%%%%%%%%%%%

% Title
{
\twocolumn[
\begin{center}
 \Huge Low-rank and Adaptive Sparse Signal (LASSI) Models for Highly Accelerated Dynamic Imaging: Supplementary Material
\vspace{0.2in}
\end{center}]
}

This supplement presents a proof of the low-rank atom update formula (in the LASSI algorithms), a review of the OptShrink low-rank estimator, and additional experimental results to accompany our manuscript \citeSupp{saibrianrajfes:supp}.

\section{Proof of Atom Update Formula} \label{app1}
\input{atomupdate}

\section{OptShrink Background} \label{app3}
\input{optshrinkbackground}

\section{Additional Results and Discussion of Numerical Experiments}
\input{additionalresults}

\bibliographystyleSupp{./IEEEtran}
\bibliographySupp{./IEEEabrv,./arXiv}

%%%%%%%%%%%%%%%%%%%%%%%%%%%%%%%%%%%%%%%%%%%%%%%%%%
% END SUPPLEMENT
%%%%%%%%%%%%%%%%%%%%%%%%%%%%%%%%%%%%%%%%%%%%%%%%%%

\end{document}

%% file: abstract3.tex
Sparsity-based approaches have been popular in many applications in image processing and imaging. Compressed sensing exploits the sparsity of images in a transform domain or dictionary to improve image recovery from undersampled measurements.
In the context of inverse problems in dynamic imaging, recent research has demonstrated the promise of sparsity and low-rank techniques.
For example, the patches of the underlying data are modeled as sparse in an adaptive dictionary domain, and the resulting image and dictionary estimation from undersampled measurements is called dictionary-blind compressed sensing, or the dynamic image sequence is modeled as a sum of low-rank and sparse (in some transform domain) components (L+S model) that are estimated from limited measurements.
In this work, we investigate a data-adaptive extension of the L+S model, dubbed LASSI, where the temporal image sequence is decomposed into a low-rank component and a component whose spatiotemporal (3D) patches are sparse in some adaptive dictionary domain. We investigate various formulations and efficient methods for jointly estimating the underlying dynamic signal components and the spatiotemporal dictionary from limited measurements. 
We also obtain efficient sparsity penalized dictionary-blind compressed sensing methods as special cases of our LASSI approaches.
Our numerical experiments demonstrate the promising performance of LASSI schemes for dynamic magnetic resonance image reconstruction from limited k-t space data compared to recent methods such as k-t SLR and L+S, and compared to the proposed dictionary-blind compressed sensing method.

%inverse problems such as those in dynamic imaging. 

%Sparsity-based approaches have been popular in many applications in image processing and imaging. Compressed sensing exploits the sparsity of images in a transform domain or dictionary to improve image recovery from undersampled measurements. In the context of inverse problems in dynamic imaging, recent research has demonstrated the promise of sparsity and low-rank techniques. For example, the dynamic image sequence is modeled as a sum of low-rank and sparse (in some transform domain) components (L+S model) that are estimated from undersampled measurements. In this work, we investigate a data-adaptive extension of the L+S model, where the temporal image sequence is decomposed into a low-rank component and a component whose spatiotemporal (3D) patches are sparse in some adaptive dictionary domain. We investigate various formulations and highly efficient methods to jointly estimate the underlying dynamic signal components and the spatiotemporal dictionary from undersampled measurements. Our numerical experiments demonstrate the promising performance of our schemes for dynamic magnetic resonance image reconstruction from highly undersampled k-t space data compared to recent methods such as k-t SLR and L+S, and compared to a proposed efficient dictionary-blind compressed sensing method. 

%% file: introduction6.tex
Sparsity-based techniques are popular in many applications in image processing and imaging. Sparsity in either a fixed or data-adaptive dictionary or transform is fundamental to the success of popular techniques such as compressed sensing that aim to reconstruct images from limited sensor measurements. In this work, we focus on low-rank and adaptive dictionary-sparse models for dynamic imaging data %(e.g., dynamic magnetic resonance imaging data, or videos), 
and exploit such models to perform image reconstruction from limited (compressive) measurements. In the following, we briefly review compressed sensing (CS), CS-based magnetic resonance imaging (MRI), and dynamic data modeling, before outlining the contributions of this work.

\vspace{-0.1in}
\subsection{Background}\label{sec1a}

CS \cite{don, tao1, feng96a, BreFen-C96c} is a popular technique that enables recovery of signals or images from far fewer measurements (or at a lower rate) than the number of unknowns or than required by Nyquist sampling conditions. CS assumes that the underlying signal is sparse in some transform domain or dictionary and that the measurement acquisition procedure is incoherent in an appropriate sense with the dictionary.
CS has been shown to be very useful for MRI \cite{lustig, lustig2}.
MRI is a relatively slow modality because the data, which are samples in the Fourier space (or k-space) of the object, are acquired sequentially in time.
In spite of advances in scanner hardware and pulse sequences, the rate at which MR data are acquired is limited by MR physics and  physiological constraints \cite{lustig}. 
%on RF energy deposition \cite{lustig}.

%a popular diagnostic imaging modality that is non-invasive and enables excellent visualization of both anatomical structure and physiological function. However, it is

%(see \cite{feng96a, BreFen-C96c, Fen-PT97, VenBre-C98b, BreGasVen-C99, GasBre-C00a, YeBreMou-J02, Bre-C2008a} for early versions of CS for Fourier-sparse signals and for Fourier imaging)

CS has been applied to a variety of MR techniques such as static MRI \cite{lustig, josh, Yoo}, dynamic MRI (dMRI) \cite{lustig2, kal, urs, jong3}, parallel imaging (pMRI) \cite{liang1, ota4, wu, liu}, and perfusion imaging and diffusion tensor imaging (DTI) \cite{gan}. 
For static MR imaging, CS-based MRI (CSMRI) involves undersampling the k-space data (e.g., collecting fewer phase encodes) using random sampling techniques to accelerate data acquisition.
However, in dynamic MRI the data is inherently undersampled because the object is changing as the data is being collected, so in a sense \emph{all} dynamic MRI scans (of k-t space) involve some form of CS because one must reconstruct the dynamic images from under-sampled data.
The traditional approach to this problem in MRI is to use ``data sharing'' where data is pooled in time to make sets of k-space data (e.g., in the form of a Casorati matrix \cite{zliang16}) that appear to have sufficient samples, but these methods do not fully model the temporal changes in the object.
CS-based dMRI can achieve improved temporal (or spatial) resolution by using more explicit signal models rather than only implicit k-space data sharing, albeit at the price of increased computation.

%by collecting fewer (quicker) k-t measurements.
%CS exploits s  fewer measurements in k-t space in dMRI, and still recover the underlying images by exploiting sparsity.
%from fewer (quicker) k-t measurements

%CS has been applied to parallel imaging (pMRI) \cite{liang1, ota4} and to dynamic MRI (dMRI) \cite{urs, lustig2}.

CSMRI reconstructions with fixed, non-adaptive signal models (e.g., wavelets or total variation sparsity) typically suffer from artifacts at high undersampling factors \cite{bresai}. Thus, there has been growing interest in image reconstruction methods where the dictionary is adapted to provide highly sparse representations of data. 
Recent research has shown benefits for such data-driven adaptation of dictionaries \cite{ols, eng, elad, Mai}
%in many applications \cite{elad2, elad3, elad5, elad6, gyu, bresai}. 
in many applications \cite{elad2, elad3, super2, bresai}.
%Recent research has shown benefits for such data-driven adaptation of dictionaries \cite{ols, eng, elad, Mai} in applications such as denoising, inpainting, deblurring, demosaicing, super-resolution, and medical image reconstruction \cite{elad2, elad3, elad5, elad6, mai23, super2, gusup, bresai}.
%While the adaptation of synthesis dictionaries \cite{ols, eng, elad, Yagh, skret, Mai} has been extensively studied, recent work has shown advantages in terms of computation and application-specific performance, for the adaptation of transform models \cite{sabres, sbclsTS2, saiwen}. 
For example, the  DLMRI method \cite{bresai} jointly estimates the image and a synthesis dictionary for the image patches from undersampled k-space measurements.  The model there is that the unknown (vectorized) image patches can be well approximated by a sparse linear combination of the columns or atoms of a learned (a priori unknown) dictionary $D$.
This idea of joint dictionary learning and signal reconstruction from undersampled measurements \cite{bresai}, known as (dictionary) blind compressed sensing (BCS) \cite{Glei12}, has been the focus of several recent works (including for dMRI reconstruction) \cite{bresai, symul, lingal16, lingal655, wangying, wangying2, josecab1, huangbays, awate, josecab2, swang1}. 
The BCS problem is harder than conventional (non-adaptive) compressed sensing. However, the dictionaries learned in BCS typically reflect the underlying image properties better than pre-determined models, thus improving image reconstructions.
%syber, sabressiims1

%Another well-known and classical model for signals is the PCA (principal component analysis), where the data is assumed to live in a low-dimensional subspace (i.e., a low-rank model).
%In an early work \cite{fowl1}, Fowler proposed a method for recovering the principal eigenvectors of data (principal components) from random projections. This work shares similarities with BCS in its attempt to estimate a model for data from compressive measurements. However, while the prior work \cite{fowl1} learns an under-complete principal components (low-rank) model, BCS enables the learning of much richer (union of subspaces \cite{vidal2}) data models by exploiting sparsity criteria.

While CS methods use sparse signal models, various alternative models have been explored for dynamic data in recent years. Several works have demonstrated the efficacy of low-rank models (e.g., by constraining the Casorati data matrix to have low-rank) for dynamic MRI reconstruction \cite{zliang16, hald11, bzhao14, peder1}.
A recent work \cite{locallow} also considered a low-rank property for local space-time image patches.
For data such as videos (or collections of related images \cite{peng55}), there has been growing interest in decomposing the data into the sum of a low-rank (L) and a sparse (S) component \cite{Can2, venka12, vas1}. In this L+S (or equivalently Robust Principal Component Analysis (RPCA) \cite{Can2}) model, the L component may capture the background of the video, while the S component captures the sparse (dynamic) foreground. The L+S model has been recently shown to be promising for CS-based dynamic MRI \cite{ota1, ben11}. The S component of the L+S decomposition could either be sparse by itself or sparse in some known dictionary or transform domain. 
%While some works considered modeling the image sequence in dMRI as both low-rank and sparse (L \& S) \cite{lingal1, bo1} (with some recent works \cite{Majumdar33} learning dictionaries for the S part of L \& S), the more general (flexible) L+S model may provide better quality reconstructions \cite{ota1}.
Some works alternatively consider modeling the dynamic image sequence as both low-rank and sparse (L \& S) \cite{lingal1, bo1}, with a recent work \cite{Majumdar33} learning dictionaries for the S part of L \& S. In practice, which model provides better image reconstructions may depend on the specific properties of the underlying data.

%Majumdar2,

When employing the L+S model, the CS reconstruction problem can be formulated as follows:
\begin{align*}
 (\text{P0}) \;\; \min_{x_{L},\, x_{S}} &  \frac{1}{2} \left \| A(x_{L}+ x_{S})-d \right \|_{2}^{2} + \lambda_{L}\left \| R_{1}(x_{L}) \right \|_{*} \\
 &\; \; \; \; + \lambda_{S} \left \| T x_{S} \right \|_{1}.
\end{align*}
In (P0), the underlying unknown dynamic object is $x = x_{L} + x_{S} \in \mathbb{C}^{N_{x}N_{y} N_{t}}$, where $x_{L}$ and $x_{S}$ are vectorized versions of space-time (3D) tensors corresponding to $N_{t}$ temporal frames, each an image\footnote{We focus on 2D + time for simplicity but the concepts generalize readily to 3D + time.} of size $N_{x} \times N_{y}$.
%where L and S are space-time matrices, where each column corresponds to a temporal frame. Each temporal frame is a vectorized version of an Nx x Ny image.
The operator $A$ is the sensing or encoding operator and $d$ denotes the (undersampled) measurements. For parallel imaging with $N_{c}$ receiver coils, applying the operator $A$ involves frame-by-frame multiplication by coil sensitivities followed by applying an undersampled Fourier encoding (i.e., the SENSE method) \cite{prues2}. The operation $R_{1}(x_{L}) $ reshapes $x_{L}$ into an $N_{x}N_{y}  \times N_{t}$ matrix, and $\left \| \cdot \right \|_{*}$ denotes the nuclear norm that sums the singular values of a matrix. The nuclear norm serves as a convex surrogate for matrix rank in (P0). Traditionally, the operator $T$ in (P0) is a \emph{known} sparsifying transform for $x_{S}$, and $\lambda_{L}$ and $\lambda_{S}$ are non-negative weights.

\vspace{-0.12in}
\subsection{Contributions} \label{sec1b}

This work investigates in detail the extension of the L+S model for dynamic data to a Low-rank + Adaptive Sparse SIgnal (LASSI) model. In particular, we decompose the underlying temporal image sequence into a low-rank component and a component whose overlapping spatiotemporal (3D) patches are assumed sparse in some \emph{adaptive} dictionary domain\footnote{The LASSI method differs from the scheme in \cite{Majumdar44} that is not (overlapping) patch-based and involves only a 2D (spatial) dictionary.
The model in \cite{Majumdar44} is that $R_{1}(x_{S}) = D Z$ with sparse $Z$ and the atoms of $D$ have size $N_{x}N_{y}$ (typically very large). Since often $N_{t} < N_{x}N_{y}$, one can easily construct trivial (degenerate) sparsifying dictionaries (e.g., $D=R_{1}(x_{S})$) in this case. 
On the other hand, in our framework, the dictionaries are for small spatiotemporal patches, and there are many such overlapping patches for a dynamic image sequence to enable the learning of rich models that capture local spatiotemporal properties.}.
We propose a framework to jointly estimate the underlying signal components and the spatiotemporal dictionary from limited measurements. We compare using $\ell_{0}$ and $\ell_{1}$ penalties for sparsity in our formulations, and also investigate adapting structured dictionaries, where the atoms of the dictionary, after being reshaped into space-time matrices are low-rank.
%Structured dictionary models may be less prone to over-fitting in applications involving limited or corrupted data. 
The proposed iterative LASSI reconstruction algorithms involve efficient block coordinate descent-type updates of the dictionary and sparse coefficients of patches, and an efficient proximal gradient-based update of the signal components.
We also obtain novel sparsity penalized dictionary-blind compressed sensing methods as special cases of our LASSI approaches.
%efficient 

Our experiments demonstrate the promising performance of the proposed data-driven schemes for dMRI reconstruction from limited k-t space data. In particular, we show that the LASSI methods give much improved reconstructions compared to the recent L+S method and methods involving joint L \& S modeling \cite{lingal1}.
We also show improvements with LASSI compared to the proposed spatiotemporal dictionary-BCS methods (that are special cases of LASSI). 
%and improvements compared to a proposed efficient dictionary-BCS method (that is a variation of LASSI).
Moreover, learning structured dictionaries and using the $\ell_{0}$ sparsity ``norm'' in LASSI are shown to be advantageous in practice.
Finally, in our experiments, we compare the use of conventional singular value thresholding (SVT) for updating the low-rank signal component in the LASSI algorithms to alternative approaches including the recent OptShrink method \cite{nadakuditi2013, moore2014a, moore2014b}.

A short version of this work investigating a specific LASSI method appears elsewhere \cite{saibrrajfes1jk}. Unlike \cite{saibrrajfes1jk}, here, we study several dynamic signal models and reconstruction approaches in detail, and illustrate the convergence and learning behavior of the proposed methods, and demonstrate their effectiveness for several datasets and undersampling factors.

%We also compare two approaches for updating the low-rank signal component in our experiments: one based on singular value thresholding (SVT), and another based on the recent OptShrink method.

\vspace{-0.15in}
\subsection{Organization} \label{sec1c}

The rest of this paper is organized as follows. Section \ref{sec2} describes our models and problem formulations for dynamic image reconstruction. Section \ref{sec3} presents efficient algorithms for the proposed problems and discusses the algorithms' properties. Section \ref{sec4} presents experimental results demonstrating the convergence behavior and performance of the proposed schemes for the dynamic MRI application. Section \ref{sec5} concludes with proposals for future work.

%% file: formulations5.tex
\subsection{LASSI Formulations}

We model the dynamic image data as $x = x_{L} + x_{S}$, where $x_{L}$ is low-rank when reshaped into a (space-time) matrix, and we assume that the spatiotemporal (3D) patches in the vectorized tensor $x_{S}$ are sparse in some adaptive dictionary domain. 
We replace the regularizer $\zeta(x_{s}) = \left \| T x_{S} \right \|_{1}$ with weight $\lambda_{S}$ in (P0) with the following patch-based dictionary learning regularizer
\begin{align}
 \zeta(x_{s}) = & \min_{D,Z}\:  \sum_{j=1}^{M}\left \| P_{j}x_{S} - D z_{j} \right \|_{2}^{2} + \lambda_Z^{2}\left \| Z \right \|_{0} \label{rrr} \\
\nonumber &\; \mathrm{s.t.}\; \: \left \| Z \right \|_{\infty} \leq a,~\text{rank}\left ( R_{2}(d_{i}) \right ) \leq r, \,~\left \| d_{i} \right \|_2 =1 \,\,\forall \, i
\end{align}
to arrive at the following  problem for joint image sequence reconstruction and dictionary estimation:
\begin{align*}
(\text{P1}) \; & \min_{D, Z, x_{L}, x_{S}} \frac{1}{2} \left \| A(x_{L}+x_{S})-d \right \|_{2}^{2} + \lambda_{L}\left \| R_{1}(x_{L}) \right \|_{*} \\
& \; \; \; \; \; \; + \lambda_{S} \begin{Bmatrix}
\sum_{j=1}^{M}\left \| P_{j}x_{S} - D z_{j} \right \|_{2}^{2} + \lambda_Z^{2}\left \| Z \right \|_{0}
\end{Bmatrix}\\
& \;\;\; \text{s.t.}\; \left \| Z \right \|_{\infty} \leq a,~\text{rank}\left ( R_{2}(d_{i}) \right ) \leq r, \,~\left \| d_{i} \right \|_2 =1 \,\,\forall \, i.
\end{align*}
Here, $P_{j}$ is a patch extraction matrix that extracts an $m_{x} \times m_{y} \times m_{t}$ spatiotemporal patch from $x_{S}$ as a vector. A total of $M$ (spatially and temporally) overlapping 3D patches are assumed.
Matrix $D \in \mathbb{C}^{m \times K}$ with $m = m_{x} m_{y} m_{t}$ is the synthesis dictionary to be learned and $z_{j} \in \mathbb{C}^{K}$ is the unknown sparse code for the $j$th patch, with $P_{j}x_{S} \approx D z_{j}$.

We use $Z \in \mathbb{C}^{K \times M}$ to denote the matrix that has the sparse codes $z_{j}$ as its columns, $\left \| Z \right \|_{0}$ (based on the $\ell_{0}$ ``norm'') counts the number of nonzeros in the matrix $Z$, and $\lambda_Z \geq 0$.
Problem (P1) penalizes the number of nonzeros in the (entire) coefficient matrix $Z$, allowing variable sparsity levels across patches.
This is a general and flexible model for image patches (e.g., patches from different regions in the dynamic image sequence may contain different amounts of information and therefore all patches may not be well represented at the same sparsity) and leads to promising performance in our experiments.
The constraint $\left \| Z \right \|_{\infty} \triangleq \max_{j} \left \| z_{j} \right \|_{\infty}
 \leq a$ with $a>0$ is used in (P1) because the objective (specifically the regularizer \eqref{rrr}) is non-coercive with respect to $Z$ \cite{sairajfes}. \footnote{Such a non-coercive function remains finite even in cases when $\left \|Z \right \| \to \infty$. For example, consider a dictionary $D$ that has a column $d_{i}$ that repeats. Then, in this case, the patch coefficient vector $z_{j}$ in (P1) could have entries $\alpha$ and $- \alpha$ respectively, corresponding to the two repeated atoms in $D$, and the objective would be invariant to arbitrarily large scaling of $\left | \alpha \right |$ (i.e., non-coercive).}
The $\ell_{\infty}$ constraint prevents pathologies that could theoretically arise (e.g., unbounded algorithm iterates) due to the non-coercive objective. In practice, we set $a$ very large, and the constraint is typically inactive.
%(see \cite{sairajfes}).

The atoms or columns of $D$, denoted by $d_{i}$, are constrained to have unit norm in (P1) to avoid scaling ambiguity between $D$ and $Z$ \cite{kar, sairajfes}. 
We also model the reshaped dictionary atoms $R_2(d_{i})$ as having rank at most $r>0$, where the operator $R_{2}(\cdot)$ reshapes $d_{i}$ into a $m_{x} m_{y} \times m_{t}$ space-time matrix.
Imposing low-rank (small $r$) structure on reshaped dictionary atoms is motivated by our empirical observation that the dictionaries learned on image patches (without such a constraint) tend to have reshaped atoms with only a few dominant singular values.
Results included in the supplement\footnote{Supplementary material is available in the supplementary files/multimedia tab.} show that dictionaries learned on dynamic image patches with low-rank atom constraints tend to represent such data as well as learned dictionaries with full-rank atoms.
Importantly, such structured dictionary learning may be less prone to over-fitting in scenarios involving limited or corrupted data. We illustrate this for the dynamic MRI application in Section \ref{sec4}.

When  $z_{j}$ is highly sparse (with $\left \| z_{j} \right \|_{0} \ll \min (m_{t}, m_{x} m_{y})$) and $R_{2}(d_{i})$ has low rank (say rank-1), the model $P_{j}x_{S} \approx D z_{j}$ corresponds to approximating the space-time patch matrix as a sum of a few reshaped low-rank (rank-1) atoms. This special (extreme) case would correspond to approximating the patch itself as low-rank.
However, in general the decomposition $D z_{j}$ could involve numerous ($> \min (m_{t}, m_{x} m_{y})$) active atoms, corresponding to a rich, not necessarily low-rank, patch model. Experimental results in Section \ref{sec4} illustrate the benefits of such rich models.

%This rank constraint enables local low-rank and sparse modeling in $x_{S}$. 

Problem (P1) jointly learns a decomposition $x = x_{L} + x_{S}$ and a dictionary $D$ along with the sparse coefficients $Z$ (of spatiotemporal patches) from the measurements $d$. Unlike (P0), the fully-adaptive Problem (P1) is nonconvex.
An alternative to (P1) involves replacing the $\ell_{0}$ ``norm'' with the convex $\ell_{1}$ norm (with  $\left \| Z \right \|_{1} = \sum_{j=1}^{M}\left \| z_{j} \right \|_{1}$) as follows:
\begin{align*}
(\text{P2}) \; & \min_{D, Z, x_{L}, x_{S}} \frac{1}{2} \left \| A(x_{L}+x_{S})-d \right \|_{2}^{2} + \lambda_{L}\left \| R_{1}(x_{L}) \right \|_{*} \\
& \; \; \; \; \; \; + \lambda_{S} \begin{Bmatrix}
\sum_{j=1}^{M}\left \| P_{j}x_{S} - D z_{j} \right \|_{2}^{2} + \lambda_Z \left \| Z \right \|_{1}
\end{Bmatrix}\\
& \;\;\; \text{s.t.}\; \left \| Z \right \|_{\infty} \leq a,~\text{rank}\left ( R_{2}(d_{i}) \right ) \leq r, \,~\left \| d_{i} \right \|_2 =1 \,\,\forall \, i.
\end{align*} 
Problem (P2) is also nonconvex due to the product $D z_{j}$ (and the nonconvex constraints), so the question of choosing (P2) or (P1) is one of image quality, not convexity.

Finally, the convex nuclear norm penalty $\left \| R_{1}(x_{L}) \right \|_{*}$ in (P1) or (P2) could be alternatively replaced with a nonconvex penalty on the rank of $R_{1}(x_{L})$, or the function $\left \| \cdot \right \|_{p}^{p}$ for $p<1$ (based on the Schatten $p$-norm) that is applied to the vector of singular values of $R_{1}(x_{L})$ \cite{lingal1}.
While we focus mainly on the popular nuclear norm penalty in our investigations, we also briefly study some of the alternatives in Section \ref{sec3} and Section \ref{sec4d}.

%as it leads to a simple update procedure for $x_{L}$ (cf. Section \ref{sec3})
%the alternative $\ell_{p}$ penalty

\vspace{-0.16in}
\subsection{Special Case of LASSI Formulations: Dictionary-Blind Image Reconstruction}

When $\lambda_{L} \to \infty$ in (P1) or (P2), the optimal low-rank component of the dynamic image sequence becomes inactive (zero).
The problems then become pure spatiotemporal dictionary-blind image reconstruction problems (with $x_{L}=0$ and $x=x_{S}$) involving $\ell_{0}$ or $\ell_{1}$ overall sparsity \cite{sairajfes} penalties. For example, Problem (P1) reduces to the following form:
\begin{align}
\nonumber & \min_{D, Z, x} \frac{1}{2} \left \| Ax-d \right \|_{2}^{2} + \lambda_{S} \begin{Bmatrix}
\sum_{j=1}^{M}\left \| P_{j}x - D z_{j} \right \|_{2}^{2} + \lambda_Z^{2}\left \| Z \right \|_{0}
\end{Bmatrix}\\
& \;\; \text{s.t.}\; \left \| Z \right \|_{\infty} \leq a,~\text{rank}\left ( R_{2}(d_{i}) \right ) \leq r, \,~\left \| d_{i} \right \|_2 =1 \,\,\forall \, i. \label{dino11}
\end{align}

\vspace{-0.05in}
We refer to formulation \eqref{dino11} with its low-rank atom constraints as the DINO-KAT (DIctioNary with lOw-ranK AToms) blind image reconstruction problem. A similar formulation is obtained from (P2) but with an $\ell_{1}$ penalty.
These formulations differ from the ones proposed for dynamic image reconstruction in prior works such as \cite{lingal16, josecab2}, \cite{wangying2}.
In \cite{josecab2}, dynamic image reconstruction is performed by learning a common real-valued dictionary for the spatio-temporal patches of the real and imaginary parts of the dynamic image sequence. 
The algorithm therein involves dictionary learning using K-SVD \cite{elad}, where sparse coding is performed using the approximate and expensive orthogonal matching pursuit method \cite{pati}.
In contrast, the algorithms in this work (cf. Section \ref{sec3}) for the overall sparsity penalized  DINO-KAT blind image reconstruction problems involve simple and efficient updating of the \emph{complex-valued} spatio-temporal dictionary (for complex-valued 3D patches) and sparse coefficients (by simple thresholding) in the formulations.
The advantages of employing sparsity penalized dictionary learning over conventional approaches like K-SVD are discussed in more detail elsewhere \cite{sairajfes}.
In \cite{wangying2}, a spatio-temporal dictionary is learned for the complex-valued 3D patches of the dynamic image sequence (a total variation penalty is also used), but the method again involves dictionary learning using K-SVD.
In the blind compressed sensing method of \cite{lingal16}, the time-profiles of individual image pixels were modeled as sparse in a learned dictionary. The 1D voxel time-profiles are a special case of general overlapping 3D (spatio-temporal) patches. Spatio-temporal dictionaries as used here may help capture redundancies in both spatial and temporal dimensions in the data.
Finally, unlike the prior works, the DINO-KAT schemes in this work involve structured dictionary learning with low-rank reshaped atoms.

%% file: algorithms6.tex
\subsection{Algorithms} \label{sec2b}

We propose efficient block coordinate descent-type algorithms for (P1) and (P2), where, in one step, we update $(D,Z)$ keeping $(x_{L},x_{S})$ fixed (\emph{Dictionary Learning Step}), and then we update $(x_{L},x_{S})$ keeping $(D,Z)$ fixed (\emph{Image Reconstruction Step}). We repeat these alternating steps in an iterative manner. The algorithm for the DINO-KAT blind image reconstruction problem \eqref{dino11} (or its $\ell_{1}$ version) is similar, except that $x_{L}=0$ during the update steps. Therefore, we focus on the algorithms for (P1) and (P2) in the following.

\subsubsection{Dictionary Learning Step}   %{Update of $D$ and $Z$}
Here, we optimize (P1) or (P2) with respect to $(D, Z)$. We first describe the update procedure for (P1).
Denoting by $P$ the matrix that has the patches $P_{j}x_{S}$ for $1 \leq j \leq M$ as its columns, and with $C \triangleq Z^{H}$, the optimization problem with respect to $(D,Z)$ in the case of (P1) can be rewritten as follows:
\begin{align*}
(\text{P3}) \; & \min_{D, C} \left \| P - DC^{H} \right \|_{F}^{2} + \lambda_Z^{2}\left \| C \right \|_{0}\\
& \;\; \text{s.t.}\; \left \| C \right \|_{\infty} \leq a, ~\text{rank}\left ( R_{2}(d_{i}) \right ) \leq r, \, ~\left \| d_{i} \right \|_2 =1 \,\,\forall \, i.
\end{align*}
Here, we express the matrix $D C^{H}$ as a Sum of OUter Products (SOUP) $\sum_{i=1}^{K} d_{i}c_{i}^{H}$.
We then employ an iterative block coordinate descent method for (P3), where the columns $c_{i}$ of $C$ and atoms $d_{i}$ of $D$ are updated sequentially by cycling over all $i$ values \cite{sairajfes}. Specifically, for each $1 \leq i \leq K$, we solve (P3) first with respect to $c_{i}$ (\emph{sparse coding}) and then with respect to $d_{i}$ (\emph{dictionary atom update}).

For the minimization with respect to $c_{i}$, we have the following subproblem, where $E_{i} \triangleq P - \sum_{k\neq i} d_{k}c_{k}^{H}$ is computed using the most recent estimates of the other variables:
\begin{equation} \label{eqop5}
\min_{c_{i} \in \mathbb{C}^{M}} \; \begin{Vmatrix}
E_{i} - d_{i}c_{i}^{H}
\end{Vmatrix}_{F}^{2} + \lambda_{Z}^{2} \left \| c_{i} \right \|_{0}  \;\; \mathrm{s.t.}\; \: \left \| c_{i} \right \|_{\infty} \leq a.
\end{equation}
The minimizer $\hat{c}_{i}$ of \eqref{eqop5} is given by \cite{sairajfes}
\begin{equation} \label{tru1ch4}
\hat{c}_{i} =  \min\left ( \begin{vmatrix}
H_{\lambda_{Z}} \left ( E_{i}^{H}d_{i} \right )
\end{vmatrix}, a 1_{M} \right ) \, \odot \, e^{j \angle  E_{i}^{H}d_{i}},
\end{equation}
where the hard-thresholding operator $H_{\lambda_{Z}} (\cdot)$ zeros out vector entries with magnitude less than $\lambda_{Z}$ and leaves the other entries (with magnitude $\geq \lambda_{Z}$) unaffected.
Here, $\left | \cdot  \right |$ computes the magnitude of vector entries, $1_{M}$ denotes a vector of ones of length $M$, ``$\odot$'' denotes element-wise multiplication, $\min(\cdot, \cdot)$ denotes element-wise minimum, and we choose $a$ such that $a > \lambda_{Z}$.
For a vector $c \in \mathbb{C}^{M}$, $e^{j \angle c} \in \mathbb{C}^{M}$ is computed element-wise, with ``$\angle$'' denoting the phase.

Optimizing (P3) with respect to the atom $d_{i}$ while holding all other variables fixed yields the following subproblem:
\begin{equation} \label{eqop6}
 \min_{d_{i} \in \mathbb{C}^{m}} \; \begin{Vmatrix}
E_{i} - d_{i}c_{i}^{H}
\end{Vmatrix}_{F}^{2}  \;\: \mathrm{s.t.}\; \: \text{rank}\left ( R_{2}(d_{i}) \right ) \leq r, \, \left \| d_{i} \right \|_2 =1.
\end{equation}

Let $U_{r} \Sigma_{r} V_{r}^{H}$ denote an optimal rank-$r$ approximation to $ R_{2}\left ( E_{i}c_{i} \right ) \in \mathbb{C}^{m_{x} m_{y} \times m_{t}}$ that is obtained using the $r$ leading singular vectors and singular values of the full singular value decomposition (SVD) $ R_{2}\left ( E_{i}c_{i} \right ) \triangleq U \Sigma V^{H}$. Then a global minimizer of \eqref{eqop6}, upon reshaping, is
\begin{equation} \label{tru1ch4g}
R_{2}(\hat{d}_{i}) =  \left\{\begin{matrix}
\frac{U_{r} \Sigma_{r} V_{r}^{H}
}{\left \| \Sigma_{r} \right \|_{F}}, & \mathrm{if}\,\, c_{i}\neq 0 \\ 
W , & \mathrm{if}\,\, c_{i}= 0 
\end{matrix}\right. \\
\end{equation}
where $W$ is any normalized matrix with rank at most $r$, of appropriate dimensions (e.g., we use the reshaped first column of the $m \times m$ identity matrix).  The proof for \eqref{tru1ch4g} is included in the supplementary material.

%The proof for \eqref{tru1ch4g} is included in Appendix \ref{app1}.
If $r=\min(m_x m_y, m_{t})$, then no SVD is needed and the solution is \cite{sairajfes}
\begin{equation} \label{tru1ch4gtyu6}
\hat{d}_{i} =  \left\{\begin{matrix}
\frac{E_{i}c_{i}}{\left \| E_{i}c_{i} \right \|_{2}}, & \mathrm{if}\,\, c_{i}\neq 0 \\ 
w, & \mathrm{if}\,\, c_{i}= 0 
\end{matrix}\right.
\end{equation}
where $w$ is any vector on the $m$-dimensional unit sphere (e.g., we use the first column of the $m \times m $ identity).
%The solution is unique if and only if $\mathbf{c}_{j}\neq 0$. 

%not included here due to space constraints and is presented elsewhere \cite{saibrianrajfes}.

In the case of (P2), when minimizing with respect to $(D, Z)$, we again set $C= Z^{H}$, which yields an $\ell_{1}$ penalized dictionary learning problem (a simple variant of (P3)).
The dictionary and sparse coefficients are then updated using a similar block coordinate descent method as for (P3). In particular, the coefficients $c_{i}$ are updated using soft thresholding: 
\begin{equation}
\hat{c}_{i} =  \max \left (\begin{vmatrix}
  E_{i}^{H}d_{i}
\end{vmatrix} - \frac{\lambda_{Z}}{2} 1_{M}, \, 0 \right ) \, \odot \, e^{j \angle \, E_{i}^{H}d_{i} }. \label{tru1ch4gg}
\end{equation}

\subsubsection{Image Reconstruction Step} %{Update of $x_L$ and $x_S$}

Minimizing (P1) or (P2) with respect to $x_{L}$ and $x_{S}$ yields the following subproblem:
\begin{align*}
(\text{P4}) \; & \min_{x_{L}, x_{S}} \frac{1}{2} \left \| A(x_{L}+ x_{S})-d \right \|_{2}^{2} + \lambda_{L}\left \| R_{1}(x_{L}) \right \|_{*} \\
& \; \; \; \; \; \;\;\; + \lambda_{S} \sum_{j=1}^{M}\left \| P_{j}x_{S} - D z_{j} \right \|_{2}^{2}.
\end{align*}
Problem (P4) is convex but nonsmooth, and its objective has the form
%Problem (P4) is convex but nonsmooth. The objective in (P4) can be written in the form
$f(x_L, x_S) + g_{1}(x_L) + g_{2}(x_S)$, with $f(x_L, x_S) \triangleq 0.5 \left \| A(x_{L} + x_{S}) - d \right \|_{2}^{2}$, $g_{1}(x_L) \triangleq \lambda_{L}\left \| R_{1}(x_{L}) \right \|_{*}$, and $g_{2}(x_S) \triangleq  \lambda_{S} \sum_{j=1}^{M}\left \| P_{j}x_{S} - D z_{j} \right \|_{2}^{2}$. 
We employ the proximal gradient method \cite{ota1} for (P4), 
%Similar to prior work \cite{ota1}, we employ a proximal gradient method for (P4). 
whose iterates, denoted by superscript $k$, take the following form:
%The iterates of the proximal gradient scheme, denoted by superscript $k$, take the following form:
\begin{align}
x_{L}^{k} & = \text{prox}_{t_{k}g_{1}}(x_{L}^{k-1} - t_{k} \nabla_{x_L}f(x_{L}^{k-1},x_{S}^{k-1})), \label{ls1} \\
x_{S}^{k} & = \text{prox}_{t_{k}g_{2}}(x_{S}^{k-1} - t_{k} \nabla_{x_S}f(x_{L}^{k-1},x_{S}^{k-1})), \label{ls2}
\end{align}
where the proximity function is defined as
\begin{equation} \label{rr}
\text{prox}_{t_{k}  g}(y) = \underset{z}{\arg \min} \; \frac{1}{2} \left \| y-z \right \|_{2}^{2} + t_{k} \, g(z),
\end{equation}
and the gradients of $f$ are given by
\begin{equation*}
\nabla_{x_L}f(x_{L}, x_{S}) =\nabla_{x_S}f(x_{L}, x_{S}) = A^{H}A(x_{L} + x_{S}) - A^{H}d.
\end{equation*}

%%%%%%%%%%%%%%%%%%%%%%%%%%%%%%%%%%%%%%%%%%%%%%%%%
% Old version
%%%%%%%%%%%%%%%%%%%%%%%%%%%%%%%%%%%%%%%%%%%%%%%%%
\iffalse
The update in \eqref{ls1} corresponds to the singular value thresholding (SVT) operation \cite{cai1}.
Specifically, denoting by $Q \Lambda W^{H}$ the SVD of $R_{1}(\tilde{x}_{L}^{k-1})$, where $\tilde{x}_{L}^{k-1} \triangleq $ $ x_{L}^{k-1} - t_{k} \nabla_{x_L}f(x_{L}^{k-1},x_{S}^{k-1})$, it follows from \eqref{ls1} and \eqref{rr} \cite{cai1} that $R_{1}(x_{L}^{k}) = Q \hat{\Lambda} W^{H}$, where $\hat{\Lambda}$ is a diagonal matrix with entries
\begin{align} \label{ls3}
\hat{\Lambda}_{ii} & =  (\Lambda_{ii}  - t_{k} \lambda_{L})^{+},
\end{align}
and $(\cdot)^{+} = \max(\cdot,0)$ sets negative values to zero.
\fi
%%%%%%%%%%%%%%%%%%%%%%%%%%%%%%%%%%%%%%%%%%%%%%%%%

%%%%%%%%%%%%%%%%%%%%%%%%%%%%%%%%%%%%%%%%%%%%%%%%%
% New version
%%%%%%%%%%%%%%%%%%%%%%%%%%%%%%%%%%%%%%%%%%%%%%%%%
The update in \eqref{ls1} corresponds to the singular value thresholding (SVT) operation \cite{cai1}.
Indeed, defining $\tilde{x}_{L}^{k-1} \triangleq x_{L}^{k-1} - t_{k} \nabla_{x_L}f(x_{L}^{k-1},x_{S}^{k-1})$, it follows from \eqref{ls1} and \eqref{rr} \cite{cai1} that
\begin{equation} \label{Lupdate}
R_{1}(x_{L}^{k}) = \mathbf{SVT}_{t_{k} \lambda_{L}}(R_{1}(\tilde{x}_{L}^{k-1})).
\end{equation}
%Here, the SVT operator is defined for a given threshold $\tau > 0$ as
Here, the SVT operator for a given threshold $\tau > 0$ is
\begin{equation} \label{svt_defn}
\mathbf{SVT}_{\tau}(Y) = \sum_i (\sigma_i - \tau)^+u_i v_i^H,
\end{equation}
where $U \Sigma V^H$ is the SVD of $Y$ with $\sigma_{i}$ denoting the $i$th largest singular value and $u_{i}$ and $v_{i}$ denoting the $i$th columns of $U$ and $V$, and $(\cdot)^{+} = \max(\cdot,0)$ sets negative values to zero.
%%%%%%%%%%%%%%%%%%%%%%%%%%%%%%%%%%%%%%%%%%%%%%%%%

Let $\tilde{x}_{S}^{k-1} \triangleq x_{S}^{k-1} - t_{k} \nabla_{x_S}f(x_{L}^{k-1},x_{S}^{k-1})$. 
%Then it follows from \eqref{ls2} and \eqref{rr} that  $x_{S}^{k}$ satisfies the following Normal equation:
Then \eqref{ls2} and \eqref{rr} imply that  $x_{S}^{k}$ satisfies the following Normal equation:
\begin{equation} \label{ls4}
\begin{pmatrix}
I + 2 t_k \lambda_{S} \sum_{j=1}^{M}P_{j}^{T}P_{j}
\end{pmatrix} x_{S}^{k} = \tilde{x}_{S}^{k-1} +  2 t_k \lambda_{S} \sum_{j=1}^{M} P_{j}^{T}Dz_{j}.
\end{equation}
Solving \eqref{ls4} for $x_{S}^{k}$ is straightforward because the matrix pre-multiplying $x_{S}^{k}$ is diagonal, and thus its inverse can be computed cheaply. The term $2 t_{k} \lambda_{S}  \sum_{j=1}^{M} P_{j}^{T}Dz_{j}$ in \eqref{ls4} can also be computed cheaply using patch-based operations.

%using (spatiotemporal) patch-based operations.

The proximal gradient method for (P4) converges \cite{pat2} for a constant step-size $t_{k} = t < 2/\ell$, where $\ell$ is the Lipschitz constant of $\nabla f(x_L, x_S)$. For (P4), \mbox{$\ell = 2 \left \| A \right \|_{2}^{2}$}.
In practice, 
$\ell$
%this Lipschitz constant
can be precomputed using standard techniques such as the power iteration method.
In our dMRI experiments in Section \ref{sec4}, we normalize the encoding operator $A$ so that $\|A\|_{2} = 1$ for fully-sampled measurements (cf. \cite{ota1, ota2}) to ensure that $\left \| A \right \|_{2}^{2} \leq 1$ in undersampled (k-t space) scenarios.

When the nuclear norm penalty in (P4) is replaced with a rank penalty, i.e., $g_{1}(x_L) \triangleq \lambda_{L} \, \text{rank}(R_{1}(x_{L}))$,
the proximity function is a modified form of the 
SVT
%singular value thresholding 
operation in \eqref{Lupdate} (or \eqref{svt_defn}), where the singular values smaller than $\sqrt{2 t_{k} \lambda_{L}}$ are set to zero and the other singular values are left unaffected (i.e., hard-thresholding the singular values).
Alternatively, when the nuclear norm penalty is replaced with $\left \| \cdot \right \|_{p}^{p}$ (for $p<1$) applied to the vector of singular values of $R_{1}(x_{L})$ \cite{lingal1}, the proximity function can still be computed cheaply when $p = 1/2$ or $p = 2/3$, for which the soft thresholding of singular values in \eqref{svt_defn} is replaced with the solution of an appropriate polynomial equation (see \cite{woodworth}). For general $p$, the $x_{L}$ update could be performed using strategies such as in \cite{lingal1}.

% OptShrink discussion
\input{optshrinkmaintext3}

%%%%%%%%%%%%%%%%%%%%%%%%%%%%%%%%%%%%%%%%%%%%%%%%%%
%\iffalse
%To ensure the convergence of the proximal gradient method \cite{pat2}, we choose a constant step-size $t_{k} = t < 2/l$, where $l$ is the Lipschitz constant of $\nabla f(x_L, x_S)$. In our case, $l = 2 \left \| A \right \|_{2}^{2}$. For dMRI, by normalizing the encoding operator $A$ so that $\|A\|_{2} = 1$ for fully-sampled measurements, we ensure that $\left \| A \right \|_{2}^{2} < 1$ in undersampled scenarios, and thus the universal step-size $t=1$ suffices \cite{ota1}.
%\fi

\input{overallalgorithm4}

%\subsection{Overall Algorithm} \label{sec2c}

\vspace{-0.12in}
\subsection{Convergence and Computational Cost} \label{sec2d}

The proposed LASSI algorithms for (P1) and (P2) alternate between updating $(D,Z)$ and $(x_{L},x_{S})$. Since we update the dictionary atoms and sparse coefficients using an exact block coordinate descent approach, the objectives in our formulations only decrease in this step. When the $(x_{L},x_{S})$ update is performed using proximal gradients (which is guaranteed to converge to the global minimizer of (P4)), by appropriate choice of the constant-step size \cite{Parikha1}, the objective functions can be ensured to be monotone (non-increasing) in this step. Thus, the costs in our algorithms are monotone decreasing, and because they are lower-bounded (by $0$), they must converge.
Whether the iterates in the LASSI algorithms converge to the critical points \cite{vari1} in (P1) or (P2) \cite{sairajfes} is an interesting question that we leave for future work. 

In practice, the computational cost per outer iteration of the proposed algorithms is dominated by the cost of the dictionary learning step, which scales (assuming $K \propto m$ and $M \gg K, m$) as $O(m^{2} M J)$, where $J$ is the number of times the matrix $D$ is updated in the dictionary learning step. 
The SOUP dictionary learning cost is itself dominated by various matrix-vector products, whereas the costs of the truncated hard-thresholding \eqref{tru1ch4} and low-rank approximation \eqref{tru1ch4g} steps are negligible.
On the other hand, when dictionary learning is performed using methods like K-SVD \cite{elad} (e.g., in \cite{bresai, wangying}), the associated cost (assuming per-patch sparsity $\propto m$) may scale worse\footnote{In \cite{sairajfes}, we have shown that efficient SOUP learning-based image reconstruction methods outperform methods based on K-SVD in practice.}
as $O(m^{3} M J)$. 
Section \ref{sec4} illustrates that our algorithms converge quickly in practice.

%(which is ensured to converge to the critical points \cite{vari1} of the $(D,Z)$ update problems \cite{sairajfes})

%% file: optshrinkmaintext3.tex
The nuclear norm-based low-rank regularizer $\|R_1(x_L)\|_{\star}$ is popular because it is the tightest convex relaxation of the (nonconvex) matrix rank penalty. However, this does not guarantee that the nuclear norm (or its alternatives) is the optimal (in any sense) low-rank regularizer in practice.
Indeed, the argument $R_{1}(\tilde{x}_{L}^{k-1})$ of the SVT operator in \eqref{Lupdate} can be interpreted as an estimate of the underlying (true) low-rank matrix $R_1(x_L)$ plus a residual (noise) matrix. In \cite{nadakuditi2013}, the low-rank denoising problem was studied from a random-matrix-theoretic perspective and an algorithm -- OptShrink -- was derived that asymptotically achieves minimum squared error among all estimators that shrink the singular values of their argument. We leverage this result for dMRI by proposing the following modification of  \eqref{Lupdate}:
%We leverage this result for dMRI by proposing a modified version of  \eqref{Lupdate}, where we replace the SVT-based $x_L$ update with
\begin{equation} \label{Loptupdate}
R_{1}(x_{L}^{k}) = \mathbf{OptShrink}_{r_L}(R_{1}(\tilde{x}_{L}^{k-1})).
\vspace{-0.04in}
\end{equation}
Here, $\mathbf{OptShrink}_{r_L}(.)$ is the data-driven OptShrink estimator from Algorithm~1 of \cite{nadakuditi2013}  (see the supplementary material for more details and discussion of OptShrink).
%(see Appendix \ref{app2} for more details and discussions).
In this variation, the regularization parameter $\lambda_L$ is replaced by a parameter $r_L \in \mathbb{N}$ that directly specifies the rank of $R_1(x_L^k)$, and the (optimal) shrinkage for each of the leading $r_L$ singular values is implicitly estimated based on the distribution of the remaining singular values.
%In this variation, the regularization parameter $\lambda_L$ is replaced by a parameter $r_L \in \mathbb{N}$ that directly specifies the rank of $R_{1}(x_{L}^{k})$.
Intuitively, we expect this variation of the aforementioned (SVT-based) proximal gradient scheme to yield better estimates of the underlying low-rank component of the reconstruction because, at each iteration $k$ (in \eqref{ls1}), the OptShrink-based update \eqref{Loptupdate} should produce an estimate of the underlying low-rank matrix $R_{1}(x_L)$ with smaller squared error than the corresponding SVT-based update \eqref{Lupdate}.
Similar OptShrink-based schemes have shown promise in practice \cite{moore2014a,moore2014b}. In particular, in \cite{moore2014a} it is shown that replacing the SVT-based low-rank updates in the algorithm \cite{ota1} for (P0) with OptShrink updates can improve dMRI reconstruction quality. In practice, small $r_L$ values perform well due to the high spatio-temporal correlation of the background in dMRI.

%In this LASSI variation, the regularization parameter $\lambda_L$ is replaced by a parameter $r_L \in \mathbb{N}$ that directly specifies the rank of $R_{1}(x_{L})$. Intuitively, we expect this LASSI variation to yield better estimates of the underlying low-rank component of the reconstruction because, at each iteration, the OptShrink-based update \eqref{Loptupdate} should produce an estimate of the underlying low-rank matrix $R_{1}(x_L)$ with smaller squared error than the corresponding SVT-based update \eqref{Lupdate}.

%Similar OptShrink-based update schemes have shown promise in practice \cite{moore2014a,moore2014b}. In particular, in \cite{moore2014a} it is shown that replacing the SVT-based low-rank updates for (P0) with OptShrink-based updates can improve dMRI reconstruction quality. In practice, small $r_L$ values perform well due to the high spatio-temporal correlation of the background in dMRI.

%% file: overallalgorithm4.tex
\begin{figure}
\begin{tabular}{p{8.3cm}}
\hline
Algorithms for (P1) and (P2)\\
\hline
 \textbf{Inputs\;:} \:\:\: measurements $d$, weights $\lambda_{L}$, $\lambda_{S}$, and $\lambda_{Z}$, rank $r$, upper bound $a$, number of dictionary learning iterations $J$, number of proximal gradient iterations $\tilde{J}$, and number of outer iterations $\hat{J}$.\\
 \textbf{Outputs\;:} \:\:\: reconstructed dynamic image sequence components $x_{L}^{\hat{J}}$ and $x_{S}^{\hat{J}}$, learned dictionary $D^{\hat{J}}$, and learned coefficients of patches $Z^{\hat{J}}$. \\
\textbf{Initial Estimates:} $\left ( x_{L}^{0}, x_{S}^{0}, D^{0}, Z^{0} \right )$, with $C^{0} = \left ( Z^{0} \right )^{H}$.\\
\textbf{For \;$t$ = $1:$ $\hat{J}$ repeat}\\
\vspace{-0.1in}
\begin{enumerate}
\item Form $P^{t-1}=\left [ P_{1}x_{S}^{t-1} \mid P_{2}x_{S}^{t-1} \mid ... \mid P_{M} x_{S}^{t-1}  \right ]$. \vspace{0.04in}
\item \textbf{Dictionary Learning:}
With training data $P^{t-1}$ and initialization $\left ( D^{t-1}, C^{t-1} \right )$, update $(c_{i}, d_{i})$ sequentially for $1 \leq i \leq K$ using \eqref{tru1ch4} (or \eqref{tru1ch4gg}) and \eqref{tru1ch4g}.
Set $\left ( D^{t}, C^{t} \right )$ to be the output after $J$ cycles of such updates, and $Z^{t} = \left ( C^{t} \right )^{H}$.
\item \textbf{Image Reconstruction:} Update $x_{L}^{t}$ and $x_{S}^{t}$ using $\tilde{J}$ iterations of the proximal gradient scheme using \eqref{ls1} and \eqref{ls2}, and with initialization $\left ( x_{L}^{t-1}, x_{S}^{t-1} \right )$.
\end{enumerate}\\
\textbf{End} \\
\hline
\end{tabular}
\caption{The LASSI reconstruction algorithms for Problems (P1) and (P2), respectively. Superscript $t$ denotes the iterates in the algorithm.
We do not compute the matrices $E_{i} \triangleq P - \sum_{k\neq i} d_{k}c_{k}^{H}$ explicitly in the dictionary learning iterations. Rather, we efficiently compute products of $E_{i}$ or $E_{i}^{H}$ with vectors \cite{sairajfes}. Parameter $a$ is set very large in practice (e.g., $a \propto \left \| A^{\dagger}d \right \|_{2}$).} \label{im6p}
\vspace{-0.24in}
\end{figure}

%as the outputs of the proximal gradient scheme below. \newline
%\textbf{For \;$k$ = $1:$ $\tilde{J}$ repeat}
%\begin{equation}
%x_{L}^{k}  = \text{prox}_{t_{k}g_{1}}(x_{L}^{k-1} - t_{k} \nabla_{x_L}f(x_{L}^{k-1},x_{S}^{k-1})), \label{ls1a} 
%\end{equation}
%\begin{equation}
%x_{S}^{k}  = \text{prox}_{t_{k}g_{2}}(x_{S}^{k-1} - t_{k} \nabla_{x_S}f(x_{L}^{k-1},x_{S}^{k-1})), \label{ls2a}
%\end{equation}
%\textbf{End} 

%\mathbf{D}=\mathbf{D}^{t-1}

Fig. \ref{im6p} shows the LASSI reconstruction algorithms for Problems (P1) and (P2), respectively.
As discussed, we can obtain variants of these proposed LASSI algorithms by replacing the SVT-based $x_{L}$ update \eqref{Lupdate} in the image reconstruction step with an OptShrink-based update \eqref{Loptupdate}, or with the update 
%(involving hard thresholding of singular values) 
arising from the rank penalty or from the Schatten $p$-norm ($p<1$) penalty.
The proposed LASSI algorithms start with an initial $\left ( x_{L}^{0}, x_{S}^{0}, D^{0}, Z^{0} \right )$. For example, $D^{0}$ can be set to an analytical dictionary, 
%(e.g., DCT), 
$Z^{0}=0$, and $x_{L}^{0}$ and $x_{S}^{0}$ could be (for example) set based on some iterations of the recent L+S method \cite{ota1}.
%When $\lambda_{L} = \infty$ in (P1) and (P2), the low-rank component of the dynamic image sequence becomes inactive (zero). The problems then become dictionary-blind compressed sensing problems (with $x=x_{S}$) involving $\ell_{0}$ or $\ell_{1}$ (overall) sparsity penalties, and 
In the case of Problem \eqref{dino11}, the proposed algorithm is an efficient SOUP-based image reconstruction algorithm.  We refer to it as the DINO-KAT image reconstruction algorithm in this case.

%% file: framework6.tex
The proposed LASSI framework can be used for inverse problems involving dynamic data, such as in dMRI, interventional imaging, video processing, etc.
Here, we illustrate the convergence behavior and performance of our methods for dMRI reconstruction from limited k-t space data.
Section \ref{sec4b} focuses on empirical convergence and learning behavior of the methods. Section \ref{sec4c} compares the image reconstruction quality obtained with LASSI to that obtained with recent techniques. Section \ref{sec4d}  investigates and compares the various LASSI models and methods in detail.
We compare using the $\ell_{0}$ ``norm'' (i.e., (P1)) to the $\ell_{1}$ norm (i.e., (P2)), structured (with low-rank atoms) dictionary learning to the learning of unstructured (with full-rank atoms) dictionaries, and singular value thresholding-based $x_{L}$ update to OptShrink-based or other alternative $x_{L}$ updates in LASSI. We also investigate the effects of the sparsity level (i.e., number of nonzeros) of the learned $Z$ and the overcompleteness of $D$ in LASSI, and demonstrate the advantages of adapting the patch-based LASSI dictionary compared to using fixed dictionary models in the LASSI algorithms. The LASSI methods are also shown to perform well for various initializations of $x_{L}$ and $x_{S}$.

%We perform simulations with several dMRI datasets used in prior works \cite{ota1, lingal1}: 
We work with several dMRI datasets from prior works \cite{ota1, lingal1}:
1) the Cartesian cardiac perfusion data \cite{ota1, ota2}, 2) a 2D cross section of the physiologically improved nonuniform cardiac torso (PINCAT) \cite{shari12} phantom data (see \cite{lingal1, ktslr44}), and 3) the \emph{in vivo} myocardial perfusion MRI data in \cite{lingal1, ktslr44}.
The 
%first 
cardiac perfusion data were acquired with a modified TurboFLASH sequence on a 3T scanner using a 12-element coil array. The fully sampled 
%perfusion image 
data with an image matrix size of $128 \times 128$ (128 phase encode lines) and $40$ temporal frames was acquired with  $\text{FOV} = 320 \times 320 \text{ mm}^{2}$, slice thickness = 8 mm, spatial resolution = $3.2$ mm$^{2}$, and temporal resolution of 307 ms \cite{ota1}.
The coil sensitivity maps 
%for this data
are provided in \cite{ota2}.
The (single coil) PINCAT data (as in \cite{ktslr44}) had image matrix size of $128 \times 128$ and $50$ temporal frames.
The single coil \emph{in vivo} myocardial perfusion data was acquired on a 3T scanner using a saturation recovery FLASH sequence with Cartesian sampling (TR/TE = 2.5/1 ms, saturation recovery time = 100 ms), and had a image matrix size of $90 \times 190$ (phase encodes $\times$ frequency encodes) and $70$ temporal frames \cite{lingal1}.

%(normalized to unit peak intensity in image space)
%along $k_{y}$

Fully sampled data (PINCAT and \emph{in vivo} data were normalized to unit peak image intensity, and the cardiac perfusion data \cite{ota1} had a peak image intensity of 1.27) were retrospectively undersampled in our experiments. We used Cartesian and pseudo-radial undersampling patterns.
In the case of Cartesian sampling, we used a different variable-density random Cartesian undersampling pattern for each time frame.
The pseudo-radial (sampling radially at uniformly spaced angles for each time frame and with a small random rotation of the radial lines between frames) sampling patterns 
%in our experiments 
were obtained by subsampling on a Cartesian grid for each time frame.
We simulate several undersampling (acceleration) factors of k-t space in our experiments.
%, each with new randomly generated undersampling patterns, in our experiments.
We measure the quality of the dMRI reconstructions using the normalized root mean square error (NRMSE) metric defined as
%\begin{equation}
%\mathrm{NRMSE} = \frac{\left \| x_{\mathrm{recon}} -x_{\mathrm{ref}}\right \|_{2}}{\left \| x_{\mathrm{ref}} \right \|_{2}},
%\end{equation}
$\left \| x_{\mathrm{recon}} -x_{\mathrm{ref}}\right \|_{2}/$ $\left \| x_{\mathrm{ref}} \right \|_{2}$,
where $x_{\mathrm{ref}}$ is a reference reconstruction from fully sampled data, and $x_{\mathrm{recon}}$ is the reconstruction from undersampled data.

We compare the quality of reconstructions obtained with the proposed LASSI methods to those obtained with the recent L+S method \cite{ota1} and the k-t SLR method involving joint L \& S modeling \cite{lingal1}. 
For the L+S and k-t SLR methods, we used the publicly available MATLAB implementations \cite{ota2, ktslr44}. 
We chose the parameters for both methods (e.g., $\lambda_{L}$ and $\lambda_{S}$ for L+S in (P0) or $\lambda_{1}$, $\lambda_{2}$, etc. for k-t SLR \cite{lingal1, ktslr44}) by sweeping over a range of values and choosing the settings that achieved good NRMSE in our experiments. We optimized parameters separately for each dataset to achieve the lowest NRMSE at some intermediate undersampling factors, and observed that these settings also worked well at other undersampling factors.
% The parameters were optimized separately for each dataset.
%at 8x undersampling. 
The L+S method was simulated for 250 iterations and k-t SLR was also simulated for sufficient iterations to ensure convergence. The operator $T$ (in (P0)) for L+S was set to a temporal Fourier transform, and a total variation sparsifying penalty (together with a nuclear norm penalty for enforcing low-rankness) was used in k-t SLR.
The dynamic image sequence in both methods was initialized with a baseline reconstruction (for the L+S method, L was initialized with this baseline and S with zero) that was obtained by first performing zeroth order interpolation at the non-sampled k-t space locations (by filling in with the nearest non-zero entry along time) and then backpropagating the filled k-t space to image space (i.e., pre-multiplying by the $A^{H}$ corresponding to fully sampled data).
%Specifically, we set $\lambda_{L} = 0.525$ and $\lambda_{S} = 0.01$, and we ran the L+S method for 250 iterations to ensure convergence. 

For the LASSI method, we extracted spatiotemporal patches of size $8 \times 8 \times 5$ from $x_{S}$ in (P1) with spatial and temporal patch overlap strides of 2 pixels.\footnote{While we used a stride of 2 pixels, a spatial and temporal patch overlap stride of 1 pixel would further enhance the reconstruction performance of LASSI in our experiments, but at the cost of substantially more computation.} 
The dictionary atoms were reshaped into $64 \times 5$ space-time matrices, and we set the rank parameter $r=1$, except for the invivo dataset \cite{lingal1, ktslr44}, where we set $r=5$. We ran LASSI for 50 outer iterations with 1 and 5 inner iterations in the $(D, Z)$ and $(x_{L}, x_{S})$ updates, respectively.
Since Problem (P1) is nonconvex, the proposed algorithm needs to be initialized appropriately.
We set the initial $Z = 0$, and the initial $x_{L}$ and $x_{S}$ were typically set based on the outputs of either the L+S or k-t SLR methods.
When learning a square dictionary, we initialized $D$ with a $320 \times 320$ DCT, and, in the overcomplete ($K>m$) case, we concatenated the square DCT initialization with normalized and vectorized patches that were selected from random locations of the initial reconstruction.
We empirically show in Section \ref{sec4d} that the proposed LASSI algorithms typically improve image reconstruction quality compared to that achieved by their initializations.
We selected the weights $\lambda_{L}$, $\lambda_{S}$, and $\lambda_{Z}$ for the LASSI methods separately for each dataset by sweeping over a range (3D grid) of values and picking the settings that achieved the lowest NRMSE at intermediate undersampling factors (as for L+S and k-t SLR) in our experiments. These tuned parameters also worked well at other undersampling factors (e.g., see Fig. \ref{imcvbcs33}(h)), and are included in the supplement for completeness.

We also evaluate the proposed variant of LASSI involving only spatiotemporal dictionary learning (i.e., dictionary blind compressed sensing).
We refer to this method as DINO-KAT dMRI, with $r=1$. We use an $\ell_{0}$ sparsity penalty for DINO-KAT dMRI (i.e., we solve Problem \eqref{dino11}) in our experiments, and the other parameters are set or optimized (cf. the supplement) similarly as described above for LASSI.

%Although tuned at a specific undersampling factor (same as for L+S and k-t SLR) for each dataset, 
%and did not vary much across the (normalized) datasets

The LASSI and DINO-KAT dMRI implementations were coded in Matlab R2016a.
Our current Matlab implementations are not optimized for efficiency. Hence, here we perform our comparisons to recent methods based on reconstruction quality (NRMSE) rather than runtimes, since the latter are highly implementation dependant.
A link to software to reproduce our results will be provided at \url{http://web.eecs.umich.edu/~fessler/}.
 
%In our experiments, we compared to recent methods in terms of reconstruction quality (NRMSE). The actual runtimes for LASSI would be quite dependant on the implementation. 

%with $\lambda_{L} = 1.2$ and $\lambda_S = 0.01$.

%we set $\lambda_{L}=0.5$, $\lambda_{S}=0.01$, and $\lambda_{Z}=0.03$  and we set $r=1$

%% file: numericalconvergence5.tex
\subsection{LASSI Convergence and Learning Behavior} \label{sec4b}

\begin{figure*}[!t]
\begin{center}
\begin{tabular}{cccccc}
\includegraphics[height=1.3in]{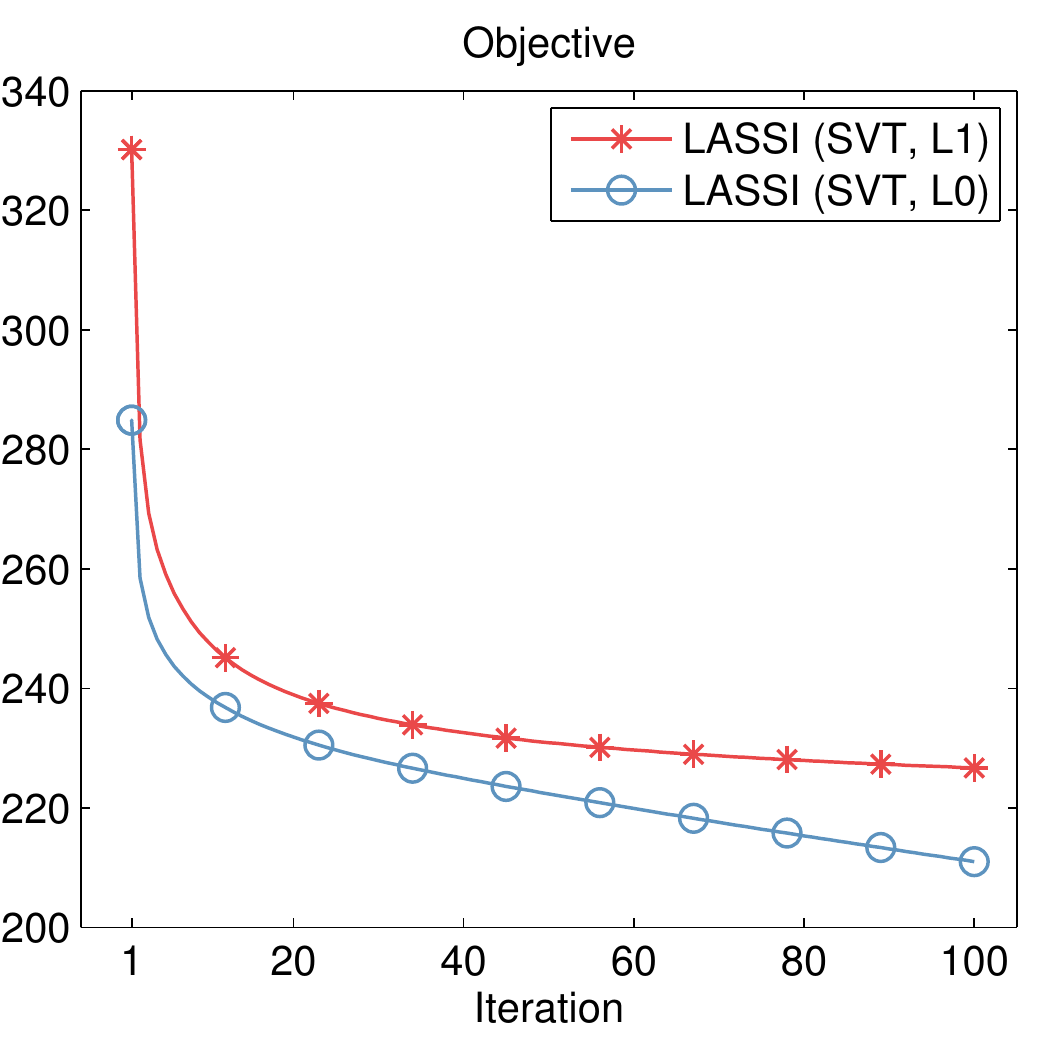}&
\includegraphics[height=1.3in]{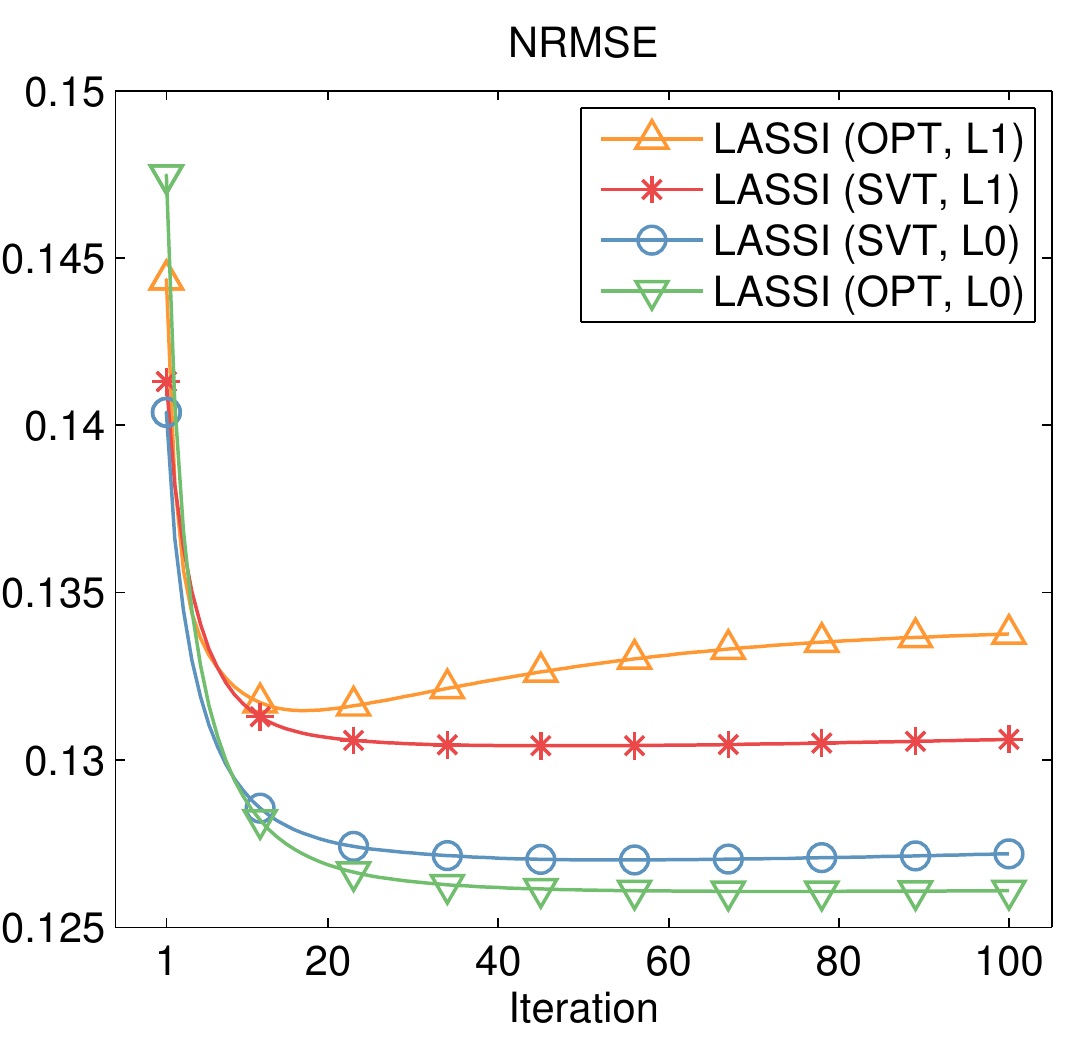}&
\includegraphics[height=1.3in]{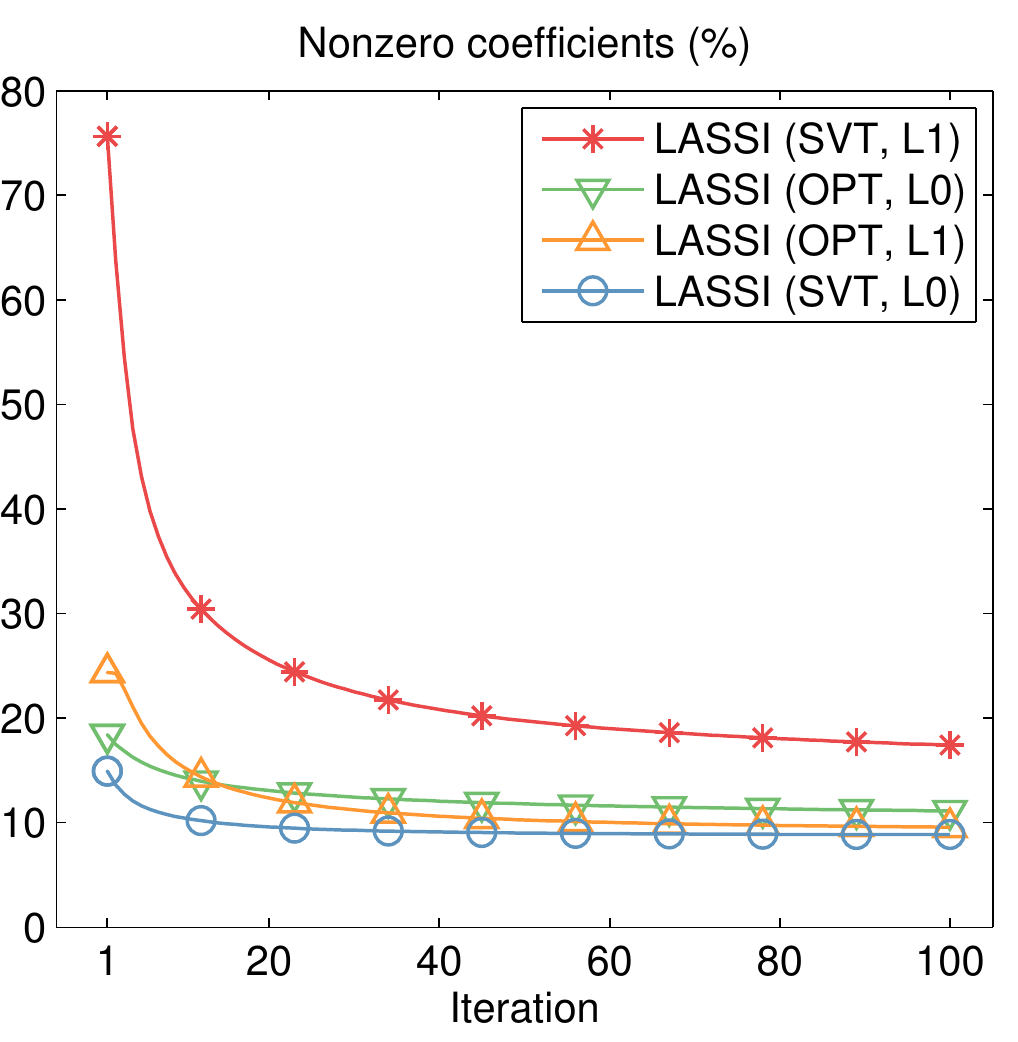}&
\includegraphics[height=1.3in]{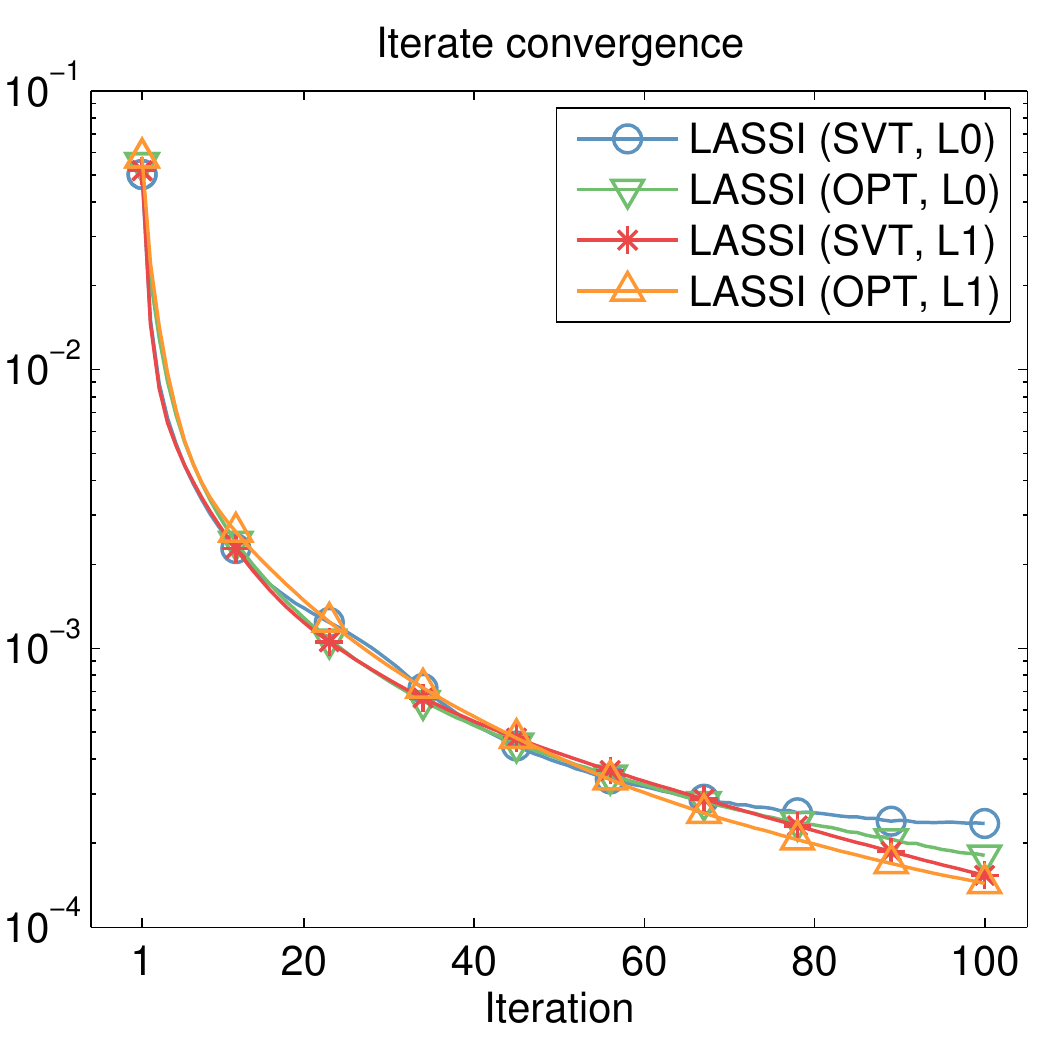}\\
(a) & (b) & (c) & (d)\\
\end{tabular}
\begin{tabular}{cc}
\includegraphics[height=0.85in]{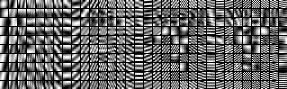}&
\includegraphics[height=0.85in]{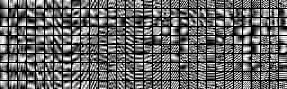}\\
(e) & (f) \\
\end{tabular}
\begin{tabular}{cc}
\includegraphics[height=1.15in]{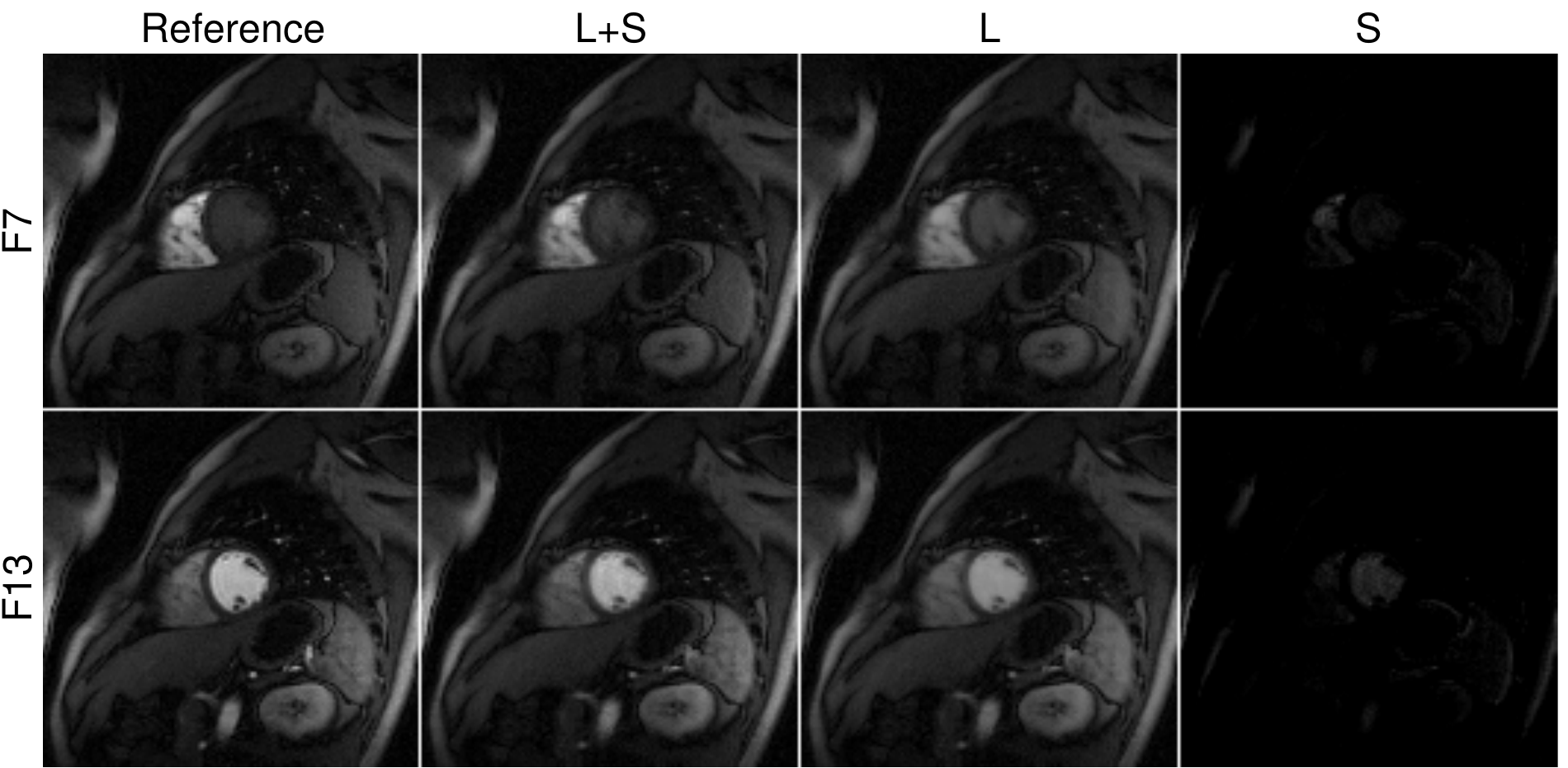}&
\includegraphics[height=1.15in]{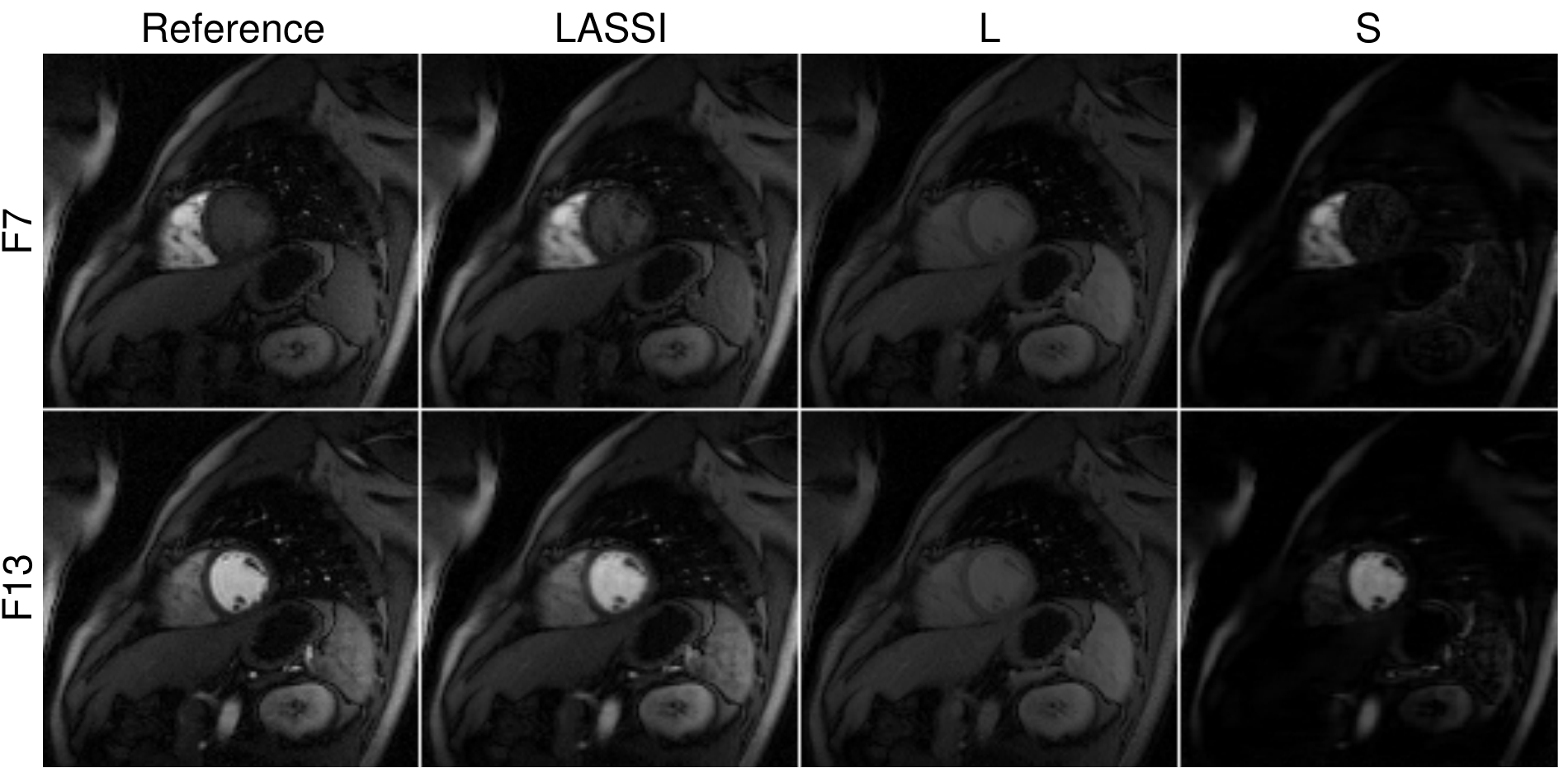}\\
 (g) & (h)\\
\end{tabular}
\caption{Behavior of the LASSI algorithms with Cartesian sampling and 8x undersampling. The algorithms are labeled according to the method used for $x_{L}$ update, i.e., SVT or OptShrink (OPT), and according to the type of sparsity penalty employed for the patch coefficients ($\ell_{0}$ or $\ell_{1}$ corresponding to (P1) or (P2)).
(a) Objectives (shown only for the algorithms for (P1) and (P2) with SVT-based updates, since OPT-based updates do not correspond to minimizing a formal cost function);
(b) NRMSE; (c) Sparsity fraction of $Z$ (i.e., $\left \| Z \right \|_{0}/mM$) expressed as a percentage; (d) normalized changes between successive dMRI reconstructions $\left \| x_{L}^{t}+x_{S}^{t} - x_{L}^{t-1} - x_{S}^{t-1} \right \|_{2}/$ $\left \| x_{\mathrm{ref}} \right \|_{2}$; (e) real and (f) imaginary parts of the atoms of the learned dictionaries in LASSI (using $\ell_{0}$ sparsity penalty and OptShrink-based $x_{L}$ update) shown as patches -- only the $8 \times 8$ patches corresponding to the first time-point (column) of the rank-1 reshaped ($64 \times 5$) atoms are shown; and frames 7 and 13 of the (g) conventional L+S reconstruction \cite{ota1} and (h) the proposed LASSI  (with $\ell_{0}$ penalty and OptShrink-based $x_{L}$ update) reconstruction shown along with the corresponding reference frames. The low-rank (L) and (transform or dictionary) sparse (S) components of each reconstructed frame are also individually shown. Only image magnitudes are displayed in (g) and (h).}
\label{imcvbcs}
\end{center}
\vspace{-0.25in}
\end{figure*}

Here, we consider the fully sampled cardiac perfusion data in \cite{ota1, ota2} and perform eight fold Cartesian undersampling of k-t space.
We study the behavior of the proposed LASSI algorithms for reconstructing the dMRI data from (multi-coil) undersampled measurements.
We consider four different LASSI algorithms in our study here: the algorithms for (P1) (with $\ell_{0}$ ``norm'') and (P2) (with $\ell_{1}$ norm) with SVT-based $x_{L}$ update; and the variants of these two algorithms where the SVT update step is replaced with an OptShrink (OPT)-type update.
The other variants of the SVT update including hard thresholding of singular values or updating based on the Schatten $p$-norm are studied later in Section \ref{sec4d}.
%The weights $\lambda_{L}$, $\lambda_{S}$, and $\lambda_{Z}$ were tuned for each LASSI algorithm to achieve good NRMSE.
We learned $320 \times 320$ dictionaries (with atoms reshaped by the operator $R_{2}(\cdot)$ into $64 \times 5$ space-time matrices) for the patches of $x_{S}$ with $r=1$, and $x_{L}$ and $x_{S}$ were initialized using the corresponding components of the L+S method with $\lambda_{L}=1.2$ and $\lambda_{S}=0.01$ in (P0) \cite{ota1}.
Here, we jointly tuned $\lambda_{L}$, $\lambda_{S}$, and $\lambda_{Z}$ for each LASSI variation, to achieve the best NRMSE.

%The weights $\lambda_{L}$, $\lambda_{S}$, and $\lambda_{Z}$ for each LASSI algorithm were jointly tuned over a range (grid) of values, and the settings achieving the best NRMSE were selected for each method here. The parameter values chosen here for the four LASSI variations are included in the supplement for completeness.

Fig. \ref{imcvbcs} shows the behavior of the proposed LASSI reconstruction methods. The objective function values (Fig. \ref{imcvbcs}(a)) in (P1) and (P2) decreased monotonically and quickly for the algorithms with SVT-based $x_{L}$ update.
The OptShrink-based $x_{L}$ update does not correspond to minimizing a formal cost function, so the OPT-based algorithms are omitted in Fig. \ref{imcvbcs}(a).
%The NRMSE (Fig. \ref{imcvbcs}(b)) improved over the iterations in all four LASSI methods and converged, with the $\ell_{0}$ ``norm''-based methods outperforming the $\ell_{1}$ penalty methods. 
All four LASSI methods improved the NRMSE over iterations compared to the initialization.
%NRMSE for initialization is 15.1\%
The NRMSE converged (Fig. \ref{imcvbcs}(b)) in all four cases, with the $\ell_{0}$ ``norm''-based methods outperforming the $\ell_{1}$ penalty methods.
Moreover, when employing the $\ell_{0}$ sparsity penalty, the OPT-based method ($r_{L}=1$) outperformed the SVT-based one for the dataset.
The sparsity fraction ($\left \| Z \right \|_{0}/mM$) for the learned coefficients matrix (Fig. \ref{imcvbcs}(c)) converged to small values (about 10-20 \%) in all cases indicating that highly sparse representations are obtained in the LASSI models.
Lastly, the difference between successive dMRI reconstructions (Fig. \ref{imcvbcs}(d)) quickly decreased to small values, suggesting iterate convergence.

Figs. \ref{imcvbcs}(g) and (h) show the reconstructions\footnote{Gamma correction was used to better display the images in this work.} and $x_{L}$ and $x_{S}$ components of two representative frames produced by the L+S \cite{ota1} (with parameters optimized to achieve best NRMSE) and LASSI
(OPT update and $\ell_{0}$ sparsity)
% (with OptShrink-based update and $\ell_{0}$ sparsity) 
methods, respectively. 
The LASSI reconstructions are sharper and a better approximation of the reference frames (fully sampled reconstructions) shown. In particular, the $x_{L}$ component of the LASSI reconstruction is clearly low-rank, and the $x_{S}$ component captures the changes in contrast and other dynamic features in the data. On the other hand, the $x_{L}$ component of the conventional L+S reconstruction varies more over time (i.e., it has higher rank), and the $x_{S}$ component contains relatively little information. The richer $(x_{L},x_{S})$ decomposition produced by LASSI suggests that both the low-rank and adaptive dictionary-sparse components of the model are well-suited for dMRI.

% and better reconstruction

%Figs. \ref{imcvbcs}(g) and (h) also show the $x_{L}$ and $x_{S}$ components of the reconstructed frames. 

Figs. \ref{imcvbcs}(e) and (f) show the real and imaginary parts of the atoms of the learned $D$
%dictionary 
in 
%the 
LASSI
%algorithm 
with OptShrink-based $x_{L}$ updating and $\ell_{0}$ 
%(patch) 
sparsity.
Only the first columns (time-point) of the (rank-1) reshaped $64 \times 5$ atoms are shown as $8 \times 8$ patches.
The learned atoms contain rich geometric and frequency-like structures that were jointly learned with the dynamic signal components from limited k-t space measurements. 

%(e.g., diagonal patterns) that are absent in the corresponding initial DCT (not shown). 

%% file: resultsandcomparisons7.tex
\subsection{Dynamic MRI Results and Comparisons} \label{sec4c}

\begin{table}[t]
\centering
\fontsize{7}{9pt}\selectfont
\begin{tabular}{|c|c|c|c|c|c|c|}
\hline
Undersampling & 4x & 8x & 12x & 16x  & 20x & 24x \\
\hline
\hline
NRMSE (k-t SLR) \%   & 11.1    & 15.4    & 18.8    & 21.7     & 24.3     &  27.0      \\
\hline
NRMSE (L+S) \%   & 10.9    &  13.9   & 15.8    & 17.8     & 20.1     & 23.0       \\
\hline
NRMSE (DINO-KAT) \%   &  10.4   & 12.6    & 14.5    & 16.7     &  18.8    &  22.1      \\
\hline
NRMSE (LASSI) \%   &  \textbf{10.0}    &  \textbf{12.6}    &  \textbf{14.3}   & \textbf{16.1}     &  \textbf{17.6}    &  \textbf{20.2}      \\
\hline
\hline
Gain over k-t SLR (dB)   &  0.9    &  1.7   & 2.4    & 2.6     &  2.8    &  2.5      \\
\hline
Gain over L+S (dB)   &  0.7   &  0.8   & 0.9    &  0.9    &  1.2    &   1.2     \\
\hline
Gain over DINO-KAT (dB)   &  0.3   &  0.0   &  0.1   &  0.3    &  0.6    &  0.8      \\
\hline
\end{tabular}
\vspace{0.06in}
\caption{NRMSE values expressed as percentages for the L+S \cite{ota1}, k-t SLR \cite{lingal1}, and the proposed DINO-KAT dMRI and LASSI methods at several undersampling (acceleration) factors for the cardiac perfusion data \cite{ota1, ota2} with Cartesian sampling. The NRMSE gain (in decibels (dB)) achieved by LASSI over the other methods is also shown. The best NRMSE for each undersampling factor is in bold.} \label{tabk1b}
\vspace{-0.35in}
\end{table}

% and the improvement (Gain) by LASSI over the other methods at each undersampling factor is indicated in decibels (dB)

Here, we consider the fully sampled cardiac perfusion data \cite{ota1, ota2}, PINCAT data  \cite{lingal1, ktslr44}, and \emph{in vivo} myocardial perfusion data \cite{lingal1, ktslr44}, and simulate k-t space undersampling at various acceleration factors. Cartesian sampling was used for the first dataset, and pseudo-radial sampling was employed for the other two.
The performance of LASSI and DINO-KAT dMRI is compared to that of L+S \cite{ota1} and k-t SLR \cite{lingal1}.
The LASSI and DINO-KAT dMRI algorithms were simulated with an $\ell_{0}$ sparsity penalty and a $320 \times 320$ dictionary. OptShrink-based $x_{L}$ updates were employed in LASSI for the cardiac perfusion data, and SVT-based updates were used in the other cases.
For the cardiac perfusion data, the initial $x_{L}$ and $x_{S}$ in LASSI were from the L+S framework \cite{ota1} (and the initial $x$ in DINO-KAT dMRI was an L+S dMRI reconstruction).
For the PINCAT and \emph{in vivo} myocardial perfusion data, the initial $x_{S}$ in LASSI (or $x$ in DINO-KAT dMRI) was the (better) \mbox{k-t} SLR reconstruction and the initial $x_{L}$ was zero.
All other settings are as discussed in Section \ref{sec4a}.

%\footnote{We used such an initialization as it led to slightly better NRMSE for LASSI in our experiments compared to alternative ones such as initializing both $x_{L}$ and $x_{S}$ using the corresponding outputs of the L+S method.}

%For the cardiac perfusion data, $x_{S}$ in LASSI (or $x$ in DINO-KAT dMRI) was initialized with the dMRI reconstruction from the L+S method \cite{ota1}, whereas it was initialized with the (better) k-t SLR reconstructions in the other cases, and $x_{L}$ was initialized to zero\footnote{We used such an initialization as it led to slightly better NRMSE for LASSI in our experiments compared to alternative ones such as initializing both $x_{L}$ and $x_{S}$ using the corresponding outputs of the L+S method.}. All other settings are as discussed in Section \ref{sec4a}.

\begin{table}[t]
\centering
\fontsize{7}{9pt}\selectfont
\begin{tabular}{|c|c|c|c|c|c|c|}
\hline
Undersampling & 5x & 6x & 7x & 9x  & 14x & 27x \\
\hline
\hline
NRMSE (k-t SLR) \%   & 9.7    &  10.7   & 12.2    & 14.5     &  18.0    &  23.7      \\
\hline
NRMSE (L+S) \%   & 11.7    & 12.8    & 14.2    & 16.3     & 19.6     &  25.4      \\
\hline
NRMSE (DINO-KAT) \%   & 8.6    &  9.5   &  10.7   &  12.6    & 15.9     &  21.8      \\
\hline
NRMSE (LASSI) \%   &  \textbf{8.4}    &  \textbf{9.1}    &  \textbf{10.1}   & \textbf{11.4}     &  \textbf{13.6}    &  \textbf{18.3}      \\
\hline
\hline
Gain over k-t SLR (dB)   &  1.2   &  1.4   & 1.7    &  2.1    &  2.4    &  2.2      \\
\hline
Gain over L+S (dB)   &  2.8   &  2.9   & 3.0    & 3.1     & 3.2     & 2.8       \\
\hline
Gain over DINO-KAT (dB)   & 0.2    &  0.3   &  0.6   &  0.9    &  1.4    & 1.5       \\
\hline
\end{tabular}
\vspace{0.06in}
\caption{NRMSE values expressed as percentages for the L+S \cite{ota1}, k-t SLR \cite{lingal1}, and the proposed DINO-KAT dMRI and LASSI methods at several undersampling (acceleration) factors for the PINCAT data  \cite{lingal1, ktslr44} with pseudo-radial sampling. The best NRMSE values for each undersampling factor are marked in bold.} \label{tabk2b}
\vspace{-0.2in}
\end{table}

\vspace{-0.02in}
Tables \ref{tabk1b}, \ref{tabk2b} and \ref{tabk3b} list the reconstruction NRMSE values for LASSI, DINO-KAT dMRI, L+S \cite{ota1} and k-t SLR \cite{lingal1} for the cardiac perfusion, PINCAT, and \emph{in vivo} datasets, respectively.
The LASSI method provides the best NRMSE values, and the proposed DINO-KAT dMRI method also outperforms the prior L+S and k-t SLR methods.
The NRMSE gains achieved by LASSI over the other methods are indicated in the tables for each dataset and undersampling factor.
The LASSI framework provides an average improvement of 1.9 dB, 1.5 dB, and 0.5 dB respectively, over the L+S, k-t SLR, and (proposed) DINO-KAT dMRI %reconstruction 
methods.
This suggests the suitability of the richer LASSI model for dynamic image sequences compared to the jointly low-rank and sparse (k-t SLR), low-rank plus non-adaptive sparse (L+S), and purely adaptive dictionary-sparse (DINO-KAT dMRI) signal models.

It is often of interest to compute the reconstruction NRMSE over a region of interest (ROI) containing the heart. Additional tables included in the supplement show the reconstruction NRMSE values computed over such ROIs for LASSI, DINO-KAT dMRI, L+S, and k-t SLR for the cardiac perfusion, PINCAT, and \emph{in vivo} datasets. The proposed LASSI and DINO-KAT dMRI methods provide much lower NRMSE in the heart ROIs compared to the other methods.

%The proposed LASSI and DINO-KAT dMRI methods provide much lower NRMSE in the heart ROIs compared to the other methods.

%he proposed LASSI methods usually provide much lower NRMSE in the heart ROIs compared to the other methods.

%The weights such as $\lambda_{L}$, $\lambda_{S}$, and $\lambda_{B}$ were tuned for each LASSI algorithm to achieve good NRMSE.

% and the improvement (Gain) by LASSI over the other methods at each undersampling factor is indicated in decibels (dB)

\begin{table}[t]
\centering
\fontsize{7}{9pt}\selectfont
\begin{tabular}{|c|c|c|c|c|c|c|}
\hline
Undersampling & 4x & 5x & 6x & 8x  & 12x & 23x \\
\hline
\hline
NRMSE (k-t SLR) \%   & 10.7    &  11.6   & 12.7    & 14.0     & 16.7     & 22.1       \\
\hline
NRMSE (L+S) \%   &  12.5   & 13.4    & 14.6    & 16.1     &  18.8    &  24.2      \\
\hline
NRMSE (DINO-KAT) \%   & 10.2    &  11.0   &   12.1  &  13.5    &  16.4    &  21.9      \\
\hline
NRMSE (LASSI) \%   &  \textbf{9.9}    &  \textbf{10.7}    &  \textbf{11.8}   & \textbf{13.2}     &  \textbf{16.2}    &  \textbf{21.9}      \\
\hline
\hline
Gain over k-t SLR (dB)   &  0.7   &  0.7   &  0.6   & 0.5     & 0.3     &  0.1      \\
\hline
Gain over L+S (dB)   &  2.1   & 2.0    & 1.8    &  1.7    & 1.3     &  0.9      \\
\hline
Gain over DINO-KAT (dB)   & 0.3    &  0.3   &  0.2   &  0.2    &  0.1    &  0.0      \\
\hline
\end{tabular}
\vspace{0.06in}
\caption{NRMSE values expressed as percentages for the L+S \cite{ota1}, k-t SLR \cite{lingal1}, and the proposed DINO-KAT dMRI and LASSI methods at several undersampling (acceleration) factors for the myocardial perfusion MRI data in \cite{lingal1, ktslr44}, using pseudo-radial sampling. The best NRMSE values for each undersampling factor are marked in bold.} \label{tabk3b}
\vspace{-0.25in}
\end{table}

%and the improvement (Gain) by LASSI over the other methods at each undersampling factor is indicated in decibels (dB)

\begin{figure}[!t]
\begin{center}
\begin{tabular}{cc}
\includegraphics[height=1.42in]{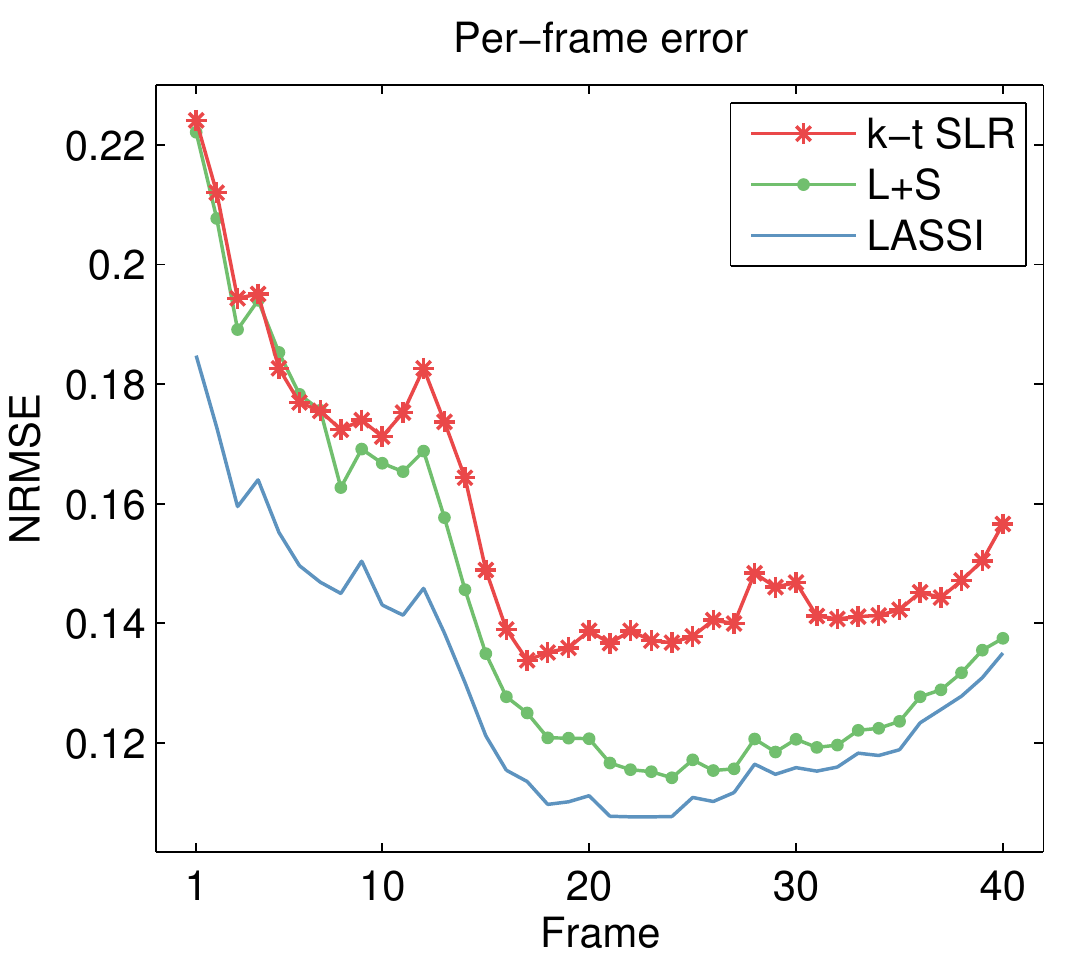}&
\includegraphics[height=1.42in]{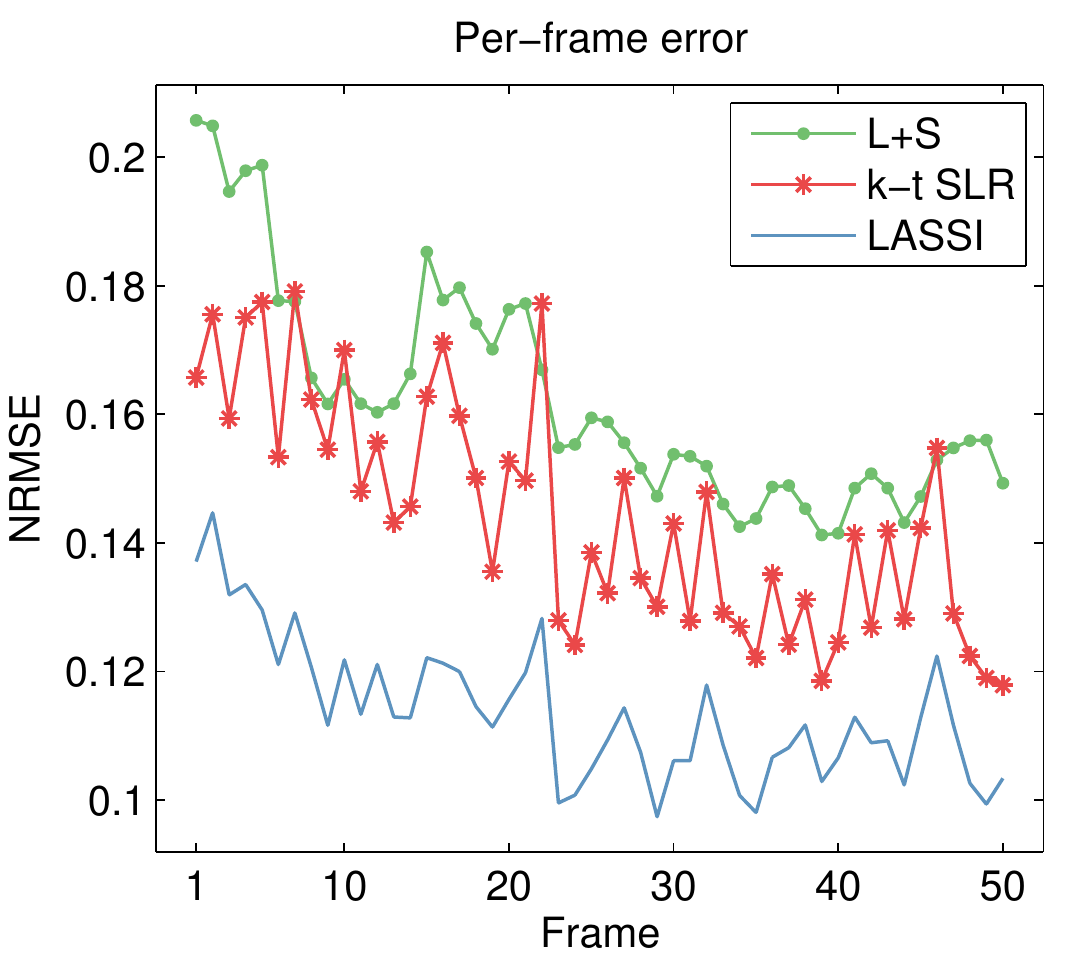}\\
(a) & (b) \\
\end{tabular}
\caption{NRMSE values computed between each reconstructed and reference frame for LASSI, L+S, and k-t SLR for (a) the cardiac perfusion data \cite{ota1, ota2} at 8x undersampling, and (b) the PINCAT data at 9x undersampling.}
\label{im100p2}
\end{center}
\vspace{-0.25in}
\end{figure}

\begin{figure}[!t]
\begin{center}
\begin{tabular}{c}
\hspace{-0.16in} \includegraphics[height=1.28in, trim={0 0 1cm 0}]{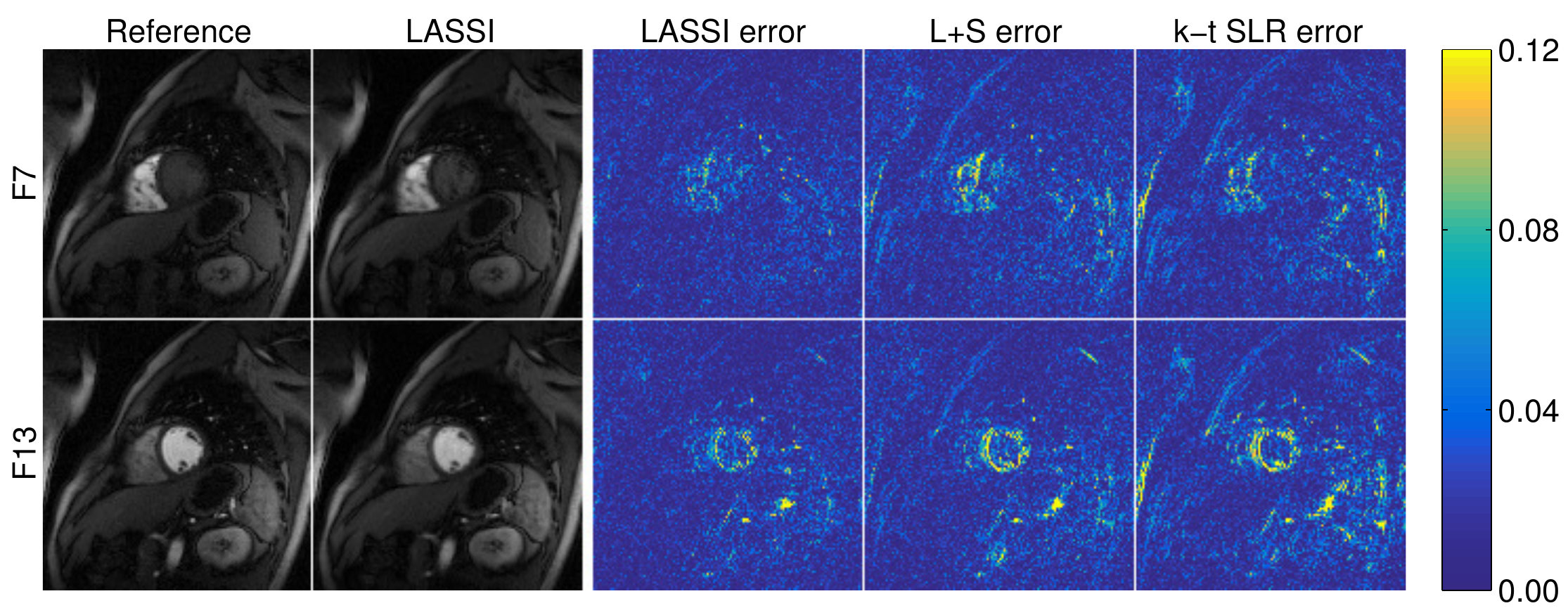}\\
\hspace{-0.16in} \includegraphics[height=1.28in, trim={0 0 1cm 0}]{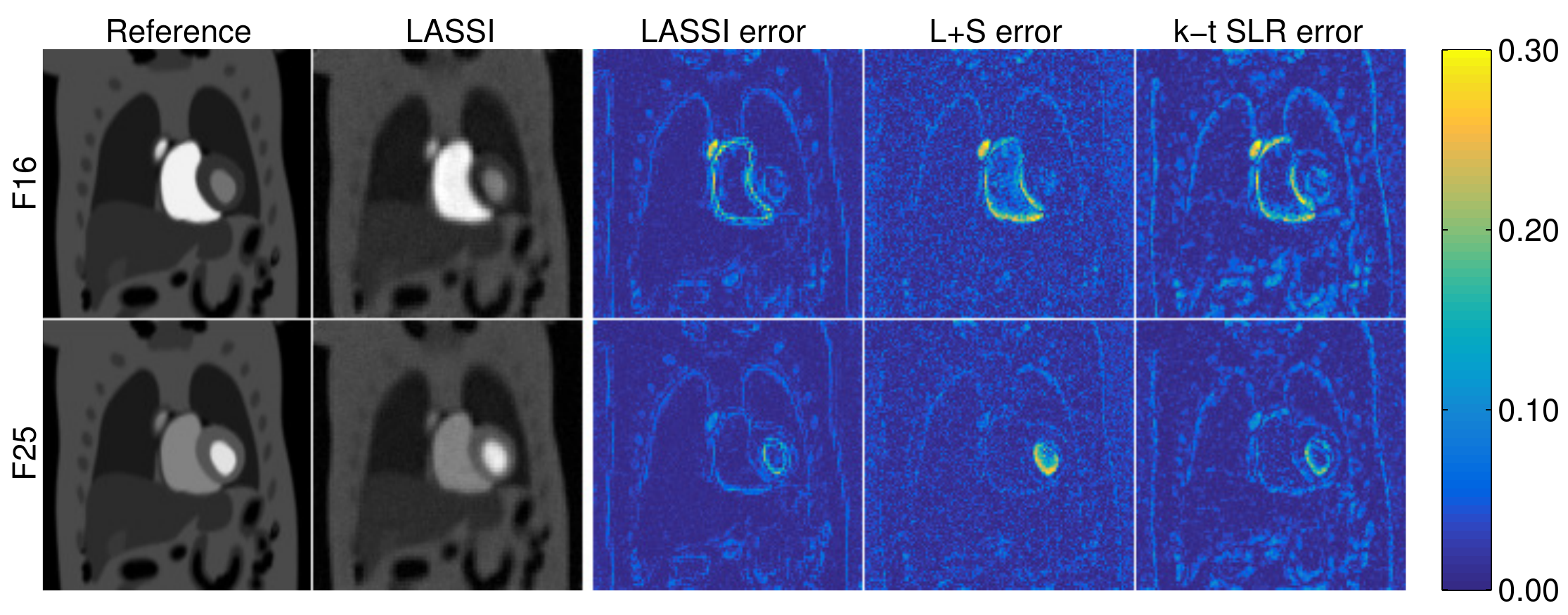}\\
\hspace{-0.16in} \includegraphics[height=1.81in, trim={0 0 1cm 0}]{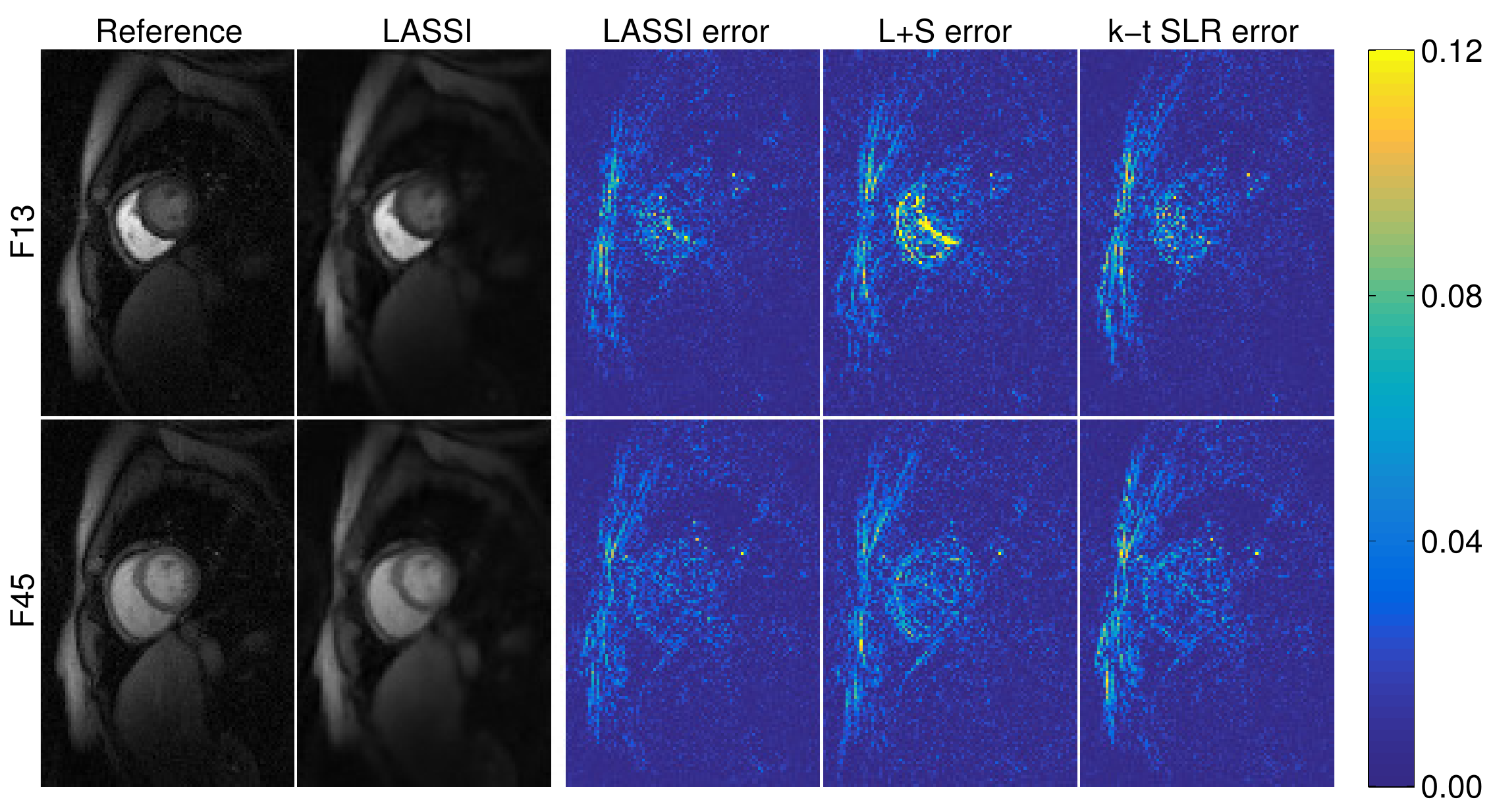}\\
\end{tabular}
\caption{LASSI reconstructions and the error maps (clipped for viewing) for LASSI, L+S, and k-t SLR for frames of the cardiac perfusion data \cite{ota1, ota2} (first row),  PINCAT data  \cite{lingal1, ktslr44} (second row), and \emph{in vivo} myocardial perfusion data \cite{lingal1, ktslr44} (third row), shown along with the reference reconstruction frames. Undersampling factors (top to bottom): 8x, 9x, and 8x. The frame numbers and method names are indicated on the images.}
\label{im100p}
\end{center}
\vspace{-0.25in}
\end{figure}

Fig. \ref{im100p2} shows the NRMSE values computed between each reconstructed and reference frame for the LASSI, L+S, and k-t SLR outputs for two datasets. The proposed LASSI scheme clearly outperforms the previous L+S and k-t SLR methods across frames (time).
Fig. \ref{im100p} shows the LASSI reconstructions of some representative frames (the supplement  shows more such reconstructions) for each dataset in Tables \ref{tabk1b}-\ref{tabk3b}. The reconstructed frames are visually similar to the reference frames (fully sampled reconstructions) shown.
Fig. \ref{im100p} also shows the reconstruction error maps (i.e., the magnitude of the difference between the magnitudes of the reconstructed and reference frames) for LASSI, L+S, and k-t SLR for the representative frames of each dataset.
The error maps for LASSI show fewer artifacts and smaller distortions than the other methods.
Results included in the supplement show that LASSI recovers temporal ($x-t$) profiles in the dynamic data with greater fidelity than other methods.

%%%%%%%%%%%%%%%%%%%%%%%%%
%%%Moved to supplement from here%%%%%%%%%%%
%%%%%%%%%%%%%%%%%%%%%%%%%%%%%%%%%%%%%

%\begin{figure*}[!t]
%\begin{center}
%\begin{tabular}{c}
%\includegraphics[height=2.2in]{figure4/tmi_xt_pincat}\\
%\end{tabular}
%\caption{A frame of the reference PINCAT \cite{lingal1, ktslr44} reconstruction is shown (left) with a spatial line cross section marked in green. The temporal ($x-t$) profiles of that line are shown for the reference, LASSI, DINO-KAT dMRI, L+S \cite{ota1}, and k-t SLR \cite{lingal1} reconstructions for pseudo-radial sampling and nine fold undersampling. The NRMSE values computed between the reconstructed and reference $x-t$ profiles are 0.107,  , 0.153, and 0.131 respectively, for LASSI, DINO-KAT dMRI, L+S, and k-t SLR.}
%\label{im35bcs33}
%\end{center}
%\end{figure*}

%Fig. \ref{im35bcs33} shows the reconstruction results for the PINCAT data at nine fold undersampling. The time series ($x-t$) plots, which correspond to the line marked in green on a reference PINCAT frame (Fig. \ref{im35bcs33}), are shown for the reference, LASSI, DINO-KAT dMRI, L+S \cite{ota1}, and k-t SLR \cite{lingal1} reconstructions. The result for LASSI clearly shows fewer artifacts and distortions (with respect to the reference) compared to the L+S and k-t SLR results. The LASSI result is also better than the DINO-KAT dMRI reconstruction that shows more smoothing (blur) effects (particularly in the top and bottom portions of the $x-t$ map).

%% file: parametermodelstudies6.tex
\subsection{A Study of Various LASSI Models and Methods} \label{sec4d}

%%%%%%%%%%%%%%%%%%%%%%%%%%%%%%%%%%%%%%%%%%%%%%%%%%
% OLD FIGURE: NO DINO-KAT
%\iffalse
%%%%%%%%%%%%%%%%%%%%%%%%%%%%%%%%%%%%%%%%%%%%%%%%%%
\begin{figure*}[!t]
\begin{center}
\begin{tabular}{cccc}
\includegraphics[height=1.4in]{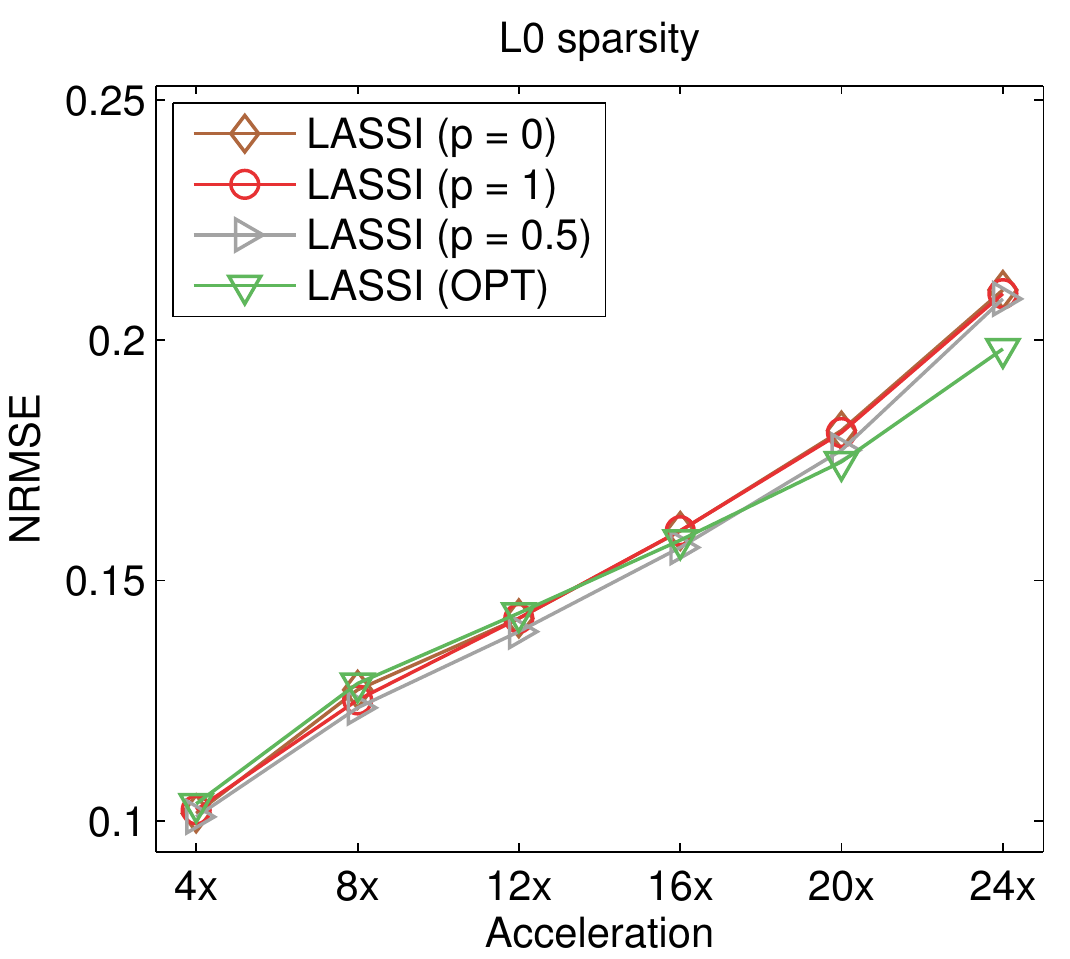}&
\includegraphics[height=1.4in]{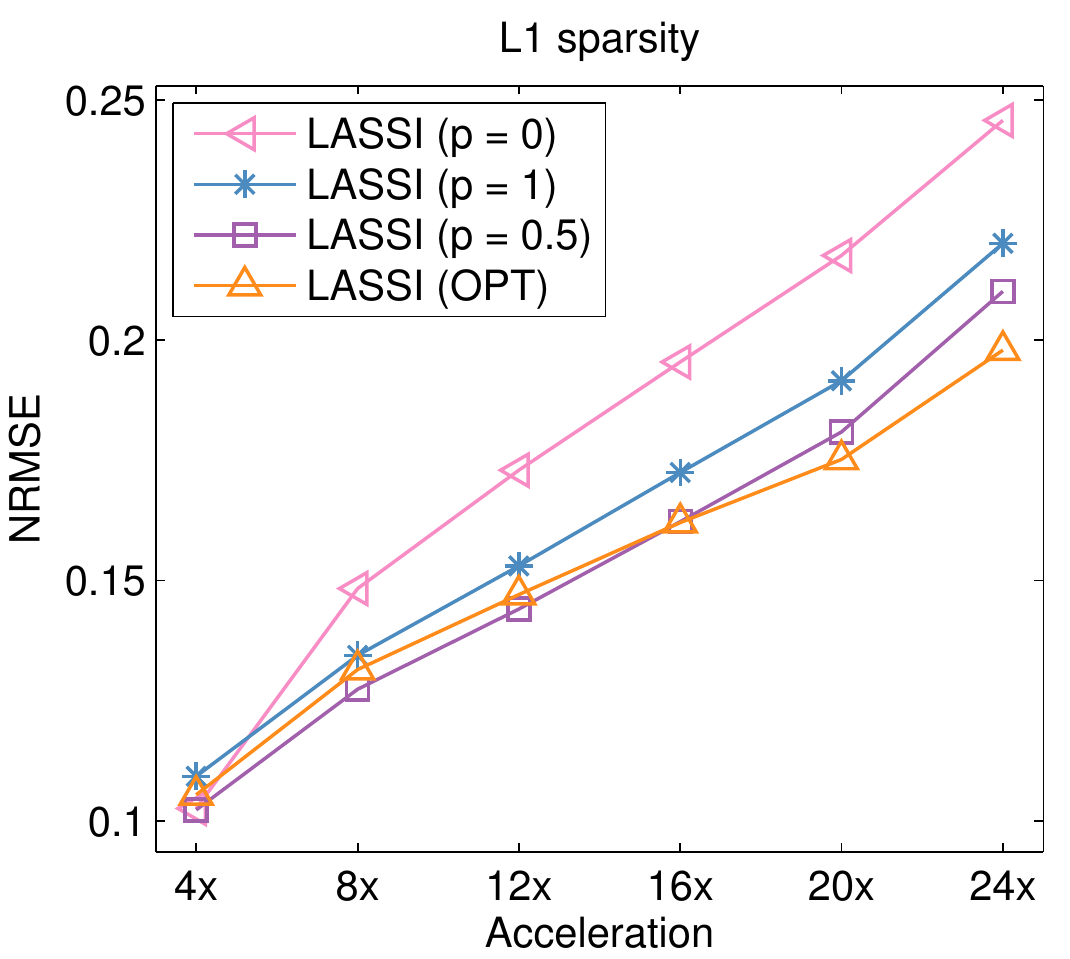}&
\includegraphics[height=1.4in]{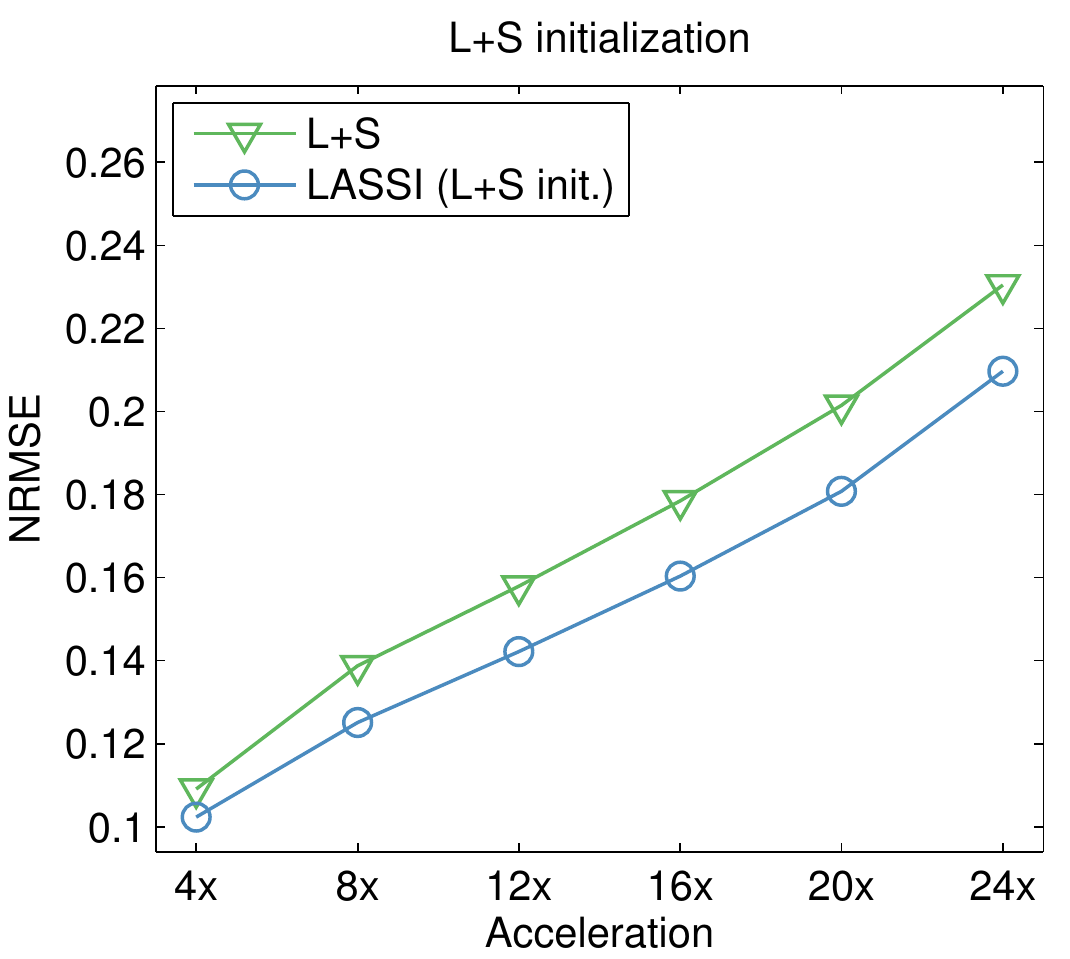}&
\includegraphics[height=1.4in]{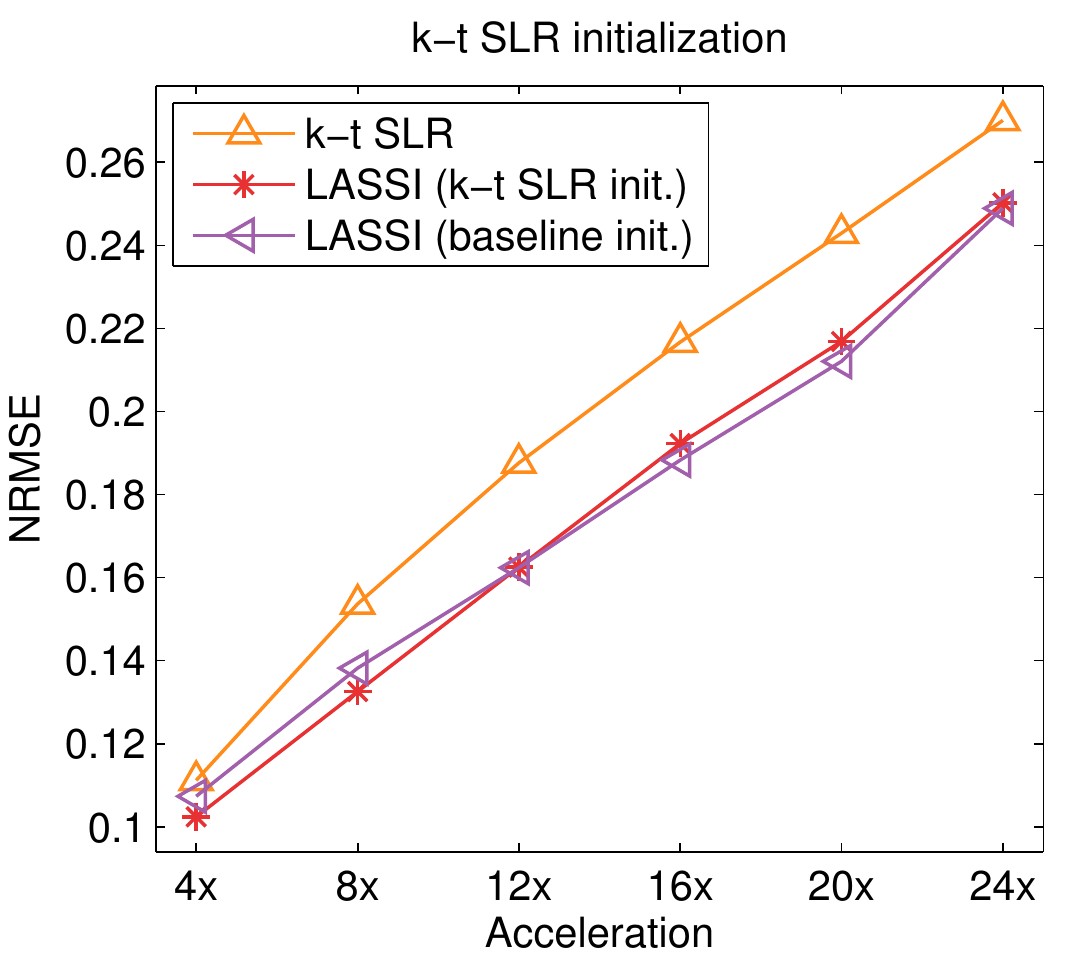}\\
(a) & (b) & (c) & (d)\\
\includegraphics[height=1.4in]{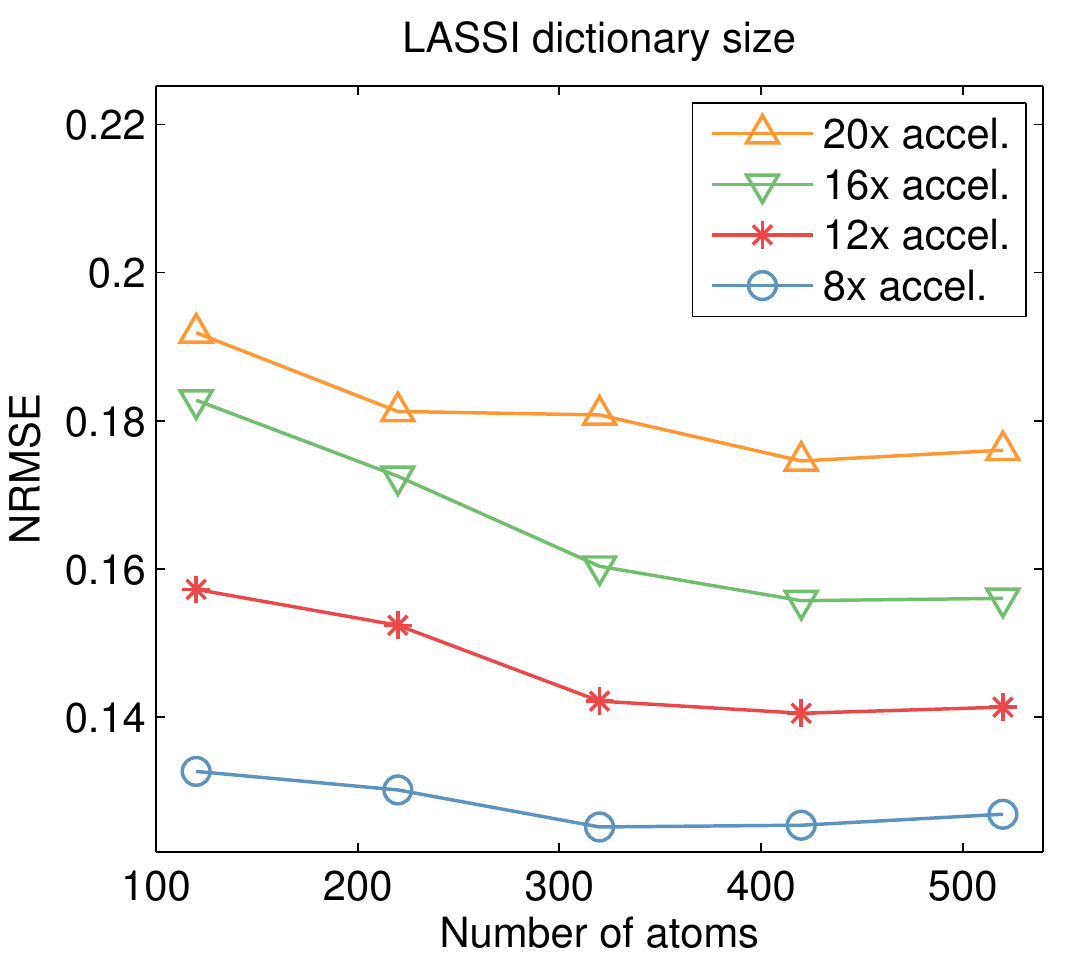}&
\includegraphics[height=1.4in]{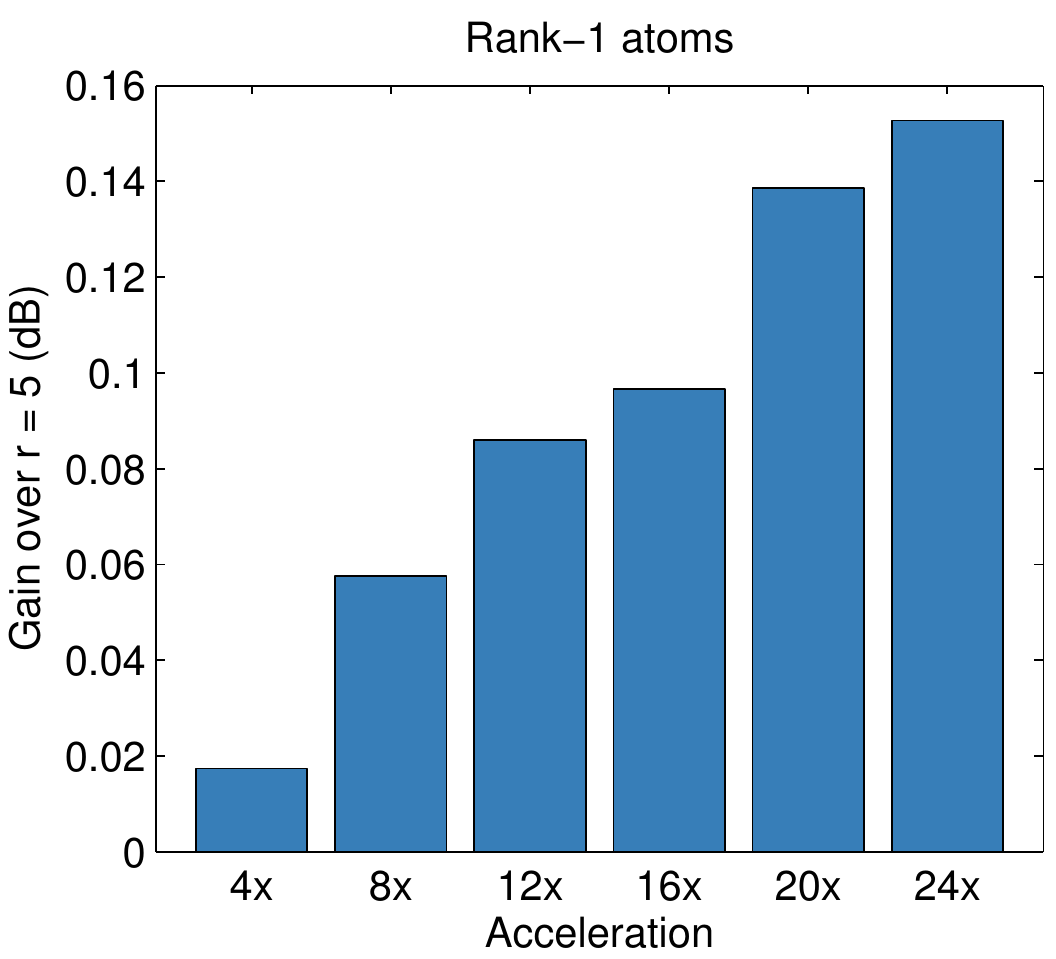}&
\includegraphics[height=1.4in]{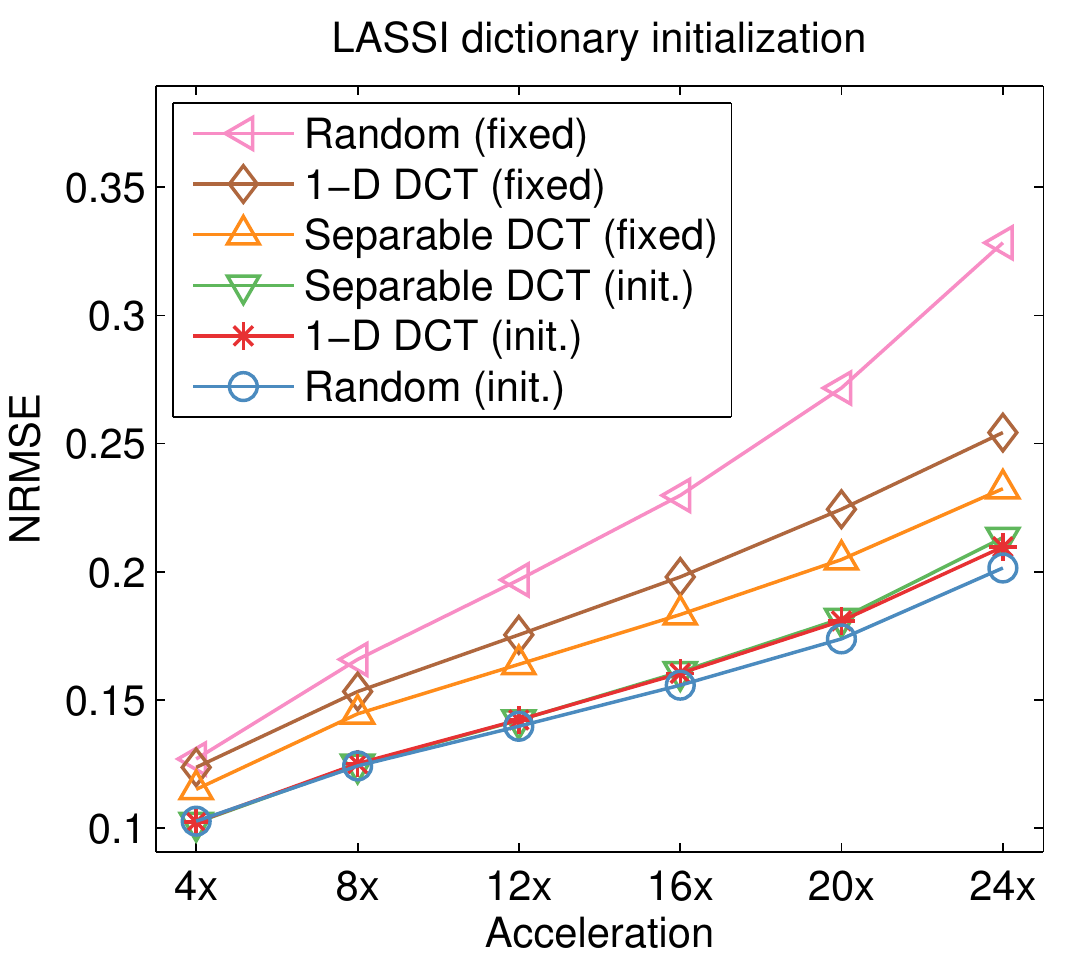}&
\includegraphics[height=1.4in]{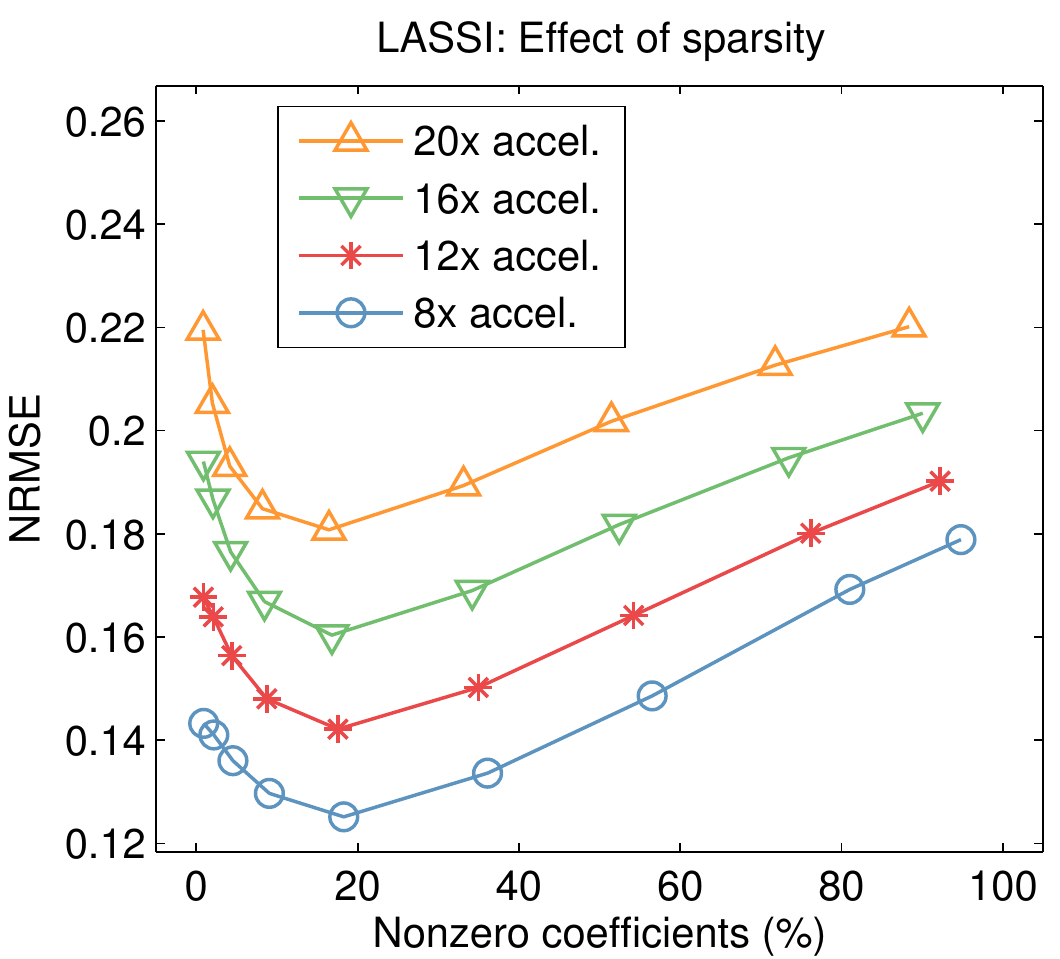}\\
(e) & (f) & (g) & (h) \\
\end{tabular}
\caption{Study of LASSI models, methods, and initializations at various undersampling (acceleration) factors for the cardiac perfusion data in \cite{ota1, ota2} with Cartesian sampling: (a) NRMSE for LASSI with $\ell_{0}$ ``norm'' for sparsity and with $x_{L}$ updates based on SVT ($p=1$), OptShrink (OPT), or based on the Schatten $p$-norm ($p=0.5$) or rank penalty ($p =0$); (b) NRMSE for LASSI with $\ell_{1}$ sparsity and with $x_{L}$ updates based on SVT ($p=1$), OptShrink (OPT), or based on the Schatten $p$-norm ($p=0.5$) or rank penalty ($p =0$); (c) NRMSE for LASSI when initialized with the output of the L+S method \cite{ota1} (used to initialize $x_{S}$ with $x_{L}^{0}=0$) together with the NRMSE for the L+S method; (d) NRMSE for LASSI when initialized with the output of the k-t SLR method \cite{lingal1} or with the baseline reconstruction (performing zeroth order interpolation at the nonsampled k-t space locations and then backpropagating to image space) mentioned in Section \ref{sec4a} (these are used to initialize $x_{S}$ with $x_{L}^{0}=0$), together with the NRMSE values for k-t SLR; (e) NRMSE versus dictionary size at different acceleration factors; (f) NRMSE improvement (in dB) achieved with $r=1$ compared to the $r=5$ case in LASSI; (g) NRMSE for LASSI with different dictionary initializations (a random dictionary, a $320 \times 320$ 1D DCT and a separable 3D DCT of the same size) together with the NRMSEs achieved in LASSI when the dictionary is fixed to its initial value; and (h) NRMSE versus the fraction of nonzero coefficients (expressed as percentage) in the learned $Z$ at different acceleration factors.}
\label{imcvbcs33}
\end{center}
\vspace{-0.28in}
\end{figure*}

Here, we investigate the various LASSI models and methods in 
%more 
detail. We work with the cardiac perfusion data \cite{ota1} 
%in this subsection 
and simulate the reconstruction performance of LASSI for Cartesian sampling at various undersampling 
%(acceleration) 
factors. %of k-t space.
Unless otherwise stated, we simulate LASSI here %in this subsection 
with the $\ell_{0}$ sparsity penalty, 
%(i.e., (P1)),
the SVT-based $x_{L}$ update, $r=1$, an initial $320 \times 320$ (1D) DCT dictionary, and $x_{S}$ initialized with the dMRI reconstruction from the L+S method \cite{ota1} and $x_{L}$ initialized to zero.
In the following, we first compare SVT-based updating of $x_{L}$ to alternatives in the algorithms and the use of $\ell_{0}$ versus $\ell_{1}$ sparsity penalties. The weights $\lambda_{L}$, $\lambda_{S}$, and $\lambda_{Z}$ were tuned for each LASSI variation.
%(i.e., with $\ell_{0}$/$\ell_{1}$ or SVT/OptShrink).
Second, we study the behavior of LASSI for different initializations of the underlying signal components or dictionary. Third, we study the effect of the number of atoms of 
%the dictionary 
$D$ on LASSI performance.
Fourth, we study the effect of the sparsity level 
%(number of nonzeros) 
of the learned $Z$ on the reconstruction quality in LASSI.
Lastly, we study the effect of the atom rank parameter $r$ in LASSI.

%versus OptShrink

\subsubsection{SVT vs. Alternatives and $\ell_{0}$ vs. $\ell_{1}$ patch sparsity} \label{sec4d1}

Figs. \ref{imcvbcs33}(a) and (b) show the behavior of the LASSI algorithms using $\ell_{0}$ and $\ell_{1}$ sparsity penalties, respectively.
In each case, the results obtained with $x_{L}$ updates based on SVT, OptShrink (OPT), or based on the Schatten $p$-norm ($p=0.5$), and rank penalty are shown.
The OptShrink-based singular value shrinkage (with $r_{L}=1$) and Schatten $p$-norm-based shrinkage typically outperform the conventional SVT (based on nuclear norm penalty) as well as the hard thresholding of singular values (for rank penalty) for the cardiac perfusion data.
The OptShrink and Schatten $p$-norm-based $x_{L}$ updates also perform quite similarly at lower undersampling factors, but OptShrink outperforms the latter approach at higher undersampling factors. 
Moreover, the $\ell_{0}$ ``norm''-based methods outperformed the corresponding $\ell_{1}$ norm methods in many cases (with SVT or alternative approaches).
These results demonstrate the benefits of appropriate nonconvex regularizers in practice.

%The results suggest benefits for good nonconvex regularizers in practice compared to the convex versions.  

\subsubsection{Effect of Initializations}

Here, we explore the behavior of LASSI for different initializations of the dictionary and the dynamic signal components.
First, we consider the LASSI algorithm initialized by
%with the outputs of 
the L+S and k-t SLR methods as well as with the baseline reconstruction (obtained by performing zeroth order interpolation at the nonsampled k-t space locations and then backpropagating to image space) mentioned in Section \ref{sec4a} (all other parameters fixed). The reconstructions from the prior methods are used to initialize $x_{S}$ in LASSI with $x_{L}^{0}=0$\footnote{We have also observed that LASSI improves the reconstruction quality over other alternative initializations such as initializing $x_{L}$ and $x_{S}$ using corresponding outputs of the L+S framework.}.
Figs. \ref{imcvbcs33}(c) and (d) show that LASSI significantly improves the dMRI reconstruction quality compared to the initializations
%(outputs of L+S and k-t SLR)
at all undersampling factors tested.
The baseline reconstructions had high NRMSE values (not shown in Fig. \ref{imcvbcs33}) of about 0.5. 
%Clearly, LASSI provides dramatic improvements over the baseline method.
Importantly, the reconstruction NRMSE for LASSI with the simple baseline initialization (Fig. \ref{imcvbcs33}(d)) is comparable to the NRMSE obtained with the more sophisticated k-t SLR initialization. In general, better initializations (for $x_{L}$, $x_{S}$) in LASSI may lead to a better final NRMSE in practice.

%or L+S
%especially at high undersampling factors
%LASSI provides dramatic improvements in NRMSE over the baseline method

%\footnote{We used such an initialization as it led to slightly better NRMSE for LASSI in our experiments compared to alternative ones such as initializing both $x_{L}$ and $x_{S}$ using the corresponding outputs of the L+S method.}

Next, we consider initializing the LASSI method with the following types of dictionaries (all other parameters fixed): a random i.i.d. gaussian matrix with normalized columns, the $320 \times 320$ 1D DCT, and the separable 3D DCT of size $320 \times 320$.
Fig. \ref{imcvbcs33}(g) shows that LASSI performs well for each choice of initialization.
We also simulated the LASSI algorithm by keeping the dictionary $D$ fixed (but still updating $Z$) to each of the aforementioned initializations.
%In this case, the separable DCT performs the best (Fig. \ref{imcvbcs33}(h)), while the random dictionary performs the worst.
Importantly, the NRMSE values achieved by the adaptive-dictionary LASSI variations are substantially better than the values achieved by the fixed-dictionary schemes.
%Importantly, the NRMSE values achieved in LASSI with fixed patch dictionaries are substantially worse than the values achieved by adapting the dictionary based on the imaging measurements. 

%Thus, although the LASSI formulations are nonconvex, the

\subsubsection{Effect of Overcompleteness of $D$} \label{sec4d3}

Fig. \ref{imcvbcs33}(e) shows the performance (NRMSE) of LASSI for various choices of the number of atoms ($K$) in $D$ at several acceleration factors. The weights in (P1) were tuned for each $K$.
%The weights in the LASSI formulation were tuned for each $K$.
As $K$ is increased, the NRMSE initially shows significant improvements (decrease) of more than 1 dB.
%The NRMSE generally improves (decreases) as $K$ increases. 
This is because LASSI learns richer models that provide sparser representations of patches and, hence, better reconstructions. However, for very large $K$ values, the NRMSE saturates or begins to degrade, since it is harder to learn very rich models using limited imaging measurements (without overfitting artifacts).

\subsubsection{Effect of the Sparsity Level in LASSI}
While Section \ref{sec4d1} compared the various ways of updating the low-rank signal component in LASSI, here we study the effect of the sparsity level of the learned $Z$ 
%(sparse coefficients matrix in (P1) or (P2)) 
on LASSI performance.
In particular, we simulate LASSI at various values of the parameter $\lambda_{Z}$ that controls sparsity (all other parameters fixed).
Fig. \ref{imcvbcs33}(h) shows the NRMSE of LASSI at various sparsity levels of the learned $Z$ and at several acceleration factors. 
The weight $\lambda_{Z}$ decreases from left to right in the plot and the same set of $\lambda_{Z}$ values were selected (for the simulation) at the various acceleration factors.
Clearly, the best NRMSE values occur
%are achieved 
around 10-20\% sparsity (when 32-64 dictionary atoms are used on the average to represent the reshaped $64 \times 5$ space-time patches of $x_{S}$), and the NRMSE degrades when the number of nonzeros in $Z$ is either too high (non-sparse) or too low (when the dictionary model reduces to a low-rank approximation of space-time patches in $x_{S}$).
This illustrates the effectiveness of the rich sparsity-driven modeling in LASSI\footnote{Fig. \ref{imcvbcs33}(h) shows that the same $\lambda_{Z}$ value is optimal at various accelerations. An intuitive explanation for this is that as the undersampling factor increases, the weighting of the (first) data-fidelity term in (P1) or (P2) decreases (fewer k-t space samples, or rows of the sensing matrix are selected). Thus, even with fixed $\lambda_{Z}$, the relative weighting of the sparsity penalty would increase, creating a stronger sparsity regularization at higher undersampling factors.}.

%Fig. \ref{imcvbcs33}(h) shows that the same $\lambda_{Z}$ is optimal at various acceleration factors. A perhaps intuitive explanation of this behavior is that as the undersampling factor increases, the weighting of the (first) data-fidelity term in (P1) or (P2) decreases (because fewer k-t space samples, or rows for the sensing matrix are selected).
%Thus, even with a fixed weight $\lambda_{Z}$, the relative weighting of the sparsity penalty would increase, which creates a stronger sparsity regularization at higher undersampling factors.

\subsubsection{Effect of Rank of Reshaped Atoms}

Here, we simulate LASSI with (reshaped) atom ranks $r=1$ (low-rank) and $r=5$ (full-rank).
Fig. \ref{imcvbcs33}(f) shows that LASSI with $r=1$ provides somewhat improved NRMSE values over the $r=5$ case at several undersampling factors, with larger improvements at higher accelerations. This result suggests that structured (fewer degrees of freedom) dictionary adaptation may be 
%particularly 
useful in scenarios involving very limited measurements.
In practice, the effectiveness of the low-rank model for reshaped dictionary atoms also depends on the properties of the underlying data.

%% file: conclusion5.tex
In this work, we investigated a novel framework for reconstructing spatiotemporal data from limited measurements. The proposed LASSI framework jointly learns a low-rank and dictionary-sparse decomposition of the underlying dynamic image sequence together with a spatiotemporal dictionary. The proposed algorithms involve simple updates.
Our experimental results showed the superior performance of LASSI methods for dynamic MR image reconstruction from limited k-t space data compared to recent works such as L+S and k-t SLR.
The LASSI framework also outperformed the proposed efficient dictionary-blind compressed sensing framework (a special case of LASSI) called DINO-KAT dMRI.
We also studied and compared various LASSI methods and formulations such as with $\ell_{0}$ or $\ell_{1}$ sparsity penalties, or with low-rank or full-rank reshaped dictionary atoms, or involving singular value thresholding-based optimization versus some alternatives including OptShrink-based optimization. The usefulness of LASSI-based schemes in other inverse problems and image processing applications merits further study.
The LASSI schemes involve parameters (like in most regularization-based methods) that need to be set (or tuned) in practice. We leave the study of automating the parameter selection process to future work. The investigation of dynamic image priors that naturally lead to OptShrink-type low-rank updates in the LASSI algorithms is also of interest, but is beyond the scope of this work, and will be presented elsewhere.

% (structured)

%In particular, our low-rank and adaptive sparse signal (LASSI) framework outperformed the previous L+S method as well as image reconstruction involving only dictionary learning. We also studied and compared various LASSI models and formulations such as with $\ell_{0}$ or $\ell_{1}$ sparsity penalties, or with low-rank or full-rank reshaped dictionary atoms, or involving singular value thresholding-based optimization versus OptShrink-based optimization. The usefulness of LASSI-based schemes in other inverse problems and image processing applications merits further study.

%% file: atomupdate.tex
\newtheorem{theorem}{Theorem}
\newtheorem{corollary}{Corollary}
\newtheorem{lemma}{Lemma}
\newtheorem{prop}{Proposition}

Here, we provide the proof of the low-rank atom update formula in Section III of our manuscript \citeSupp{saibrianrajfes:supp}. The corresponding optimization problem is as follows:
\begin{align} \label{eqop6gv}
 \min_{d_{i} \in \mathbb{C}^{m}} \ \ & \begin{Vmatrix} E_{i} - d_{i}c_{i}^{H} \end{Vmatrix}_{F}^{2}  \\
 \nonumber \mathrm{s.t.} \ \ &  \text{rank}\left ( R_{2}(d_{i}) \right ) \leq r, \, \left \| d_{i} \right \|_2 =1
\end{align}
where $E_{i} \triangleq P - \sum_{k\neq i} d_{k}c_{k}^{H}$ is computed using the most recent estimates of the variables, $P$ denotes the matrix that has the patches $P_{j}x_{S}$ for $1\leq j \leq M$ as its columns, and $d_{i}$ and $c_{i}$ are the $i$th columns of $D$ and $C = Z^{H}$, respectively.
The following Proposition \ref{prop2} provides the solution to Problem \eqref{eqop6gv}. It relies on the full singular value decomposition (SVD) of an appropriate matrix. We assume $R_{2}(d_{i})  \in \mathbb{C}^{m_{x} m_{y} \times m_{t}}$, and let $\sigma_{i}$ denote the $i$th entry on the main diagonal of the matrix $\Sigma$. 

\begin{prop}\label{prop2} \vspace{0.02in}
Given $E_{i} \in \mathbb{C}^{m \times M}$ and $c_{i} \in \mathbb{C}^{M}$, let $U_{r} \Sigma_{r} V_{r}^{H}$ denote an optimal rank-$r$ approximation to $ R_{2}\left ( E_{i}c_{i} \right )  \in \mathbb{C}^{m_{x} m_{y} \times m_{t}}$ that is obtained using the $r$ leading singular vectors and singular values of the full SVD $ R_{2}\left ( E_{i}c_{i} \right ) \triangleq U \Sigma V^{H}$. Then, a global minimizer in Problem \eqref{eqop6gv}, upon reshaping, is
\begin{equation} \label{tru1ch4ggv}
R_{2}(\hat{d}_{i}) =  \left\{\begin{matrix}
\frac{U_{r} \Sigma_{r} V_{r}^{H}
}{\left \| \Sigma_{r} \right \|_{F}}, & \mathrm{if}\,\, c_{i}\neq 0 \\ 
W, & \mathrm{if}\,\, c_{i}= 0 
\end{matrix}\right. \\
\end{equation}
where $W$ is the reshaped first column of the $m \times m$ identity matrix. The solution is unique if and only if $c_{i}\neq 0$, and $\sigma_{r}> \sigma_{r+1}$  or $\sigma_{r} = 0$.
\end{prop}

\textit{Proof:}
First, because $\left \| d_{i} \right \|_{2}=1$, the following result holds:
\begin{align} 
& \begin{Vmatrix}
E_{i} - d_{i}c_{i}^{H}
\end{Vmatrix}_{F}^{2}= \left \| E_{i} \right \|_{F}^{2} +  \left \| c_{i} \right \|_{2}^{2} - 2\, \text{Re}\left \{d_{i}^{H} E_{i} c_{i} \right \}.
\label{eqop12}
\end{align}
Upon substituting \eqref{eqop12} into \eqref{eqop6gv}, Problem \eqref{eqop6gv} simplifies to
\begin{align} \label{eqop6b}
\max_{d_{i} \in \mathbb{C}^{m}} \ \ & \text{Re} \left \{ \text{tr}\begin{pmatrix}
R_{2}(d_{i})^{H} R_{2}\begin{pmatrix}
E_{i} c_{i}
\end{pmatrix}
\end{pmatrix} \right \} \\
\nonumber \mathrm{s.t.} \ \ & \text{rank}\left ( R_{2}(d_{i}) \right ) \leq r,~\left \| d_{i} \right \|_2 =1.
\end{align}
Next, let $R_{2}(d_{i}) = G \Gamma B^{H}$, and $R_{2}\begin{pmatrix}
E_{i} c_{i}
\end{pmatrix}= U \Sigma V^{H}$ be full SVDs, with $\gamma_{k}$ and $\sigma_{k}$ the entries on the main diagonals of $\Gamma$ and $\Sigma$, respectively. The problem then becomes
% ($A, \Gamma, B, U, \Sigma, V$ are $\sqrt{n} \times \sqrt{n}$ matrices)
\begin{align} \label{eqop6b3}
\max_{\Gamma} \ & \ \max_{G,B} \ \ \text{Re} \begin{Bmatrix} \text{tr}\begin{pmatrix}
B \Gamma^{T} G^{H} U \Sigma V^{H}
\end{pmatrix} \end{Bmatrix}   \\
\nonumber \mathrm{s.t.} \ \ & \text{rank}(\Gamma) \leq r, \left \| \Gamma \right \|_F =1, G^{H}G=B^{H}B=I.
\end{align}
For the inner maximization above, we use $\text{Re}\left \{ \text{tr}\left ( B \Gamma^{T} G^{H} U\Sigma V^{H} \right ) \right \}$ $\leq \text{tr}\left ( \Gamma^{T} \Sigma \right )$ \citeSupp{matal:supp}, with the upper bound attained with $G=U$ and $B=V$. The remaining problem with respect to $\Gamma$ is then
\begin{align} \label{eqop6b4}
& \max_{\{\gamma_{k}\}} \sum_{k=1}^{r} \gamma_{k} \sigma_{k} \quad \mathrm{s.t.}\; \sum_{k=1}^{r} \gamma_{k}^{2}=1, \gamma_{j}=0 \, \forall \, j>r.
\end{align}
Using the Cauchy Schwarz inequality, $ \hat{\gamma}_{k} =$ $ \sigma_{k}/\sqrt{\sum_{k=1}^{r} \sigma_{k}^{2}}$ for $1 \leq k \leq r$, and $\hat{\gamma}_{k}=0$ for $k>r$ is clearly optimal. The derived solution for the optimal $R_{2}(\hat{d}_{i}) $ then simply corresponds to a normalized version of the rank-$r$ approximation to $ R_{2}\left ( E_{i}c_{i} \right )$. Clearly, the solution to \eqref{eqop6b} is unique if and only if $E_{i}c_{i} \neq 0$, and $\sigma_{r}> \sigma_{r+1}$  or $\sigma_{r}= \sigma_{r+1} = 0$. Any $d \in \mathbb{C}^{m}$ satisfying the constraints in \eqref{eqop6b} is a (non-unique) minimizer when $E_{i}c_{i}=0$. In particular $R_{2}(\hat{d}_{i})  =W$ works.

Lastly, to complete the Proposition's proof, we show that $E_{i}c_{i} = 0 $ in our algorithm if and only if $c_{i}=0$. Since $c_{i}$ here was obtained as a minimizer in the preceding sparse coding step, we have the following result $\forall$ $c \in \mathbb{C}^{M}$ with $\left \| c \right \|_{\infty} \leq a$ and $\tilde{d}_{i}$ denoting the $i$th atom in the preceding sparse coding step:
\begin{align} 
 & \hspace{-0.1in} \begin{Vmatrix}
E_{i} - \tilde{d}_{i}c_{i}^{H}
\end{Vmatrix}_{F}^{2} + \lambda_{Z}^{2} \left \| c_{i} \right \|_{0}  \leq \begin{Vmatrix}
E_{i} - \tilde{d}_{i}c^{H}
\end{Vmatrix}_{F}^{2} + \lambda_{Z}^{2} \left \| c \right \|_{0}.
\label{eqop16b}
\end{align}
If $E_{i}c_{i}=0$, the left hand side above is $\left \| E_{i} \right \|_{F}^{2}$ $ + \left \| c_{i} \right \|_{2}^{2}$ $+ \lambda_{Z}^{2} \left \| c_{i} \right \|_{0} $, which is clearly minimal (only) when $c_{i}=0$. Thus, when $E_{i}c_{i}=0$, we must have $c_{i}=0$.
$\;\;\; \blacksquare$

%% file: optshrinkbackground.tex
Here, we provide some additional detail about the OptShrink algorithm that we employ in Section III of our manuscript \citeSupp{saibrianrajfes:supp}. We begin by motivating the need for OptShrink by discussing the suboptimality of singular value thresholding (SVT) for low-rank matrix denoising, and then we explicitly state the algorithm.

In Section III of \citeSupp{saibrianrajfes:supp}, we argued that the LASSI low-rank update
\begin{equation} \label{Lupdate2}
R_{1}(x_{L}^{k}) = \mathbf{SVT}_{t_{k} \lambda_{L}}(R_{1}(\tilde{x}_{L}^{k-1}))
\end{equation}
can be interpreted as a low-rank denoising step, where the matrix $R_{1}(\tilde{x}_{L}^{k-1})$ is a noisy version of the underlying low-rank matrix $R_{1}(x_L)$ that we are interested in recovering, and the SVT is the chosen low-rank estimator.

A natural question to ask is what is the quality of the low-rank estimates produced by the SVT operator. To address this question, suppose that we are given a matrix $\widetilde{X} \in \mathbb{R}^{m \times n}$ of the form
\begin{equation} \label{eq:lowrankmodel2}
\widetilde{X} = \underbrace{\sum_{i=1}^{r} \theta_i u_i v_i ^{H}}_{=:L} + X,
\end{equation}
where $L$ is an unknown rank-$r$ matrix with singular values $\theta_i$ and singular vectors $u_i$ and $v_i$, and $X$ is an additive noise matrix. For example, in \eqref{Lupdate2}, we identity $L = R_{1}(x_L)$ and $X = R_{1}(\tilde{x}_{L}^{k-1} - x_L)$.

Now, consider the oracle low-rank denoising problem
\begin{equation} \label{eq:denoising2}
w^{\star} = \argmin_{[w_{1}, \ldots, w_{r}]^{T} \in \mathbb{R}^{r}} \Big\| \sum_{i=1}^{r} \theta_i u_i v_i ^{H} - \sum_{i=1}^{r} w_{i} \widetilde{u}_{i} \widetilde{v}_{i}^{H} \Big\|_{F},
\end{equation}
where $\widetilde{u}_i$ and $\widetilde{v}_i$ are the singular vectors of $\widetilde{X}$, and we denote its singular values by $\widetilde{\sigma}_i$. Problem \eqref{eq:denoising2} seeks the best approximation of the latent low-rank signal matrix $L$ by an optimally weighted combination of estimates of its left and right singular vectors. The truncated SVD (of rank $r$) and SVT are both feasible approaches for \eqref{eq:denoising2}. Indeed,  the truncated SVD corresponds to choosing weights $w_i = \widetilde{\sigma}_i \mathds{1}\{i \leq r\}$ and SVT with parameter $\tau \geq \widetilde{\sigma}_{r+1}$ corresponds to $w_i = (\widetilde{\sigma}_i - \tau)^+$. However, \eqref{eq:denoising2} can be solved in closed-form (see \citeSupp{nadakuditi2013:supp}), yielding the expression
\begin{equation}\label{eq:denoising solution2}
w^{\star}_{i} = \sum_{j=1}^{r} \theta_j\left(\widetilde{u}_{i}^{H} u_{j}\right) \left(\widetilde{v}_{i}^H v_{j}\right), \quad i=1,\ldots,r.
\end{equation}
Of course, \eqref{eq:denoising solution2} cannot be computed in practice because it depends on the latent low-rank singular vectors $u_i$ and $v_i$ that we would like to estimate, but it gives insight into the properties of the optimal weights $w^{\star}$. Indeed, when $\widetilde{u}_{i}$ and $\widetilde{v}_{i}$ are good estimates of $u_{i}$ and $v_i$, respectively, we expect $\widetilde{u}_{i}^{H} u_{i}$ and $\widetilde{v}_{i}^H v_{i}$ to be close to $1$. Consequently, from \eqref{eq:denoising solution2}, we expect ${w}^{\star}_{i}  \approx \theta_{i}$. Conversely, when $\widetilde{u}_{i}$ and $\widetilde{v}_{i}$ are poor estimates of $u_{i}$ and $v_{i}$, respectively, we expect $\widetilde{u}_{i}^{H} u_{i}$ and $v_{i}^{H} \widetilde{v}_{i}$ to be closer to $0$ and ${w}^{\star}_{i} < \theta_i$. In other words, \eqref{eq:denoising solution2} shows that the optimal singular value shrinkage is inversely proportional to the accuracy of the estimated principal subspaces. As a special case, if $\theta_i \rightarrow \infty$, then clearly $\widetilde{u}_{i}^{H} u_{i} \to 1$ and $v_{i}^{H} \widetilde{v}_{i} \to 1$, so the optimal weights $w^{\star}_i$ must have the property that the absolute shrinkage vanishes as $\theta_{i} \to \infty$. Consequently, the SVT operator, which applies a constant shrinkage to each singular value of its input, will necessarily produce suboptimal low-rank estimates in general. See \citeSupp{nadakuditi2013:supp} for more details.

The following theorem \citeSupp{nadakuditi2013:supp} formalizes the above argument under a probabilistic model for the additive noise matrix $X$.

\begin{theorem} \label{th:optshrink}
Suppose that $X_{ij}$ are i.i.d. random variables with zero-mean, variance $\sigma^2$, and bounded higher order moments, and suppose that $\theta_1 > \theta_2 > \ldots > \theta_r > \sigma$. Then, as $m,n \to \infty$ such that $m/n \to c \in (0, \infty)$, we have that
\begin{equation} \label{eq:asympweights}
{w}^{\star}_{i}  + 2 \ \displaystyle\frac{D_{\mu_{\widetilde{X}}}(\widetilde{\sigma}_i)}{D'_{\mu_{\widetilde{X}}}(\widetilde{\sigma}_i)} \convas 0 \qquad \textrm{for} \ \ i = 1, \ldots, r,
\end{equation}
where
\begin{equation} \label{eq:empmassfcn2bb}
\mu_{\widetilde{X}}(t) = \frac{1}{q-r} \sum_{i=r+1}^{q} \delta\left(t - \widetilde{\sigma}_i\right),
\end{equation}
with $q = \min(m,n)$ and the $D$-transform is defined as
\begin{equation} \label{eq:Dtransform2}
\begin{array}{rl}
D_{\mu_{\widetilde{X}}}(z) := &\left[ \displaystyle\int \frac{z}{z^2 - t^2} \ \mathrm{d}\mu_{\widetilde{X}}(t) \right] \times \\ \vphantom{2^{2^{2^{2^{2^{2^2}}}}}} &\left[ c \displaystyle\int \frac{z}{z^2 - t^2} \ \mathrm{d}\mu_{\widetilde{X}}(t) + \frac{1 - c}{z} \right],
\end{array}
\end{equation}
and $D_{\mu_{\widetilde{X}}}'(z)$ is the derivative of $D_{\mu_{\widetilde{X}}}(z)$ with respect to $z$.
\end{theorem}

Theorem~\ref{th:optshrink} establishes that the optimal weights ${w}^{\star}_{i}$ converge in the large matrix limit to a certain non-random integral transformation of the limiting noise distribution $\mu_{\widetilde{X}}$.

In practice, Theorem~\ref{th:optshrink} also suggests the following data-driven OptShrink estimator, defined for a given matrix $Y \in \mathbb{C}^{m \times n}$ and rank $r$ as
\begin{equation} \label{eq:optshrink_defn2}
\mathbf{OptShrink}_r(Y) = \sum_{i=1}^{r}  \left(-2 \ \displaystyle\frac{D_{\mu_Y}(\sigma_i)}{D'_{\mu_Y}(\sigma_i)} \right) u_i v_i^{H},
\end{equation}
where $Y = U  \Sigma V^H$ is the SVD of $Y$ with singular values $\sigma_i$, and
\begin{equation} \label{eq:empmassfcn2}
\mu_{Y}(t) = \frac{1}{q-r} \sum_{i=r+1}^{q} \delta\left(t - \sigma_i\right),
\end{equation}
is the empirical mass function of the noise-only singular values of $Y$, and $q = \min(m,n)$. By Theorem~\ref{th:optshrink}, $\mathbf{OptShrink}_r(\widetilde{X})$ asymptotically solves the oracle low-rank denoising problem \eqref{eq:denoising2}.

OptShrink has a single parameter $r \in \mathbb{N}$ that directly specifies the rank of its output matrix. Rather than applying a constant shrinkage to each singular value of the input matrix as in SVT, the OptShrink estimator partitions the singular values of its input matrix into \emph{signals} $\{\sigma_1,\ldots,\sigma_r\}$ and \emph{noise} $\{\sigma_{r+1},\ldots,\sigma_q\}$ and uses the empirical mass function of the noise singular values to estimate the optimal (nonlinear, in general) shrinkage \eqref{eq:asympweights} to apply to each signal singular value. See \citeSupp{nadakuditi2013:supp, benaych2012:supp} for additional detail.

The computational cost of OptShrink is the cost of computing a full SVD\footnote{In practice, one need only compute the singular values $\sigma_1,\ldots,\sigma_q$ and the leading $r$ singular vectors of $Y$.} plus the $O(r(m+n))$ computations required to compute the $D$-transform terms in \eqref{eq:optshrink_defn2}, which reduce to summations for the choice of ${\mu}_{Y}$ in \eqref{eq:empmassfcn2}.

%% file: additionalresults.tex
Here, we provide additional experimental results and details to accompany Section IV of our manuscript \citeSupp{saibrianrajfes:supp}.

\subsection{Dictionary Learning for Representing Dynamic Image Patches}

\begin{figure}[!t]
\begin{center}
\begin{tabular}{c}
\includegraphics[height=2.8in]{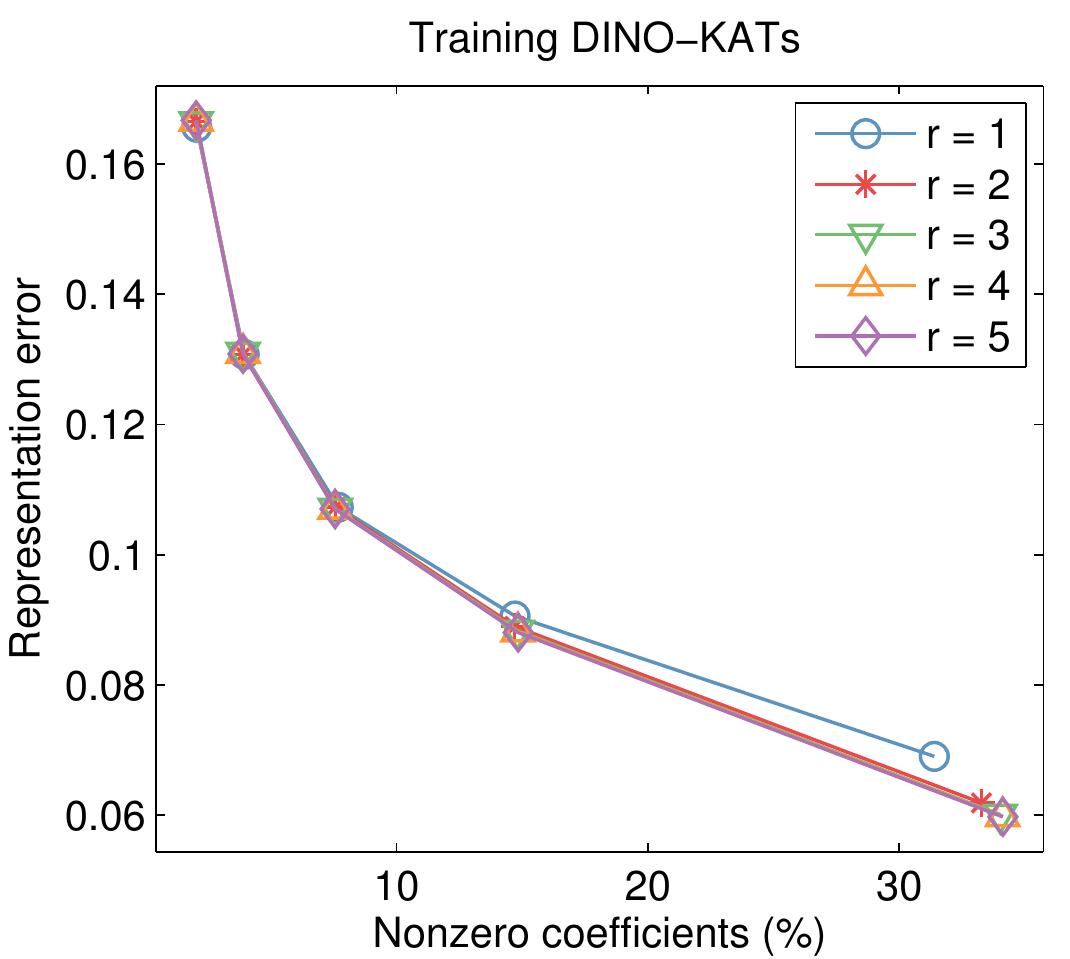}\\
\end{tabular}
\caption[]{The normalized sparse representation error (NSRE) $\left \| Y-DC^{H} \right \|_{F}/\left \| Y \right \|_{F}$ for the $320 \times 320$ dictionaries learned on the $8 \times 8 \times 5$ overlapping spatio-temporal patches of the fully sampled cardiac perfusion data \citeSupp{ota1:supp}.
The results are shown for various choices of the $\ell_{0}$ sparsity penalty parameter $\lambda_{Z}$ corresponding to different fractions of nonzero coefficients in the learned $C$ and  for various choices of the atom rank parameter $r$.}
\label{im305b}
\end{center}
%\vspace{-0.317in}
\end{figure}

Here, we present results on the effectiveness of learned (SOUP) dictionaries for representing dynamic image data. In particular, we compare dictionary learning with low-rank atom constraints to learning without such constraints.
We extract the $8 \times 8 \times 5$ overlapping spatio-temporal patches of the fully sampled cardiac perfusion data \citeSupp{ota1:supp}, with a spatial and temporal patch overlap stride of 2 pixels.
The vectorized 3D patches are then stacked as columns of the training matrix $P$, and we solve Problem (P3) in Section III.A-1 of \citeSupp{saibrianrajfes:supp} to learn the approximation $DC^{H}$ for $P$. In particular, we employ the iterative block coordinate descent method for (P3) that was discussed in Section III.A-1.
Dictionaries of size $320 \times 320$ (with atoms reshaped into $64 \times 5$ matrices) were learned for various values of the $\ell_{0}$ sparsity penalty parameter $\lambda_{Z}$ and for $r = 1, 2, 3, 4, \text{and} \; 5$. The block coordinate descent learning method ran for $50$ iterations and was initialized with $C=0$ and a $320 \times 320$ DCT.

The quality of the learned data approximations was measured using the normalized sparse representation error (NSRE) given as $\left \| Y-DC^{H} \right \|_{F}/\left \| Y \right \|_{F}$. Fig. \ref{im305b} shows the NSRE for various choices of $\lambda_{Z}$ corresponding to different fractions of nonzero coefficients in the learned $C$ and for various choices of the reshaped atom rank $r$. The learned dictionaries achieved small NSRE values together with sparse coefficients $C$. Importantly, the learned dictionaries with low-rank ($r<5$) reshaped atoms represented the spatio-temporal patches about as well as the learned dictionaries with full-rank ($r = 5$) atoms. Thus, the low-rank model on reshaped dictionary atoms, although a constrained model, effectively captures the properties of dynamic image patches.

\subsection{LASSI and DINO-KAT dMRI Algorithm Parameters}

This section lists the weights $\lambda_{L}$, $\lambda_{S}$, and $\lambda_{Z}$ used for the LASSI and DINO-KAT dMRI algorithms in Section IV of our manuscript \citeSupp{saibrianrajfes:supp}.

%chosen by tuning

First, we discuss the weights in LASSI.
Recall that for the cardiac perfusion data \citeSupp{ota1:supp, ota2:supp} (where the fully sampled dynamic data had a peak image intensity of 1.27) in Section IV.B, we performed eight fold Cartesian undersampling of k-t space,
and considered four different LASSI variations for dMRI reconstruction: the algorithms for Problem (P1) (with $\ell_{0}$ ``norm'') and Problem (P2) (with $\ell_{1}$ norm) with SVT-based $x_{L}$ update; and the variants of these two algorithms where the SVT update step is replaced with an OptShrink (OPT)-type update. Denote these methods as Algorithms 1, 2, 3, and 4, respectively.
The $\lambda_{L}$, $\lambda_{S}$, and $\lambda_{Z}$ values are  0.5, 0.01, and 0.03  for Algorithm 1, and  0.4,   0.02, and 0.01 for Algorithm 2. The $\lambda_{S}$ and $\lambda_{Z}$ settings were 0.01 and 0.025 for Algorithm 3, and 0.005 and 0.04 for Algorithm 4, and $r_{L}=1$.
The same values as in Section IV.B were used (and found to work well) for the corresponding algorithms in Table I of Section IV.C and in the experiments of Section IV.D, except in Section IV.D-2, where a higher $\lambda_{Z} = 0.06$ was used for the LASSI scheme with the simple baseline initialization.

In Section IV.D-1, the weight $\lambda_{L}$ for the LASSI algorithms based on the rank penalty or Schatten $p$-norm-based $x_{L}$ update was 4.5 and 1.5 respectively, when involving the $\ell_{0}$ ``norm'' for sparsity, and 12.5 and 2.5 respectively, when involving $\ell_{1}$ sparsity, and $\lambda_{S}$ and $\lambda_{Z}$ for these algorithms were identical to the settings for the corresponding SVT-based methods. 
In Section IV.C, $\lambda_{L}$, $\lambda_{S}$, and $\lambda_{Z}$ were chosen to be  3, 0.075, and 0.06 for the PINCAT data, and 0.05, 0.0025, and 0.01 for the \emph{in vivo} myocardial perfusion data (with the fully sampled dynamic data normalized to unit peak image intensity).

For DINO-KAT dMRI, $\lambda_{S}$ and $\lambda_{Z}$ were chosen to be 0.0075 and 0.025 for the cardiac perfusion data, 0.079 and 0.04 for the PINCAT data, and 0.0005 and 0.01 for the \emph{in vivo} myocardial perfusion data, respectively.

%the weights $\lambda_{S}$ and $\lambda_{Z}$ for the algorithms involving the rank penalty or Schatten $p$-norm-based $x_{L}$ updates were identical to those for the SVT-based approaches
%(obtained by zeroth order interpolation of dMRI measurements at nonsampled k-t space locations and then backpropagating to image space)

\subsection{Dynamic MRI Reconstruction Results}

%Dynamic MRI $x-t$ Profile Reconstructions

\begin{figure*}[!t]
\begin{center}
\begin{tabular}{c}
\includegraphics[height=2.2in]{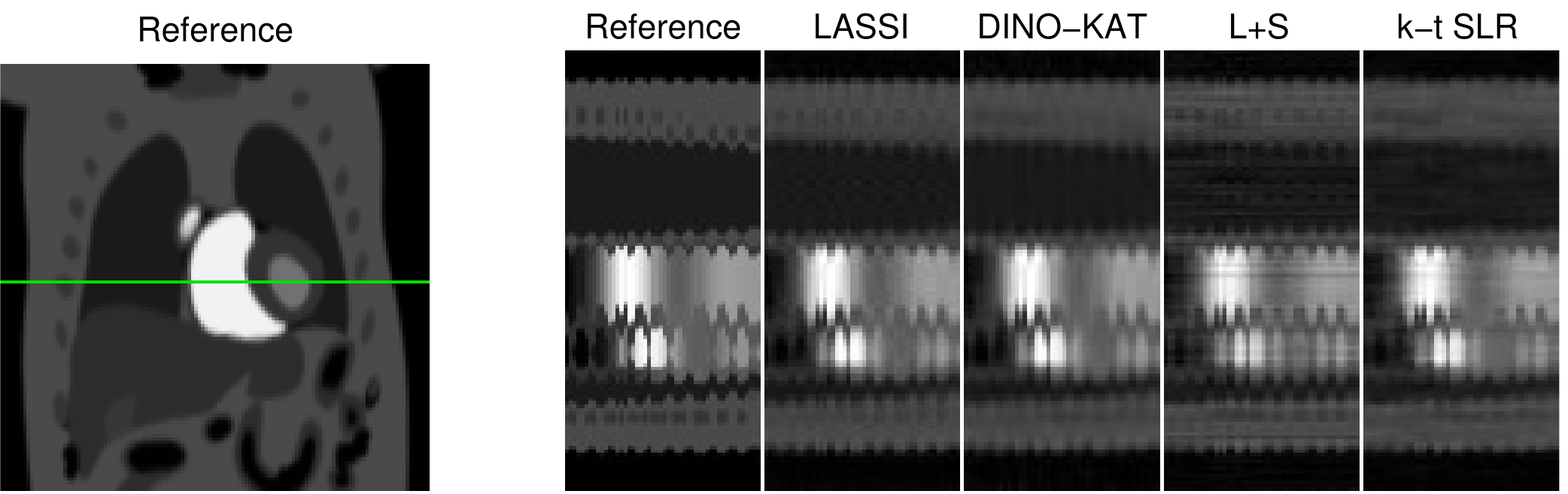}\\
\end{tabular}
\caption[]{A frame of the reference PINCAT \citeSupp{lingal1:supp, ktslr44:supp} reconstruction is shown (left) with a spatial line cross section marked in green. The temporal ($x-t$) profiles of that line are shown for the reference, LASSI, DINO-KAT dMRI, L+S \citeSupp{ota1:supp}, and k-t SLR \citeSupp{lingal1:supp} reconstructions for pseudo-radial sampling and nine fold undersampling. The NRMSE values computed between the reconstructed and reference $x-t$ profiles are 0.107, 0.116 , 0.153, and 0.131 respectively, for LASSI, DINO-KAT dMRI, L+S, and k-t SLR.}
\label{im35bcs33}
\end{center}
%\vspace{-0.317in}
\end{figure*}

\begin{figure*}[!t]
\begin{center}
\begin{tabular}{c}
\includegraphics[height=2.2in]{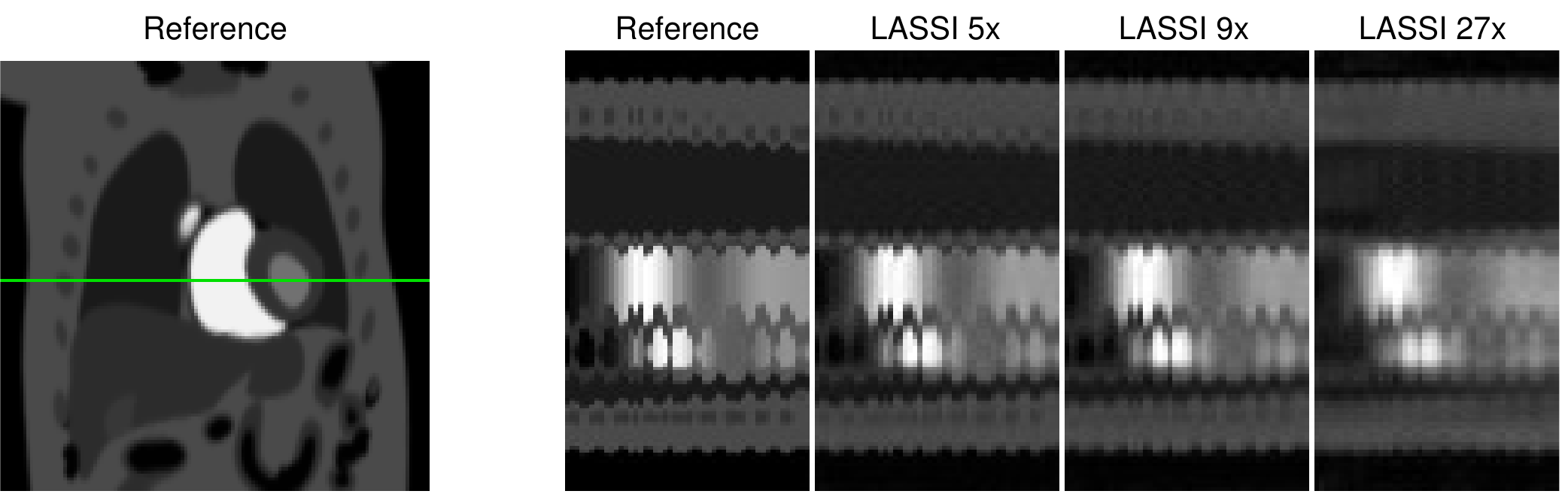}\\
\end{tabular}
\caption[]{A frame of the reference PINCAT \citeSupp{lingal1:supp, ktslr44:supp} reconstruction is shown (left) with a spatial line cross section marked in green. The temporal ($x-t$) profiles of that line are shown for the reference, and the LASSI reconstructions at 5x, 9x, and 27x undersampling and pseudo-radial sampling.}
\label{im35bcs33b}
\end{center}
%\vspace{-0.317in}
\end{figure*}

\begin{figure*}[!t]
\begin{center}
\begin{tabular}{c}
\includegraphics[height=2.05in]{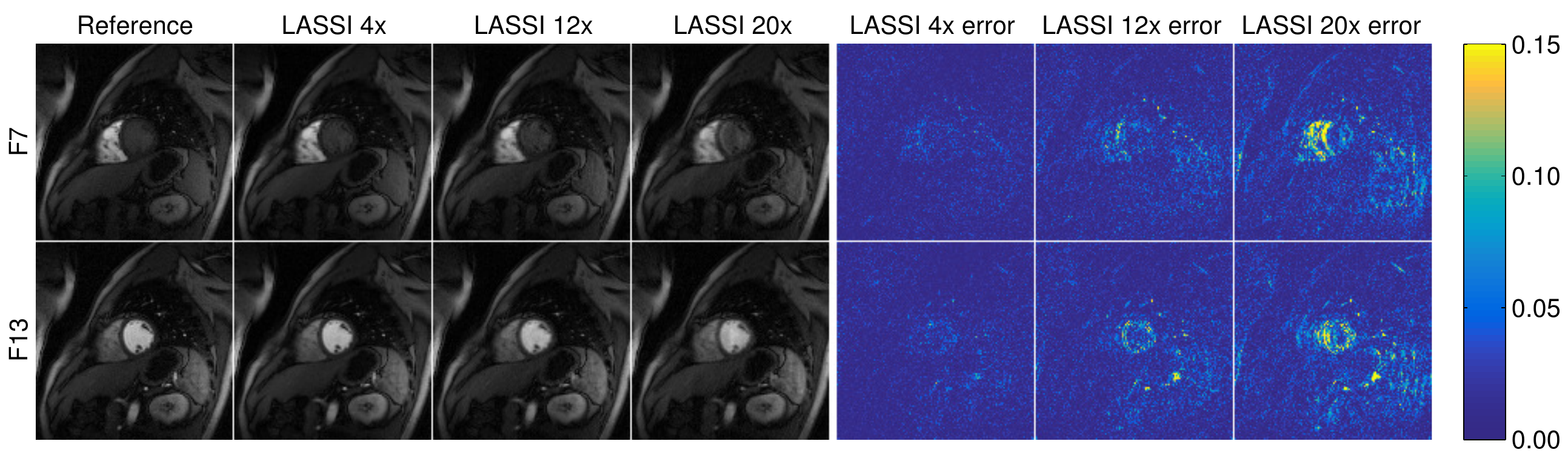}\\
\end{tabular}
\caption[]{LASSI reconstructions and error maps (clipped for viewing) for frames of the cardiac perfusion data  \citeSupp{ota1:supp, ota2:supp} at 4x, 12x, and 20x undersampling (Cartesian sampling), shown along with the reference reconstruction frames. The images are labeled with the frame numbers and undersampling factors.}
\label{im35bcs33c}
\end{center}
%\vspace{-0.317in}
\end{figure*}

Fig. \ref{im35bcs33} shows reconstruction results for the PINCAT data \citeSupp{lingal1:supp, ktslr44:supp} with pseudo-radial sampling and nine fold undersampling. The time series ($x-t$) plots, which correspond to the line marked in green on a reference PINCAT frame (Fig. \ref{im35bcs33}), are shown for the reference, LASSI, DINO-KAT dMRI, L+S \citeSupp{ota1:supp}, and k-t SLR \citeSupp{lingal1:supp} reconstructions. The NRMSE values computed between the reconstructed and reference $x-t$ slices are also shown.
The reconstruction for LASSI has lower NRMSE and clearly shows fewer artifacts and distortions (with respect to the reference) compared to the L+S and k-t SLR results. The LASSI result is also better than the DINO-KAT dMRI reconstruction that shows more smoothing (blur) effects (particularly in the top and bottom portions of the $x-t$ map).

Fig. \ref{im35bcs33b} shows time series ($x-t$) plots for the LASSI reconstructions of the PINCAT data at several undersampling factors. At an undersampling factor of 27x, the LASSI result shows temporal smoothing. Nevertheless, LASSI still reconstructs many features well, despite the high undersampling.
Fig. \ref{im35bcs33c} shows the LASSI reconstructions and reconstruction error maps for some representative frames of the cardiac perfusion data \citeSupp{ota1:supp, ota2:supp}, at several undersampling factors.
Notably, even at high undersampling factors, LASSI still accurately reconstructs many image features.

\subsection{Dynamic MRI Results over Heart ROIs}

\begin{figure}[!t]
\begin{center}
\begin{tabular}{ccc}
\includegraphics[height=1.12in]{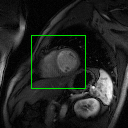}&
\includegraphics[height=1.12in]{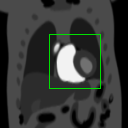}&
\includegraphics[height=1.12in]{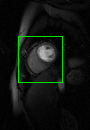}\\
(a) & (b) & (c) \\
\end{tabular}
\caption[]{Regions of interest containing the heart shown using green bounding boxes for a frame of (a) the cardiac perfusion data \citeSupp{ota1:supp}, (b) PINCAT data  \citeSupp{lingal1:supp, ktslr44:supp}, and (c) \emph{in vivo} myocardial perfusion MRI data \citeSupp{lingal1:supp, ktslr44:supp}, respectively.}
\label{im100p3}
\end{center}
%\vspace{-0.2in}
\end{figure}

\begin{table}[t]
\centering
\fontsize{7}{9pt}\selectfont
\begin{tabular}{|c|c|c|c|c|c|c|}
\hline
Undersampling & 4x & 8x & 12x & 16x  & 20x & 24x \\
\hline
\hline
NRMSE (k-t SLR) \%   & 10.4    & 14.2    & 17.2    & 19.5     & 22.4     &  24.2      \\
\hline
NRMSE (L+S) \%   & 10.7    &  14.0   & 16.3    & 18.8     & 22.2     & 24.1       \\
\hline
NRMSE (DINO-KAT) \%   &  9.8   & \textbf{12.5}    & \textbf{14.2}    & \textbf{16.4}     &  19.1    &  21.2      \\
\hline
NRMSE (LASSI) \%   &  \textbf{9.7}    &  12.7    &  14.4   & 16.7     &  \textbf{18.3}    &  \textbf{20.1}      \\
\hline
\hline
Gain over k-t SLR (dB)   &  0.6    &  0.9   & 1.5    & 1.4     &  1.8    &  1.6      \\
\hline
Gain over L+S (dB)   &  0.8   &  0.9   & 1.1    &  1.1    &  1.6    &   1.6     \\
\hline
Gain over DINO-KAT (dB)   &  0.1   &  -0.1   &  -0.1   &  -0.1    &  0.4    &  0.5      \\
\hline
\end{tabular}
\vspace{0.06in}
\caption[]{NRMSE values for an ROI (Fig. \ref{im100p3}(a)) in the cardiac perfusion data \citeSupp{ota1:supp} expressed as percentages for the L+S \citeSupp{ota1:supp}, k-t SLR \citeSupp{lingal1:supp}, and the proposed DINO-KAT dMRI and LASSI methods at several undersampling (acceleration) factors and Cartesian sampling. The NRMSE gain (in decibels (dB)) achieved by LASSI over the other methods is shown. The best NRMSE value at each undersampling factor is indicated in bold.} \label{tabk1b1}
%\vspace{-0.4in}
\end{table}

\begin{table}[t]
\centering
\fontsize{7}{9pt}\selectfont
\begin{tabular}{|c|c|c|c|c|c|c|}
\hline
Undersampling & 5x & 6x & 7x & 9x  & 14x & 27x \\
\hline
\hline
NRMSE (k-t SLR) \%   & 8.7    &  9.6   & 11.1    & 13.2     &  16.7    &  22.8      \\
\hline
NRMSE (L+S) \%   & 11.1    & 12.0    & 13.2    & 15.0     & 18.1     &  23.9      \\
\hline
NRMSE (DINO-KAT) \%   & 8.2    &  8.9   &  10.1   &  11.6    & 14.6     &  20.6      \\
\hline
NRMSE (LASSI) \%   &  \textbf{8.0}    &  \textbf{8.7}    &  \textbf{9.6}   & \textbf{10.9}     &  \textbf{13.2}    &  \textbf{18.3}      \\
\hline
\hline
Gain over k-t SLR (dB)   &  0.7   &  0.9   & 1.2    &  1.6    &  2.1    &  1.9      \\
\hline
Gain over L+S (dB)   &  2.9   &  2.8   & 2.8    & 2.8     & 2.8     & 2.3       \\
\hline
Gain over DINO-KAT (dB)   & 0.2    &  0.3   &  0.4   &  0.5    &  0.9    & 1.0       \\
\hline
\end{tabular}
\vspace{0.06in}
\caption[]{NRMSE values for an ROI (Fig. \ref{im100p3}(b)) in the PINCAT data  \citeSupp{lingal1:supp, ktslr44:supp} expressed as percentages for the L+S \citeSupp{ota1:supp}, k-t SLR \citeSupp{lingal1:supp}, and the proposed DINO-KAT dMRI and LASSI methods at several undersampling (acceleration) factors and pseudo-radial sampling. The best NRMSE value at each undersampling factor is indicated in bold.} \label{tabk2b1}
%\vspace{-0.3in}
\end{table}

\begin{table}[t]
\centering
\fontsize{7}{9pt}\selectfont
\begin{tabular}{|c|c|c|c|c|c|c|}
\hline
Undersampling & 4x & 5x & 6x & 8x  & 12x & 23x \\
\hline
\hline
NRMSE (k-t SLR) \%   & 7.6    &  8.3   & 9.2    & 10.4     & 12.4     & 17.1       \\
\hline
NRMSE (L+S) \%   &  9.2   & 10.0    & 11.0    & 12.3     &  14.5    &  18.9      \\
\hline
NRMSE (DINO-KAT) \%   & 7.1    &  7.8   &   8.7  &  10.0    &  12.0    &  16.8      \\
\hline
NRMSE (LASSI) \%   &  \textbf{6.8}    &  \textbf{7.5}    &  \textbf{8.4}   & \textbf{9.7}     &  \textbf{11.8}    &  \textbf{16.8}      \\
\hline
\hline
Gain over k-t SLR (dB)   &  0.9   &  0.9   &  0.8   & 0.6     & 0.4     &  0.2      \\
\hline
Gain over L+S (dB)   &  2.6   & 2.5    & 2.3    &  2.1    & 1.8     &  1.0      \\
\hline
Gain over DINO-KAT (dB)   & 0.4    &  0.4   &  0.3   &  0.2    &  0.1    &  0.0      \\
\hline
\end{tabular}
\vspace{0.06in}
\caption[]{NRMSE values for an ROI (Fig. \ref{im100p3}(c)) in the myocardial perfusion MRI data \citeSupp{lingal1:supp, ktslr44:supp} expressed as percentages for the L+S \citeSupp{ota1:supp}, k-t SLR \citeSupp{lingal1:supp}, and the proposed DINO-KAT dMRI and LASSI methods at several undersampling (acceleration) factors and pseudo-radial sampling. The best NRMSE value at each undersampling factor is indicated in bold.} \label{tabk3b1}
%\vspace{-0.25in}
\end{table}

Tables I-III in Section IV.C of \citeSupp{saibrianrajfes:supp} showed the NRMSE values of dynamic MRI reconstructions obtained by various methods for three datasets. Here, we report the NRMSE of the dynamic MRI reconstructions in Section IV.C, computed over specific regions of interest (ROIs) containing the heart.
Fig. \ref{im100p3} shows the ROIs (as a rectangular box in a frame) for the cardiac perfusion data \citeSupp{ota1:supp, ota2:supp}, PINCAT data  \citeSupp{lingal1:supp, ktslr44:supp}, and \emph{in vivo} myocardial perfusion MRI data \citeSupp{lingal1:supp, ktslr44:supp}.
Tables \ref{tabk1b1}, \ref{tabk2b1}, and \ref{tabk3b1} list the NRMSE values computed over these ROIs for the LASSI, DINO-KAT dMRI, L+S \citeSupp{ota1:supp}, and k-t SLR \citeSupp{lingal1:supp} reconstructions  at several undersampling factors.
The various methods tend to provide even better reconstruction quality (i.e., NRMSE) within the specific ROIs than over the entire images (cf. Tables I-III of  \citeSupp{saibrianrajfes:supp}).
Tables \ref{tabk1b1}-\ref{tabk3b1} also indicate the NRMSE gains achieved by LASSI over the other methods for each dataset and undersampling factor.
The proposed LASSI and DINO-KAT dMRI methods provide much lower NRMSE in the heart ROIs compared to the previous L+S and k-t SLR methods. The LASSI scheme also outperforms DINO-KAT dMRI in most cases, and provides an average improvement within the ROIs of 2.0 dB, 1.1 dB, and 0.3 dB respectively, over the L+S, k-t SLR, and the proposed DINO-KAT dMRI methods.